\documentclass{article}
\PassOptionsToPackage{numbers}{natbib}
\usepackage[preprint]{neurips_2020}

\usepackage[table,xcdraw]{xcolor}

\usepackage[utf8]{inputenc} 
\usepackage[T1]{fontenc}  
\usepackage{hyperref}    
\hypersetup{colorlinks=true, pagebackref=true, urlcolor=darkgray, citecolor=darkgray, linkcolor=darkgray}    

\usepackage{url}   
\usepackage{booktabs}    
\usepackage{amsfonts}    
\usepackage{nicefrac}    
\usepackage{microtype}   
\usepackage{amsmath}
\usepackage{cleveref}
 \usepackage{mathtools}
 \usepackage{mathrsfs}
 \usepackage{mathptmx}
 \usepackage{parskip}
 \usepackage{graphicx}    
\usepackage[shortlabels]{enumitem} 
\usepackage{soul}
\usepackage{lipsum}
\usepackage[pangram]{blindtext}
\usepackage[english]{babel}
\usepackage{bm}
\usepackage{subfloat}
\usepackage{subfig}
\usepackage{float}
\usepackage{booktabs}
\usepackage{tabularx}
\usepackage{makecell}
\usepackage{algorithm}
\usepackage[noend]{algpseudocode}
\usepackage{wrapfig}
\usepackage{placeins}
\usepackage{cleveref}

\title{
Pruning via Iterative Ranking of Sensitivity Statistics
}

\author{%
 Stijn Verdenius  \\
 University of Amsterdam\\
 \texttt{stijn@verdenius.com} \\
  \And
  Maarten Stol \\
  BrainCreators B.V. \\
  \texttt{maarten.stol@braincreators.com} \\
  \And
  Patrick Forr\'e \\
 University of Amsterdam\\
 \texttt{p.d.forre@uva.nl} \\
}

\newcommand{\minquad}{\;\;}

\newcommand{\iterativesnip}{SNIP-it}
\newcommand{\iterativestructuredsnip}{SNAP-it}
\newcommand{\combinediterativesnip}{CNIP-it}

\newcommand{\bb}[1]{\textbf{#1}}

\newcommand{\midd}[1]{\minquad &#1 \minquad}
\newcommand{\x}{$\times$}

\begin{document}

\maketitle

\begin{abstract}
With the introduction of SNIP \citep{lee2019snip}, it has been demonstrated that modern neural networks can effectively be pruned before training. Yet, its sensitivity criterion has since been criticized for not propagating training signal properly or even disconnecting layers. As a remedy, GraSP \citep{wang2020picking} was introduced, compromising on simplicity. However, in this work we show that by applying the sensitivity criterion iteratively in smaller steps - still before training - we can improve its performance without difficult implementation. As such, we introduce \emph{\lq\iterativesnip\rq}. We then demonstrate how it can be applied for both structured and unstructured pruning, before and/or during training, therewith achieving state-of-the-art sparsity-performance trade-offs. That is, while already providing the computational benefits of pruning in the training process from the start. Furthermore, we evaluate our methods on robustness to overfitting, disconnection and adversarial attacks as well.
\end{abstract}

\section{Introduction}
In the last decade, deep learning has undergone a period of rapid development. Since AlexNet's revolution \cite{krizhevsky2012imagenet}, advances have been noticeable on many fronts; for instance in generative image modeling \cite{kingma2013auto, goodfellow2014generative} and translation \cite{vaswani2017attention}. Be that as it may, deep learning also has some practical drawbacks. Amongst others, it is time-consuming, computationally expensive and data-hungry.

Literature has since suggested that modern neural networks are heavily overparametrised \citep{denil2013predicting, dauphin2013big, ba2014deep, han2015learning}, hypothesising that a great deal of the parameter-space that these networks are equipped with, is actually redundant. As such, the field of \emph{model compression} suggests networks can be pruned off their redundant model components, with promises of reduced storage and computational effort. This can be formalised in the following way; assume the existence of an uncompressed network $f(x ; \theta)$, that has a compressed counterpart $f(x ; \theta')$. Then, the latter is produced by some compression-algorithm, parametrised by hyperparameters and subjected to a sparsity constraint, such that:
\begin{align*}
\begin{split}
   f(x; \theta') \midd{\approx} f(x; \theta)\quad \forall_x \\
   \|\theta'\|_0 \midd{\ll} \|\theta\|_0
   \end{split}
\end{align*}
Its sparsity $\kappa$ is then the fraction of zeroes in $\theta'$. Practically, this is often achieved by multiplying the parameters with a mask $M \in \{0, 1\}^{|\theta|}$. The pair of networks also share some set of desirable properties, in which values of these properties are typically assumed to be more favourable, or at least as good, for a successfully compressed network. These properties minimally constitute some trade-off between a performance- and sparsity metric but can include many more. Modern pruning techniques are quite successful at finding these sparse solutions and prune over 95\% of the weights in a network, whilst leaving raw performance intact \citep{han2015deephuffman, li2016pruning, frankle2019lottery, evci2019rigging, you2019gate}. They often operate, by the \lq train-prune-finetune\rq\ pipeline \citep{ han2015learning, li2016pruning, liu2018rethinking}, which proceeds as follows; (a) take a pre-trained network and (b) prune some model-components, after which (c) some fine-tuning is performed \citep{han2015learning, li2016pruning}. Although these remarkable sparsity rates may give the impression that the problem is already solved, the following three reasons exhibit why they actually provide a skewed perspective:
\begin{enumerate}
  \item The \lq train-prune-finetune\rq\ routine still requires training a dense network first. On the upside, the weights are already close to their final value this way. However, pruning efforts have no contribution to the training phase. Some methods fight this by pruning during training \citep{louizos2017learning, frankle2019lottery, evci2019rigging, you2019gate}, which reduces the problem somewhat. Preferably, we would even prune \emph{before training}, so that we can exploit the sparse setting from the start. Unfortunately, this is a harder problem to solve. Recently, \citep{lee2019snip, wang2020picking} started tackling this problem, yet nevertheless only aimed at pruning individual weights, which brings us to the next point;
  
  \item Pruning methods frequently only target individual weights - i.e. unstructured pruning, which is more flexible and results in higher sparsity. However, down the line this leads to sparse weight matrices which require formatting such as \lq Compressed Sparse Row\rq\ (CSR) to be efficiently stored \citep{bulucc2009parallel, saad2003iterative} and only speeds up computation with dedicated libraries \citep{han2016eie, liu2018rethinking}. Fortunately, research concerning \emph{structured pruning} exists as well - i.e. pruning entire rows, columns or filters of a weight-tensor. Be that as it may, those methods are limited to pruning \emph{during or after training} in the literature, making the benefits available for inference only.
  
  \item Evaluation of these compressed sub-networks is usually performed along the dimensions of classifier-accuracy, computational cost and sparsity. However, this only validates a pruned solution shallowly. One can imagine, more aspects get affected in the process. For example, robustness to overfitting and adversarial attacks. Of course, some works do actually discuss these themes \citep{cosentino2019search, sehwag2020pruning, lee2019snip, arora2018stronger}. Even so, it is still uncommon to new methods in the literature.
\end{enumerate}

In the recently proposed \lq Lottery Ticket Hypothesis\rq\ \citep{frankle2018lottery}, it is established that networks contain sub-networks that are trainable in isolation, to the same performance as the parent-network - thereby, demonstrating that it would be possible to compress networks early-on. Subsequently, in SNIP \citep{lee2019snip}, it was first endeavoured to actually prune at initialisation, demonstrating notable  sparsity-rates using a simple \lq sensitivity criterion\rq\ - without resorting to difficult training schemes. Practically, they employ the gradients obtained from a single batch to rank weights in their sensitivity criterion, cut-off at a desired sparsity $\kappa$ and then train without further disturbance. Their criterion was thereafter criticised, due to unfaithful gradient propagation. To mitigate, orthogonal initialisation \citep{lee2019signal} and \lq GraSP\rq\ \citep{wang2020picking} were suggested. The latter replaces the gradient-magnitude product of SNIP with a hessian-gradient-magnitude product for a better gradient approximation \citep{wang2020picking}, yet compromises on simplicity. Our work builds upon the aforementioned research, examining structured- and unstructured pruning, before and during training. Our contributions are four-fold and can be summarised as follows: 
\begin{enumerate}
    \item We will introduce \emph{\lq\iterativesnip\rq}, a method that uses the same criterion as SNIP \citep{lee2019snip} but applies its sensitivity statistics iteratively in smaller steps, showing that this greatly improves performance and robustness in the high-sparsity regime, without compromising on simplicity.
    \item Then, we will demonstrate how this can be applied to structured pruning as well, yielding a method - \emph{\lq\iterativestructuredsnip\rq} - that doesn't require a pre-trained network or complex training schedule to obtain a competitive structured pruning algorithm. As such, to the best of our knowledge, we are introducing the first structured pruning method that operates \emph{before training}.
    \item Additionally, we will remark how the sensitivity signal is proportional to function elasticity \citep{sydsaeter1995mathematics, zelenyuk2013scale} and discuss the ranking and pruning of weights and nodes in a combined fashion.
    \item Finally, we will rigorously test and analyse our methods, on more than one dimension, and argue why pruning iteratively before or during training is effective at maintaining robustness. 
\end{enumerate}

\section{Related work}

\paragraph{Magnitude pruning}
The concept of pruning connections is not new and started with Hessian-based criterions; \lq Optimal Brain Damage\rq\ \citep{lecun1990optimal} and \lq Optimal Brain Surgeon\rq\ \citep{hassibi1993second}. Yet, it really began to gain traction in recent years, after \cite{han2015learning} demonstrated high compression rates. Consequently, research was performed into magnitude pruning, where they use the $\ell_1$-norm as pruning criterion \citep{han2015learning, han2015deephuffman, li2016pruning, frankle2018lottery}. Next, came the aforementioned Lottery Ticket Hypothesis, which demonstrated the existence of sparse sub-networks inside the randomly initialised networks, that could be trained to the same performance as the parent-network - coined \lq winning lottery tickets\rq\ \citep{frankle2018lottery}. This theory sparked a body of research examining the behaviour and obtainment of these winning tickets \citep{frankle2019lottery, yu2019playing, morcos2019one, desai2019evaluating, evci2019rigging, zhou2019deconstructing, frankle2019linear, malach2020proving}, as well as some criticism \citep{liu2018rethinking, gale2019state}. 

\paragraph{Bayesian approach}

Concurrently, using a Bayesian framework, some methods used uncertainty approximation to determine the redundancy of model components \citep{weigend1991generalization, tipping2001sparse, kingma2015variational, ullrich2017soft, molchanov2017variational, louizos2017learning}. Related to that, are methods employing a \lq sparsity inducing prior\rq, where using a distribution over model components as extra regularisation term in the loss pushes the model to a sparser solution \citep{carvalho2009handling, louizos2017learning, liu2017learning, yang2019deephoyer}. Notable, are relaxed $\ell_0$-regularisation \citep{louizos2017learning} where they approximate a differentiable $\ell_0$-norm and use it as additional penalty and DeepHoyer \citep{yang2019deephoyer}, where they instead use the ratio between $\ell_2$- and $\ell_1$-norm. Moreover, work was done into the usage of dropout as a means to pruning \citep{kingma2015variational, molchanov2017variational, gomez2018targeted}. 

\paragraph{Pruning sparsely initialised networks}
In addition, there are methods which train sparsely initialised networks. One way, is to initialise with a sparse distribution and then train with the periodical interchanging of which parameters are considered pruned and which are \lq grown back \rq\ \citep{bellec2017deep, dettmers2019sparse, evci2019rigging}. Hence, they don't train a final sparse setting from initialisation, which was later introduced in SNIP \citep{lee2019snip} and quickly followed by works that improve it \citep{lee2019signal, wang2020picking, hayou2020pruning, lee2020data}. The main criticism for SNIP has been, that its criterion is not scale-invariant for certain layers, such that it substantially influences the criterion's signal or sometimes even disconnects layers \citep{lee2019signal}. Solutions to solve these problems include using orthogonal initialisation \citep{lee2019signal}, GraSP \citep{wang2020picking} and are also the focus of this work.

\paragraph{Structured pruning}

Likewise to unstructured pruning, the beginning of structured pruning was derivative-based \citep{mozer1989skeletonization} and only really started gaining traction after the work of a magnitude based criterion \cite{li2016pruning}. Subsequently, other research focused on reconstructing output feature-maps under the strain of structured pruning, with the assumption that this implies maintaining performance \citep{he2017channel, luo2017thinet}. In other works, group-sparsity is used as sparsity inducing prior, with subsequent channel-pruning \citep{liu2017learning, wen2016learning, alvarez2016learning, lebedev2016fast, ye2018rethinking}. Most notable of them, is Network-slimming \cite{liu2017learning}, where they impose a $\ell_1$-penalty on scaling factors in batch-norm layers and thereby push neurons to deactivate during training. More recently, in GateDecorators \citep{you2019gate}, this idea is combined with a sensitivity-based criterion, alike was employed earlier in \cite{molchanov2016pruning, lee2019snip, mozer1989skeletonization} and now also in this work. Moreover, some works perform single-shot node pruning with pre-trained networks \citep{li2020ss, yu2019autoslim, li2019single}. Finally, there have been multiple papers where ADMM training is used, in which iteratively different layers get optimised in isolation \citep{liu2019autocompress, li2020ss}.

\section{Methodology}
\label{sec:methods}
\paragraph{Revisiting the sensitivity criterion}
We will adopt the sensitivity criterion that is applied in SNIP \citep{lee2019snip}, albeit practiced in a different fashion. It is worth mentioning that this criterion is not unique to SNIP \citep{mozer1989skeletonization, molchanov2016pruning, you2019gate, lee2019signal}. As such, multiple derivations coexist. In \cite{mozer1989skeletonization, lee2019snip}, authors commence by defining auxiliary gates $c$ over model parameters. They then initialise all $c = 1$ and don't update them anymore. At which point, the criterion $sc(\cdot)$ is defined as the derivative of the loss w.r.t. the gates:
\begin{align*}
  sc(\theta_{ij}) \midd{=} \frac{\partial L(\mathcal{D}\ |\ \theta \odot c)}{\partial c_{ij}} \Bigg |_{c=1}
\end{align*}
They then rank the model components by said criterion and cut-off at the desired sparsity $\kappa$. Later in \cite{lee2019signal}, it is recognised how for the unstructured case, via the chain rule, we obtain the derivative-magnitude product, rendering the auxiliary variables to be notational convention only:
\begin{align}
  \frac{\partial L(\theta \odot c)}{\partial c} \minquad = \minquad \frac{\partial L}{\partial (\theta \odot c)} \odot \frac{\partial (\theta \odot c)}{\partial c} \minquad = \minquad \frac{\partial L}{\partial \theta} \odot \theta
  \label{func:snip:saliency-updated}
\end{align}
 On the other hand, in \citep{molchanov2016pruning, you2019gate, wang2020picking}, it is concurrently shown how this is likewise found by a local Taylor-expansion. Presently, we will further demonstrate how this is also proportional to the \lq function elasticity\rq\ of the loss w.r.t. the parameters. This is a term which constitutes a mathematical way to express relative change in a function as a result of relative change in an input variable \citep{sydsaeter1995mathematics, zelenyuk2013scale}:
\begin{align}
  \varepsilon_y[x] \midd{=} \frac{\partial y(x) }{\partial x}  \frac{x}{y(x)}
  \label{func:elasticity-definition}
\end{align}
We now proceed to combine Formulas \ref{func:snip:saliency-updated} \& \ref{func:elasticity-definition} in this work, yielding the following proportionality:
\begin{align}
\frac{\partial L}{\partial \theta} \odot \theta \minquad
   \propto \minquad \frac{\partial L}{\partial \theta} \odot \frac{\theta}{L} = \varepsilon_L[\theta]
   \label{func:snip-equavalent-elasticity}
\end{align}
Incidentally, since the loss $L$ is the same for each model component $\theta_{ij}$ anyway, it would not matter for the purpose of ranking them from most to least important. In SNIP \citep{lee2019snip}, they instead divide by the sum of saliencies, which produces the same ranking too. However, by rather dividing by the loss, we gain some interpretability, as a weight is \lq very inelastic\rq\ when $\varepsilon_L[\theta_{ij}] \ll 1$, meaning the loss will not change after pruning that weight. When inspecting the distribution of elasticities, we always empirically observe that the majority of weights fall in this \lq very-inelastic\rq\ category and only a small percentage don't (see Supplementary Material \ref{sec:append:histograms}). This confirms the aforementioned finding of overparametrisation in literature \citep{denil2013predicting, dauphin2013big, ba2014deep, han2015learning} and also explains the high sparsity rates that state-of-the-art algorithms obtain. Moreover, we can now compare the sensitivity of weights and nodes on the same footing, meaning we can rank them together in one pruning event from our algorithm - \iterativesnip\ - and prune regardless of structure-type. This then yields a very flexible pruning technique, which we have explored but did not become the main focus of the paper due to limited applicational value (we report some explorations in Supplementary Material \ref{sec:append:cnip}). Additionally, it suggests something about a possible procedure of evaluating the importance of grouped model-components, as we can simply introduce a gate variable over any set of them and then employ elasticity to evaluate importance.

\paragraph{Iterative pruning as remedy to disconnection}

In the works of \cite{lee2019signal} and \cite{wang2020picking}, it is argued that the pitfall of vanilla SNIP \citep{lee2019snip} is that it stops information flow. Here, a new solution to this problem is suggested; iteratively applying the saliency criterion - with the corresponding method that is coined \emph{\lq\iterativesnip\rq}. The intuition goes as follows; certain medium-ranking model components, that are not that important to the loss initially, will be more important after the initial pruning-event. Specifically, we hypothesise the ranking may change, making specific model components more important in the sub-network than they were in the parent-network - i.e. the parent-network does not effectively dictate how information flows through the sub-network in isolation. By pruning more conservatively, in multiple rounds, we grant the model components another chance, therefore making the process more robust to the pruning algorithm and lowering the chance of disconnecting the network. Note that we still use the criterion from Formula \ref{func:snip-equavalent-elasticity} and that it can be applied both before or during training.

In practice, this materialises as dissecting the single pruning-event into a few pruning steps ($s$), that will be performed in one sequence. Its details are discussed in Algorithm \ref{alg:snipit:drawing}. As an easy rule of thumb, we start with pruning half the network's weights for the first event, since we observe that generally more than half are nearly completely inelastic [Supplementary Material \ref{sec:append:histograms}]. Thereafter we keep halving the remainder until the desired sparsity is reached with some $\epsilon$-proximity - after which we prune that final bit $\epsilon$ towards final sparsity $\kappa_{final}$. These specifics of the steps are not essential for the algorithm though. Accordingly, we do not experiment with different schedules. Instead we aim at a method that does not introduce too much complexity - i.e. it is the concept of iterative pruning that is important. That is not to say, the method cannot be improved by tuning this as well.

\begin{algorithm}
\caption{\small \iterativesnip}
\label{alg:snipit:drawing}
\begin{algorithmic}[1]
\scriptsize{
\Procedure{\iterativesnip}{ $f$, $\kappa_\textit{final}$, $\tau$, $s$ }   \Comment{i.e. network $f$, sparsity $\kappa$, interval $\tau$, steps $s$}
\State $\theta_0 \gets \text{choose what to prune from  $f_\text{weights}$ $\cup$\ $f_\text{nodes}$}$  \Comment{structured, unstructured or combined}
\For{i $\in [0, s]$ }     \Comment{we default $s=5$}
\State train $\tau$ epochs \Comment{$\tau=0$ for \emph{before-training}, $\tau>0$ for \emph{during training}}
\State  $\kappa_i \gets \kappa_\textit{final}-(\kappa_\textit{final}-\frac{1}{2})\cdot\frac{1}{2}^i$  \Comment{rule of thumb: 50\% first, then half of the remainder each time after}
\State $\theta_i \gets$ rank $\theta$ with $\epsilon_L[\theta_i]$ and prune to $\kappa_i$ \Comment{See Formulas \ref{func:snip-equavalent-elasticity} \& \ref{func:structured:criterion}}
\EndFor
\State $\theta_\textit{final}' \gets$ rank $\theta$ with $\epsilon_L[\theta_s]$ and prune to $\kappa_\textit{final}$ 
\State \textbf{start training} $f(\theta_\textit{final}')$
\EndProcedure}
\end{algorithmic}
\end{algorithm}
Note, that just like SNIP, we use one batch for each gradient evaluation in $\varepsilon_L[\theta]$. Granted, with more iterations, we will thus cumulatively see more data. Yet, this is hardly computationally expensive and each batch just constitutes an independent unbiased estimator after all \citep{bottou2018optimization, robbins1951stochastic}. In fact, in the SNIP paper it is argued that if the algorithm is performed with a batch of sufficient size, then the multiplication of that size with a constant $d>1$, does not constitute any noticeable difference \citep{lee2019snip}. In all experiments we performed, algorithms are always provided with an accumulated batch of $2560$ samples, which we deem large enough to minimise its influence sufficiently.

\paragraph{Structured pruning before training}

Additionally, \iterativesnip\ is easily supplemented with a structural equivalent. Generally, structured pruning methods substantially reduce computational time and feature maps. Combined with the benefit of pruning \emph{before training}, we unlock a potential to drastically reduce the total required compute. At the time of writing, to the best of our knowledge, no methods endeavoured structured pruning before training yet. A such, it is introduced here. Where unstructured \iterativesnip\ uses weight-elasticity $\varepsilon_L[\theta]$ as a criterion, we now employ node-elasticity. If $h$ is the activation function of a node, we thus straightforwardly produce $\varepsilon_L[h]$. However, since $h(\cdot)$ itself is a function and not actually a model parameter, we adopt the same notational convention applied in Skeletonization \citep{mozer1989skeletonization} and define auxiliary gates $c_j = 1$ over each node's input, thus substituting the activations $h(\cdot)$ for each layer $\ell \in P$ (see Formula \ref{func:structured:criterion}), subsequently leading to criterion $sc(h_{i}^{(\ell)})$. 
\begin{align}
    \begin{split}
  h^{(\ell)}\big(\Theta^{(\ell)} \cdot \vec{x}^{(\ell)} + \vec{b}^{(\ell)}\big) \midd{=} h^{(\ell)}\big( diag(\vec{c}^{(\ell)}) \cdot ( \Theta^{(\ell)} \cdot \vec{x}^{(\ell)} + \vec{b}^{(\ell)})\big)\ \Bigg |_{\vec{c}^{(\ell)}=\vec{1}}   \\
   \varepsilon_L[h_{i}^{(\ell)}] \minquad \approx \minquad sc(h_{i}^{(\ell)}) \midd{=} \frac{\partial L(\mathcal{D}\ |\ \{\Theta^{(\ell)}\}_{\ell=1}^{P})}{\partial c_{i}^{(\ell)}} \cdot \frac{1}{L}
   \end{split}
   \label{func:structured:criterion}
\end{align} 
Thereafter, the same procedure as unstructured \iterativesnip\ is followed. Additional measures are taken to not prune away entire layers, as well as not pruning away the input and output nodes.

\section{Experimental setup}

\paragraph{Datasets and networks}
Experiments are evaluated with a number of different networks and datasets. The datasets used are publicly available and consist of three common benchmarks; MNIST \citep{lecun1998gradient}, CIFAR10 \citep{krizhevsky2009learning} and an ImageNet-subset; \lq Imagenette\rq\ \citep{deng2009imagenet, imagenette}. Networks that are used are LeNet5 \citep{lecun1998gradient}, MLP5 [\ref{sec:append:networks}], Conv6 \citep{frankle2018lottery}, ResNet18 \citep{he2016deep}, AlexNet \citep{krizhevsky2012imagenet} and VGG16 \citep{simonyan2014very}. Implementation and more network, data and augmentation specifics can be found in the Supplementary Material.

\paragraph{Initialisation}

As mentioned earlier, \cite{lee2019signal} showed the significance of initialisation in SNIP. Specifically, it is argued that orthogonal initialisation - although needed to be scaled by the activation function - is the best way to keep the magnitude of the signal comparable over different layers. For this reason, we adopt this initialisation in our experiments - unless a baseline has its own preferred initialisation. Note that this concerns initialisation before the network enters the pruning algorithm.

\paragraph{Baselines} As baselines we use SNIP \citep{lee2019snip}, GraSP \citep{wang2020picking}, HoyerSquare \citep{yang2019deephoyer} and global-IMP \citep{frankle2018lottery, frankle2019lottery} for unstructured pruning, whereas $\ell_0$-regularisation \citep{louizos2017learning}, Group-HS \citep{yang2019deephoyer}, EfficientConvNets \citep{li2016pruning} and GateDecorators \citep{you2019gate} for structured pruning. For all baselines hyperparameter tuning is performed - see Supplementary Material \ref{sec:append:tuning} for the tuning details - and implementation is sourced from authors' publications as much as possible, depending on availability and possibility of integration. 

\paragraph{Optimisation}

For all experiments, the ADAM-optimiser \citep{kingma2014adam} was used with a learning rate of $2 \cdot 10^{-3}$. No additional learning rate schedules were employed to avoid over-complication. Further, a weight decay of $5 \cdot 10^{-5}$ was used in all experiments. The batch-size in all CIFAR-10 and MNIST experiments was $512$, whereas in Imagenette it is always $128$. Finally, gradient clipping at a magnitude of $10.0$ was applied as well. All methods have been given similar epoch budget, with an exception for $\ell_0$-regularisation \citep{louizos2017learning} and GateDecorators \citep{you2019gate}, which required twice that of the rest to converge.

\paragraph{Evaluation}
Evaluation is performed on sparsity, performance, connectivity and adversarial robustness. Additionally, in the structured domain; time, memory footprint and FLOPS are taken into account. The last is estimated with an implementation adopted from \cite{liu2018rethinking}. Furthermore, where applicable a confidence bound is displayed, which is calculated by: \ $1.96 \cdot \frac{\sigma}{\sqrt{n}}$, where $n$ is the sample-size with a median value of $5$, $\sigma$ the standard deviation and $1.96$ corresponds to a confidence interval of $95\%$ - assuming the points are normally distributed. We evaluate in the structured and unstructured setting, as well as before and during training. To clearly distinguish \iterativesnip's structured counterpart, we will refer to it as \lq\iterativestructuredsnip\rq\ from here forward. For metric details, see Supplementary Material \ref{sec:append:metrics}.

\section{Results \& analyses}
\begin{figure}
  \centering
    \captionsetup[subfigure]{justification=centering}

  \subfloat[\textit{\small AlexNet}]{
  \includegraphics[width=0.48\linewidth]{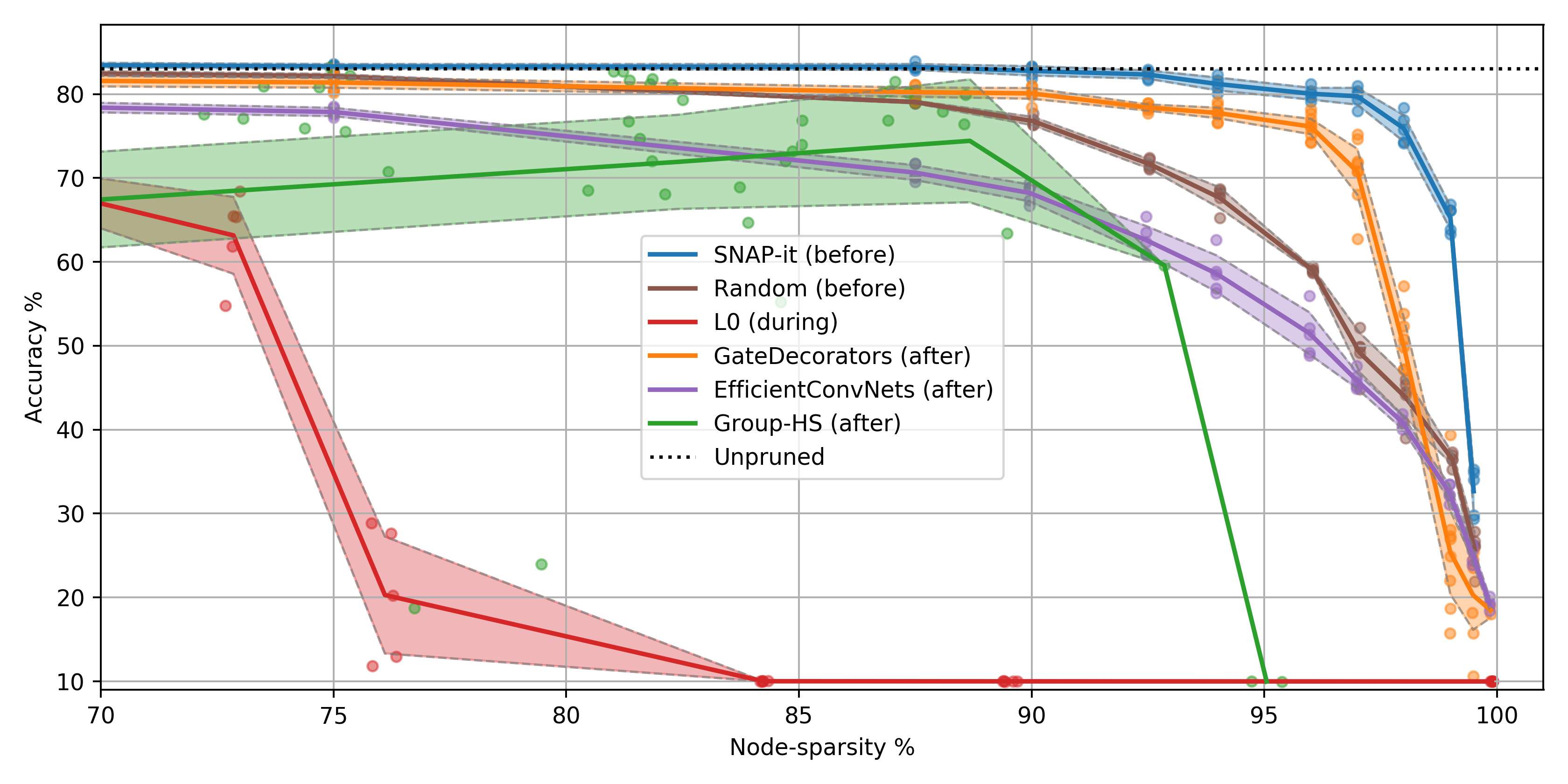}
  \label{fig:node_spars:alex}
}
  \subfloat[\textit{\small VGG16}]{
  \includegraphics[width=0.48\linewidth]{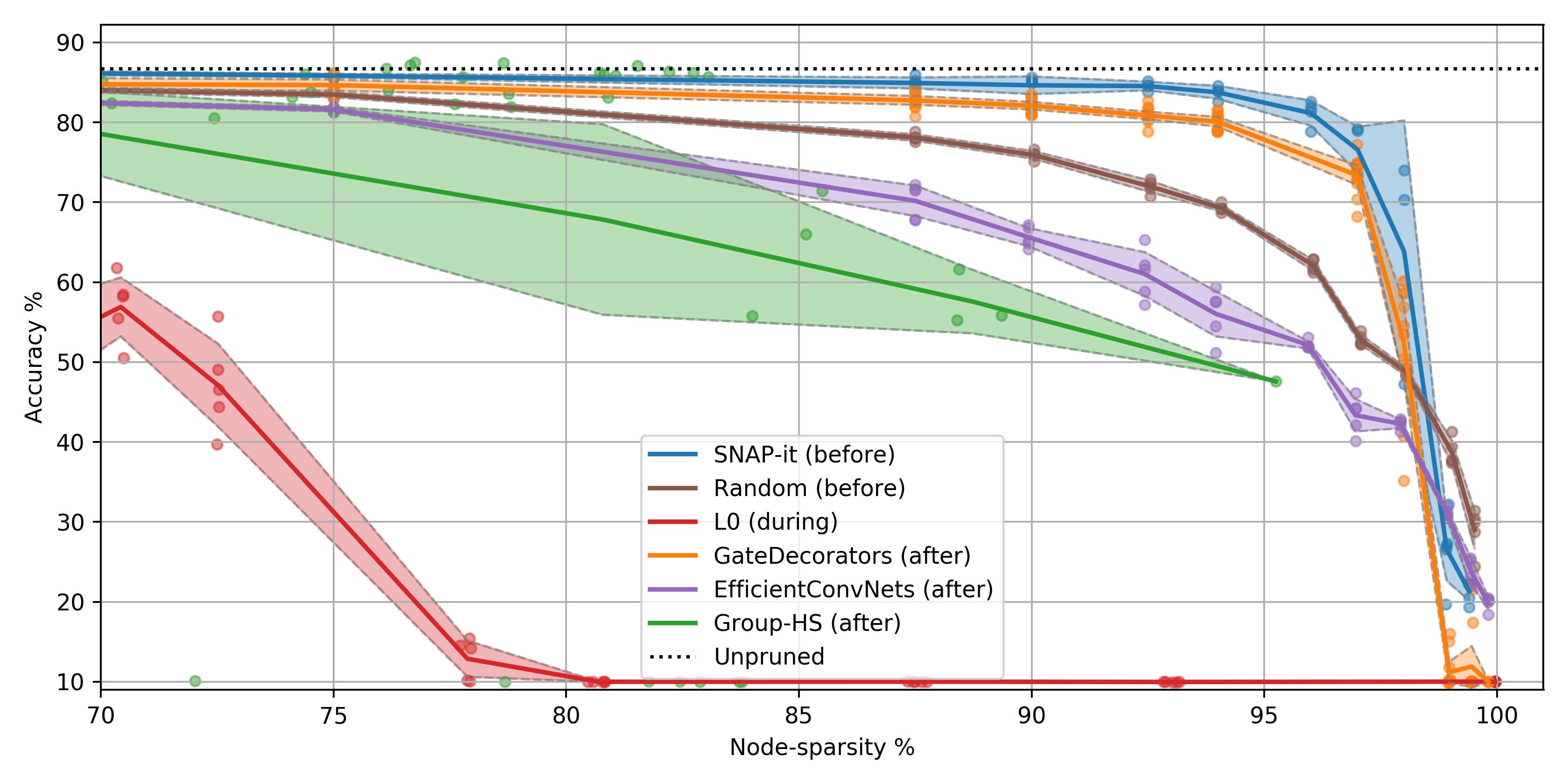}
  \label{fig:node_spars:vgg16}
}
  \caption{\textit{Structured sparsity-performance trade-off for CIFAR10 with confidence bound.}}
  \label{fig:node_spars}
\end{figure}

\paragraph{Structured pruning before training}

We first evaluate \iterativestructuredsnip\ in the domain with the most potential: structured pruning before training. We have assessed its during-training domain also, yet found it operates on-par with before training, so focused on the latter. It is illustrated in Figure \ref{fig:node_spars}, that \iterativestructuredsnip\ outperforms baselines in terms of sparsity and performance, even though it prunes \emph{before training}. Interestingly, Random forms a strong baseline. We speculate this is due to the nature of optimisation, where pruning \emph{before training} gives the network enough chance to adapt, which again indicates how these networks are overparametrised. Additionally, we find that $\ell_0$-regularisation \citep{louizos2017learning} performs notably poor. To elaborate, we empirically observe that it performs better for the specific hyperparameter settings and architectures that were reported in their paper - as is portrayed in Figure \ref{fig:l0_works_sometimes} in Supplementary Material \ref{sec:append:results:cifar}. However, the method doesn't easily extend to new settings, even with the considerable tuning we performed, which we find defeats the purpose of an easy pruning method to begin with. Next, we witness that for any $\lambda$, Group-HS \citep{yang2019deephoyer} is subject to high variance in both performance and sparsity, and collapses occasionally, which decreases its mean in Figure \ref{fig:node_spars}. 

\begin{table}

\centering
\caption{\textit{Structured results on Imagenette, showing optimal sparsity performance trade-off, as well as training-time, training-FLOPS and RAM-footprint for \iterativestructuredsnip\ - compared to baselines. Points are picked using the highest harmonic mean between sparsity and accuracy.}}
\vspace{10pt}
\label{tab:node:imagenette}
\scalebox{0.7}{
\begin{tabular}{@{}c@{\hskip 0.75cm}c@{\hskip 1.25cm}c@{\hskip 0.2cm}c@{\hskip 1.25cm}c@{\hskip 0.1cm}c@{\hskip 1.25cm}c@{\hskip 0.1cm}c@{\hskip 1.25cm}c@{\hskip 0.1cm}c@{}}
\toprule
\multicolumn{10}{c}{\textbf{VGG16}}  \\ 

          Method              & Accuracy &  \multicolumn{2}{@{}c@{\hskip 1.25cm}}{Sparsity} & \multicolumn{2}{@{}c@{ \hskip1.25cm}}{FLOPS} 					& \multicolumn{2}{@{}c@{\hskip 1.25cm}}{Time} 					& \multicolumn{2}{@{}c@{}}{Storage} \\
          					  &  	 & \scriptsize Weight & \scriptsize Node    & \scriptsize Inference & \scriptsize Train  	& \scriptsize Inference & \scriptsize Train & \scriptsize Disk & \scriptsize RAM  	\\  	

          \midrule

Baseline                       & 87\%            &&         &&    &&  &&   \\
\multicolumn{10}{c}{ \small Prunes after training}\\ 
Effic.ConvN.				   & 80\% $\pm$ 2.5       &            92\% & 88\%             &          49\x & 1 \x           &          \bb{10}\x & 1 \x      &        6e1\x & 1 \x        \\
GateDecorators                 & 83\% $\pm$ 1.8       &            96\% & \bb{93\%}        &          47\x & 1 \x           &          5 \x & 1 \x           &        1e2\x & 10\x        \\
Group-HS					   & 84\% $\pm$ 2.1       &            82\% & 68\%             &          3 \x & 1 \x           &          2 \x & 1 \x           &        1e1\x & 1 \x        \\
\multicolumn{10}{c}{ \small Prunes during training}\\ 
L0                             & 28\% $\pm$ 6.2       &            55\% & 58\%             &          2 \x & 1 \x           &          1 \x & 1 \x           &        1e0\x & 1 \x        \\
 \multicolumn{10}{c}{ \small Prunes before training}\\ 
Random						   & 82\% $\pm$ 0.4       &            93\% & 90\%             &     \bb{76}\x & \bb{76}\x       &          9 \x & \bb{7} \x       &        1e2\x & 7 \x        \\
SNAP-it \footnotesize{(ours)}  & \bb{85\% $\pm$ 1.4}  &            \bb{97\%} & \bb{93\%}   &          15\x & 15\x            &          5 \x & 4 \x           &   \bb{1e3}\x & \bb{20}\x   \\
\toprule
 \multicolumn{10}{c}{\textbf{AlexNet}}  \\ 
\midrule
Baseline                       & 86\%		         &&         &&    &&  &&    \\
\multicolumn{10}{c}{ \small Prunes after training} \\
Effic.ConvN.				   & 78\% $\pm$ 1.9       &            92\% & 90\%             &          \bb{61}\x & 1 \x      &          \bb{13}\x & 1 \x      &        4e1\x & 1 \x        \\
GateDecorators                 & 84\% $\pm$ 2.5       &            93\% & 90\%             &          14\x & 1 \x           &          5 \x & 1 \x           &        2e1\x & 4 \x        \\
Group-HS					   & 84\% $\pm$ 1.1       &            85\% & 78\%             &          2 \x & 1 \x           &          2 \x & 1 \x           &        8e0\x & 1 \x        \\
\multicolumn{10}{c}{ \small Prunes during training}\\ 
L0                             & 31\% $\pm$ 3.9       &            54\% & 58\%             &          2 \x & 1 \x           &          1 \x & 1 \x           &        6e0\x & 1 \x        \\
 \multicolumn{10}{c}{ \small Prunes before training}\\ 
Random						   & 83\% $\pm$ 0.8       &            90\% & 88\%              &          45\x & \bb{45}\x      &          10\x & \bb{6} \x      &        2e1\x & 4 \x        \\
SNAP-it \footnotesize{(ours)}  & \bb{85\% $\pm$ 0.9}  &       \bb{98\%} & \bb{96\%}         &          16\x & 16\x           &          6 \x & 5 \x           &        \bb{8e2}\x & \bb{16}\x    
\end{tabular}

}
\end{table}
Moreover, we test \iterativestructuredsnip\ on the Imagenette dataset, which is native to larger pictures. Results are found in Table \ref{tab:node:imagenette}. Again, we observe that \iterativestructuredsnip\ reaches the best sparsity-performance trade-off and random forms a strong baseline. Granted, EfficientConvNets \citep{li2016pruning} and GateDecorators \citep{you2019gate} still reach the best inference-FLOPS and inference-time reduction. However, when looking at training-FLOPS and training-time, \iterativestructuredsnip\ has a significant advantage. Which, depending on the frequency a network needs to be retrained in practice, can make an extraordinary impact on the computational effort. 

Equally promising, is the RAM-footprint reduction, which is also found in Table \ref{tab:node:imagenette} as median reduction \emph{during training}. However, when one is pruning \emph{before training}, as \iterativestructuredsnip\ does, we obtain this benefit immediately. Therefore, we can drastically lower the amount of feature maps we store, whereas the baselines require keeping all nodes with their corresponding feature maps in memory during training. Therefore, we gain the opportunity to proportionally increase the batch-size, which in turn allows us to increase the learning rate \citep{goyal2017accurate}. This observation is also reflected in Figure \ref{fig:ram_accuracy} in Supplementary Material \ref{sec:append:memory}, where we see the trade-off between accuracy after training and RAM-footprint at the start of training. Note however, that RAM and time measurements can be influenced by processes playing at test-machines and should therefore be taken with some reservation.

\paragraph{Unstructured \iterativesnip}

\begin{figure}
  \centering
    \captionsetup[subfigure]{justification=centering}

  \subfloat[\textit{\small LeNet5}]{
  \includegraphics[width=0.3\linewidth]{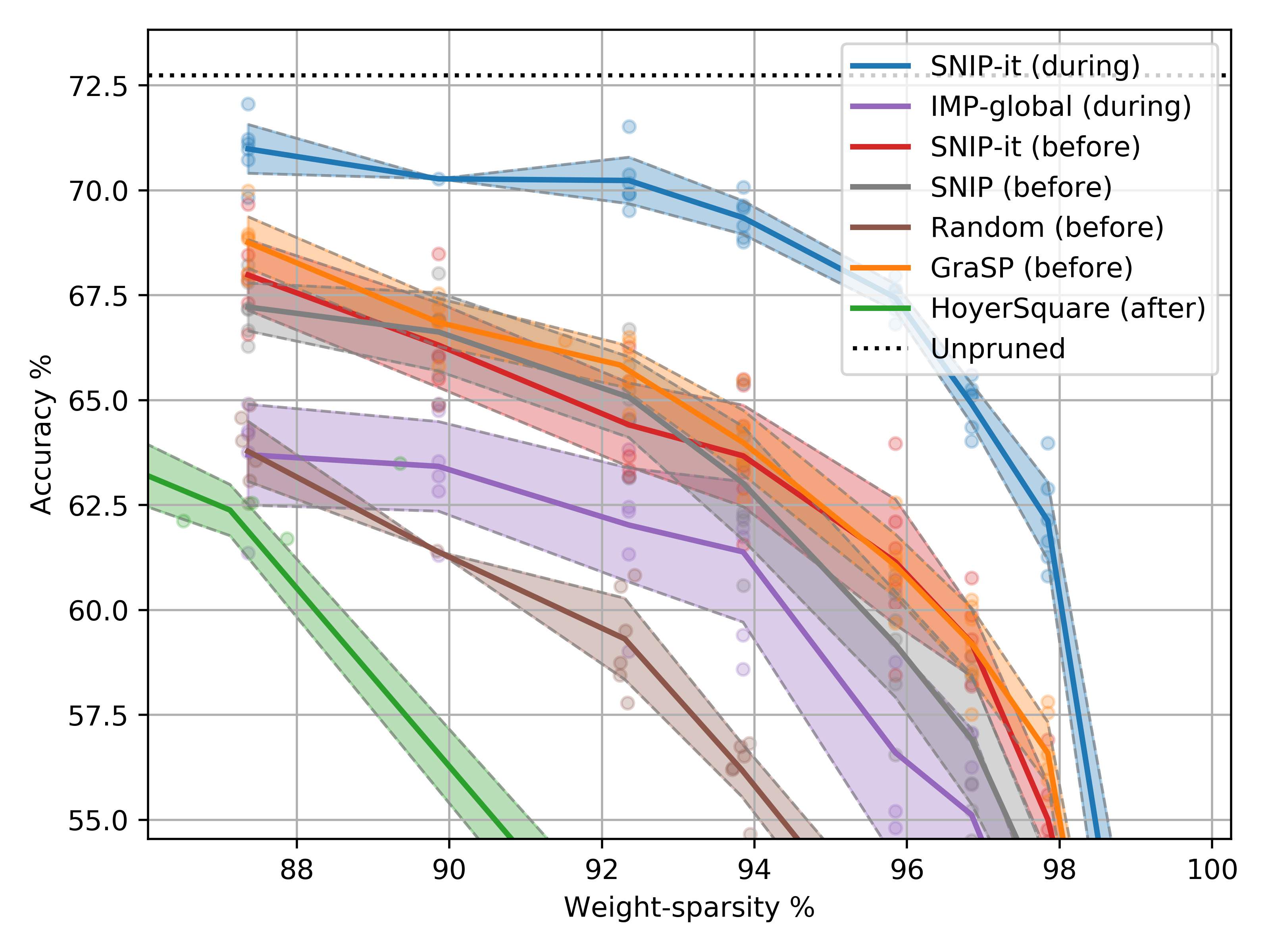} \label{fig:spars:lenet5}
}
  \subfloat[\textit{\small MLP5}]{
  \includegraphics[width=0.3\linewidth]{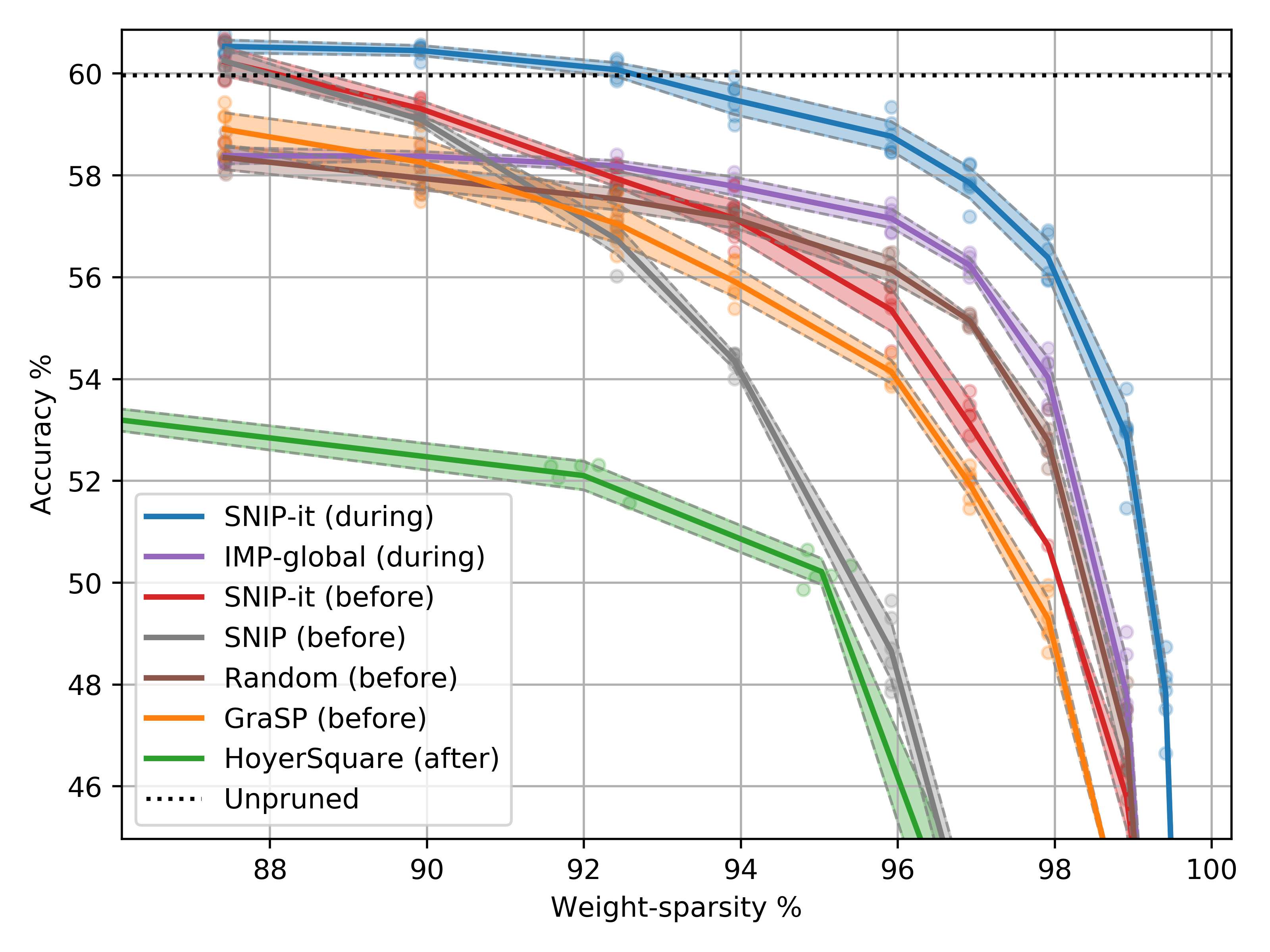} \label{fig:spars:mlp5}
}
  \subfloat[\textit{\small ResNet18}]{
  \includegraphics[width=0.3\linewidth]{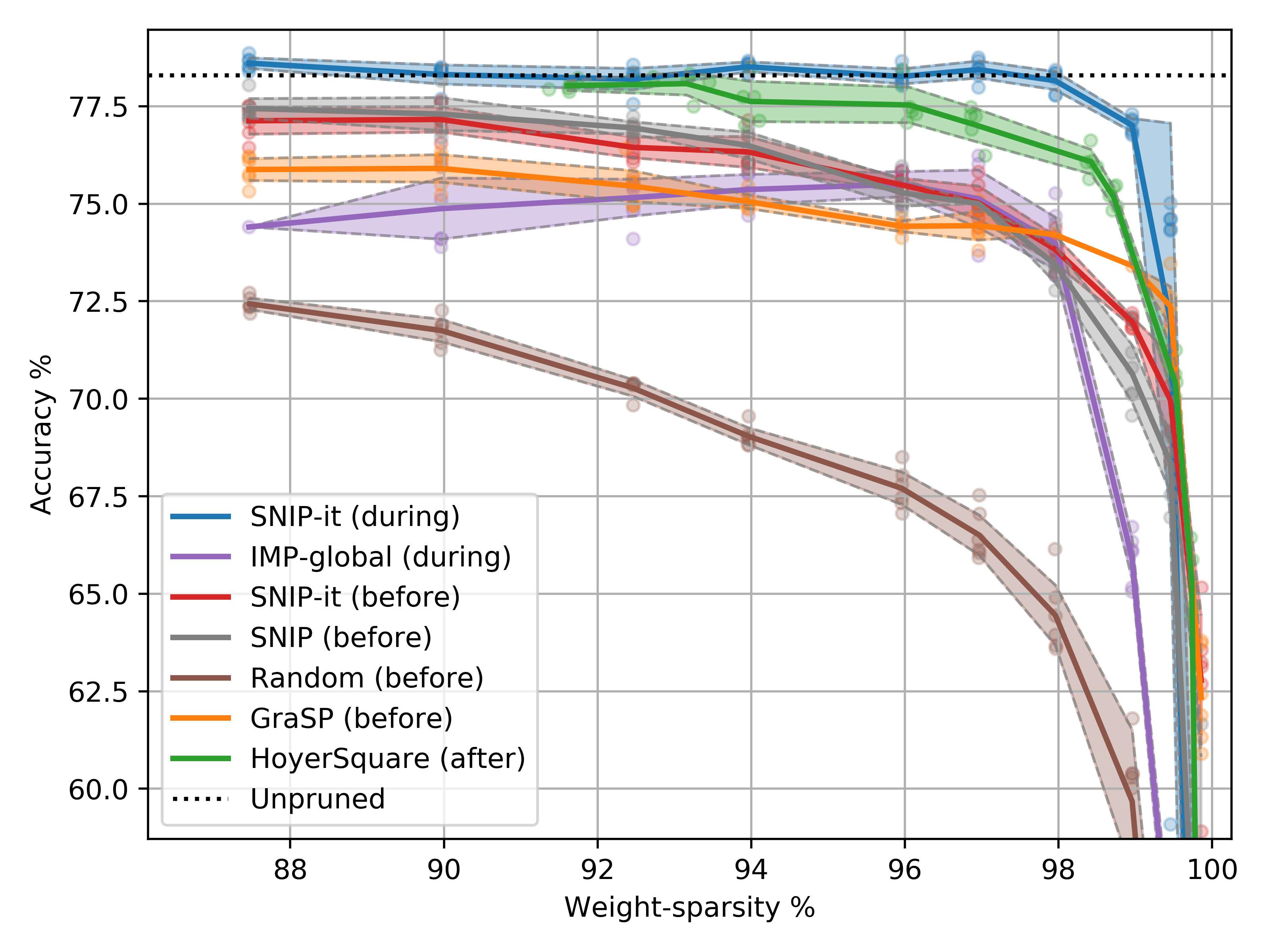} \label{fig:spars:ResNet18}
}
  \caption{\textit{Unstructured sparsity-performance trade-off for CIFAR10 with confidence bound.}}
  \label{fig:spars}
\end{figure}

In addition, we evaluate \iterativesnip\ in the unstructured domain, both before- and during-training. For this, we use $\tau=0$ and $\tau=4$ respectively (see Algorithm \ref{alg:snipit:drawing}), where the latter was based on the hyperparameter tuning that was done for IMP \citep{frankle2019lottery}, which we re-used to make \iterativesnip\ apply the same schedule - see Supplementary Material \ref{sec:append:tuning} for details. In Table \ref{tab:sparsity}, results for various network-dataset combinations are displayed, with accuracy and sparsity as percentages. Adjacently, is the harmonic mean between the two (HM), which we designate as a proxy for the trade-off between them - just like the F1-score for information retrieval. Furthermore, sparsity-accuracy curves are shown in Figure \ref{fig:spars}. What follows from both, is that \iterativesnip\ (during training) is significantly Pareto-optimal compared to all baselines on almost all setups. Moreover, \iterativesnip\ (before training) significantly surpasses SNIP \citep{lee2019snip} and performs at least as good as GraSP \citep{wang2020picking}, now and again outperforming it. Interestingly, in the case of an MLP5 random performs better than expected, which emphasizes the importance to test against a random baseline on different types of networks.

\begin{figure}
  \centering
  \captionsetup[subfigure]{justification=centering, labelformat=empty}
\subfloat[\textit{\footnotesize SNIP \scriptsize \citep{lee2019snip}}]{
  \includegraphics[width=0.145\linewidth]{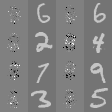} \label{fig:saliencies:snip}
}
\hfill
\subfloat[\textit{\footnotesize GraSP \scriptsize \citep{wang2020picking}}]{
  \includegraphics[width=0.145\linewidth]{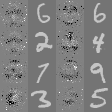} \label{fig:saliencies:grasp}
}
\hfill
\subfloat[\textit{\footnotesize \iterativesnip\ \scriptsize (before)}]{
  \includegraphics[width=0.145\linewidth]{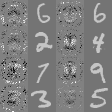} \label{fig:saliencies:snipit}
}
\hfill
\subfloat[\textit{\footnotesize IMP-global \scriptsize \citep{frankle2019lottery}}]{
  \includegraphics[width=0.145\linewidth]{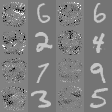} \label{fig:saliencies:IMP}
}
\hfill
\subfloat[\textit{\footnotesize HoyerSquare \scriptsize \citep{yang2019deephoyer}}]{
  \includegraphics[width=0.145\linewidth]{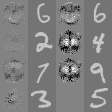} \label{fig:saliencies:hoyer}
}
\hfill
\subfloat[\textit{\footnotesize \iterativesnip\ \scriptsize (during)}]{
  \includegraphics[width=0.145\linewidth]{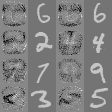} \label{fig:saliencies:snipit-during}
}
  \caption{\textit{Saliency maps \citep{simonyan2013deep} (left) w.r.t. inputs (right), in MLP5 trained on MNIST at $\kappa=0.99$.}}
  \label{fig:saliencies}
\end{figure}

\begin{figure}
  \centering
  \includegraphics[width=0.95\linewidth,trim={0.1cm 0.35cm 0.1cm 0.35cm},clip]{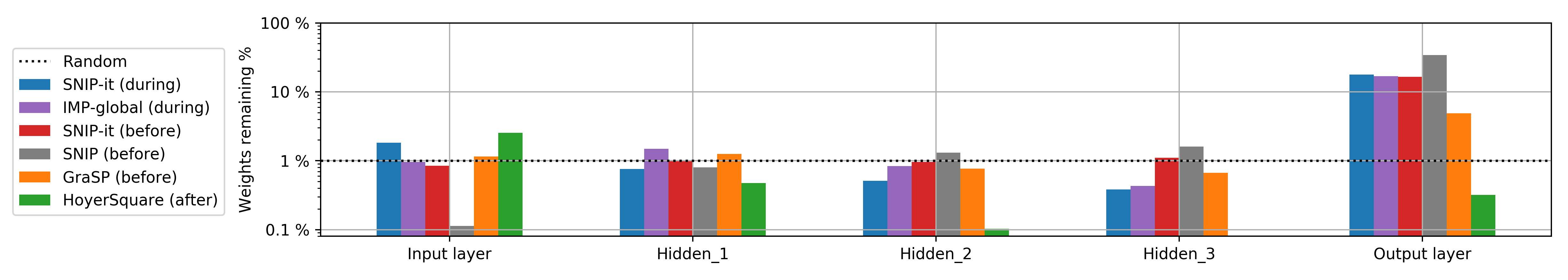}
  \caption{\textit{Histogram of remaining weights in MLP5 trained on MNIST at $\kappa=0.99$.}}
  \label{fig:weights-remaining}
  \vspace{-5pt}
\end{figure}

\begin{table}
\centering
\caption{\textit{Unstructured sparsity-performance trade-offs, showing \iterativesnip\ (during training) outperforms baselines. Points are picked using the highest harmonic mean (HM) between sparsity and accuracy.}}
\vspace{10pt}
\label{tab:sparsity}
\scalebox{0.7}{ 
\begin{tabular}{@{}r@{\hskip 0.7cm}c@{\hskip 0.17cm}c@{\hskip 0.17cm}c@{\hskip 1.051cm}c@{\hskip 0.17cm}c@{\hskip 0.17cm}c@{\hskip 1.051cm}c@{\hskip 0.17cm}c@{\hskip 0.17cm}c@{\hskip 1.051cm}c@{\hskip 0.17cm}c@{\hskip 0.17cm}c@{}}
\toprule

& \multicolumn{12}{c}{\textbf{CIFAR10}}  \\ 
& \multicolumn{3}{c}{LeNet5} & \multicolumn{3}{c}{MLP5} & \multicolumn{3}{c}{Conv6} & \multicolumn{3}{c}{ResNet18} \\
								&\small Accuracy  			 &\small Sparsity 	& \small HM 		& \small Accuracy  				& \small Sparsity    & \small HM  	  & \small Accuracy  			   & \small Sparsity  & \small HM  	   & \small Accuracy  				& \small Sparsity  & \small HM  \\
\midrule

Baseline 						&  73\%  		         &			&		    & 60\% 					&			&       & 87\%  			   &         &		   		&  78\%  	   			&         &   \\
& \multicolumn{12}{c}{\small Prunes after training}   \\ 
HoyerSquare						&  66\% $\pm$ 1.4  &     82\%     &     73     &  52\% $\pm$ 0.3  &     92\%     &     66     &  82\% $\pm$ 0.6  &     89\%     &     85     &  77\% $\pm$ 0.5  &   97\%   & 86   \\
& \multicolumn{12}{c}{\small Prunes during training}   \\ 
IMP-global         				&  63\% $\pm$ 1.3  &     90\%     &     74     &  57\% $\pm$ 0.2  &  \bb{96\%}   &     72     &  84\% $\pm$ 0.8  &     92\%     &     88     &  75\% $\pm$ 0.9  &   97\%   & 85   \\
SNIP-it \footnotesize{(ours)}   &\bb{70\% $\pm$ 0.7}& \bb{92\%}   &   \bb{80}  &\bb{59\% $\pm$ 0.4}&  \bb{96\%}  &  \bb{73}   &\bb{85\% $\pm$ 0.4}&  \bb{97\%}  & \bb{91}    &\bb{78\% $\pm$ 0.3}&\bb{98\%}&\bb{87}\\
& \multicolumn{12}{c}{\small Prunes before training}   \\ 
GraSP              				&  69\% $\pm$ 0.8  &     87\%     &     77     &  58\% $\pm$ 0.6  &     90\%     &     71     &  82\% $\pm$ 0.5  &     96\%     &     88     &  74\% $\pm$ 0.7  &\bb{98\%} & 84   \\
Random             				&  64\% $\pm$ 0.9  &     87\%     &     74     &  57\% $\pm$ 0.2  &     94\%     &     71     &  82\% $\pm$ 0.4  &     92\%     &     87     &  70\% $\pm$ 0.2  &   92\%   & 80   \\
SNIP      						&  67\% $\pm$ 1.2  &     90\%     &     77     &\bb{59\% $\pm$ 0.3}&    87\%     &     71     &  81\% $\pm$ 0.9  &     94\%     &     87     &  75\% $\pm$ 1.0  &   97\%   & 85   \\
SNIP-it \footnotesize{(ours)}   &  68\% $\pm$ 1.1  &     87\%     &     76     &\bb{59\% $\pm$ 0.2}&    90\%     &     71     &  83\% $\pm$ 0.5  &     94\%     &     88     &  75\% $\pm$ 0.5  &   97\%   & 85   \\
\toprule 

& \multicolumn{12}{c}{\textbf{Imagenette}} \\
& \multicolumn{3}{c}{LeNet5} & \multicolumn{3}{c}{MLP5} & \multicolumn{3}{c}{Conv6} & \multicolumn{3}{c}{ResNet18} \\
							 &\small Accuracy  			 &\small Sparsity 	& \small HM 		& \small Accuracy  				& \small Sparsity    & \small HM  	  & \small Accuracy  			   & \small Sparsity  & \small HM  	   & \small Accuracy  				& \small Sparsity  & \small HM  \\
\midrule

Baseline  					 & 83\%  				  & 		&		  	 & 53\%  			   		&  		 & 		     & 87\%  				& 		  &		    	 & 80\%  		 & 			 & 				\\
& \multicolumn{12}{c}{\small Prunes after training}\\
HoyerSquare					 &\bb{83\% $\pm$ 1.8}&    92\%     &     87     &  53\% $\pm$ 0.9  &     90\%     &     67     &  83\% $\pm$ 1.2  &     97\%     &     89     &  80\% $\pm$ 0.7  &   95\%     &     87     \\
& \multicolumn{12}{c}{ \small Prunes during training}\\
IMP-global         			 &  70\% $\pm$ 1.7  &   \bb{98\%}  &     82     &  54\% $\pm$ 0.7  &     97\%     &     69     &  77\% $\pm$ 3.0  &     97\%     &     86     &  77\% $\pm$ 1.4  &   98\%     &     86     \\
SNIP-it \footnotesize{(ours)}&\bb{83\% $\pm$ 2.0}&  \bb{98\%}  &  \bb{89}   &  56\% $\pm$ 1.1  &  \bb{99\%}   &  \bb{72}   &\bb{87\% $\pm$ 1.1}&  \bb{99\%}  &  \bb{93}   &\bb{82\% $\pm$ 0.8}&\bb{99\%}  &  \bb{90}   \\
& \multicolumn{12}{c}{ \small Prunes before training}\\ 
GraSP              			 &  81\% $\pm$ 0.8  &   \bb{98\%}  & \bb{89}    &  57\% $\pm$ 0.9  &     94\%     &     71     &\bb{87\% $\pm$ 1.0}&    98\%     &     92     &  79\% $\pm$ 0.9  & \bb{99\%}  &     88     \\
Random             			 &  78\% $\pm$ 0.7  &     90\%     &     84     &  54\% $\pm$ 0.5  &     96\%     &     69     &  85\% $\pm$ 0.8  &     96\%     &     90     &  79\% $\pm$ 0.7  &   96\%     &     87     \\
SNIP    					 &  82\% $\pm$ 1.1  &   \bb{98\%}  & \bb{89}    &\bb{58\% $\pm$ 1.0}&     97\%    &     73     &\bb{87\% $\pm$ 1.1}&    98\%     &     92     &  79\% $\pm$ 1.0  &   98\%     &     87     \\
SNIP-it \footnotesize{(ours)}&  81\% $\pm$ 1.5  &     97\%     &     88     &\bb{58\% $\pm$ 0.9}&     96\%    &  \bb{72}   &  86\% $\pm$ 0.7  &     98\%     &     92     &  80\% $\pm$ 0.8  &   98\%     &     88     \\ 
 
\bottomrule

\end{tabular}
}
\end{table}
\paragraph{Exploration on robustness} 
If we zoom in on resulting architectures, findings correspond to our hypothesis, which said the  ranking changes with each pruning event - introduced in Section \ref{sec:methods}. We display some saliency maps \citep{simonyan2013deep} with their respective inputs in Figure \ref{fig:saliencies}. Note that for the case of HoyerSquare \citep{yang2019deephoyer} we had to interpolate to 99\% sparsity from the closest point, due to its sparsity being harder to control. We observe that SNIP \citep{lee2019snip} overfits on only a few pixels, compared to GraSP \citep{wang2020picking} and \iterativesnip\ (before training). Even better, are the saliency maps of HoyerSquare and \iterativesnip\ \emph{(during training)}, which hardly indicate any overfitting. These findings are again reflected in Figure \ref{fig:weights-remaining}, where we see that the different pruning methods favor different layers' weights as best candidates for pruning. As mentioned before, a common criticism against SNIP is its bias towards earlier layers \citep{lee2019signal, wang2020picking}, which manifests itself again here. HoyerSquare \citep{yang2019deephoyer} on the other hand, shows a bias towards pruning the output layer. If we assume the outermost two layers are the most important to maintain, we would expect to see a downward parabola-like shape in this figure, as that would prune heavier in the layers in the middle. This is the case for \iterativesnip, in contrast to for instance; SNIP \citep{lee2019snip}. Thus, in accordance with our hypothesis, by applying the sensitivity criterion in iterations, weights in the earlier layers seem to be more important to intermediate pruning events, rather than SNIP's one-shot pruning of the full percentage, indicating the ranking changes between these intermediate events.

\begin{figure}[H]
  \centering
  \captionsetup[subfigure]{justification=centering}
  \subfloat[\textit{Success prob. vs. sparsity at $\ell_2$-distance$\ \leq 0.5$.}]{
  \includegraphics[width=0.47\linewidth]{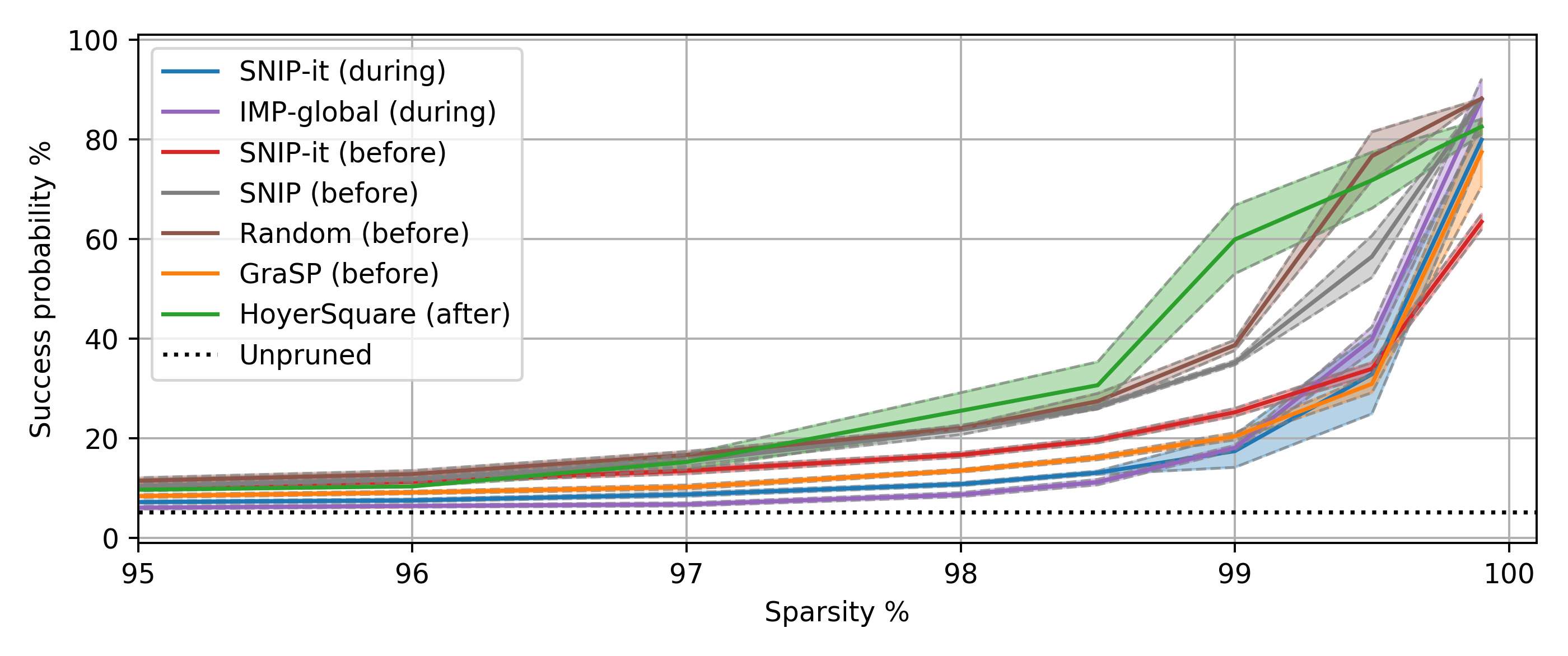} \label{fig:adversarial:eps_is_set}
}
 \subfloat[\textit{Success prob. vs. max $\ell_2$-distance at $\kappa=0.99$.}]{
  \includegraphics[width=0.47\linewidth]{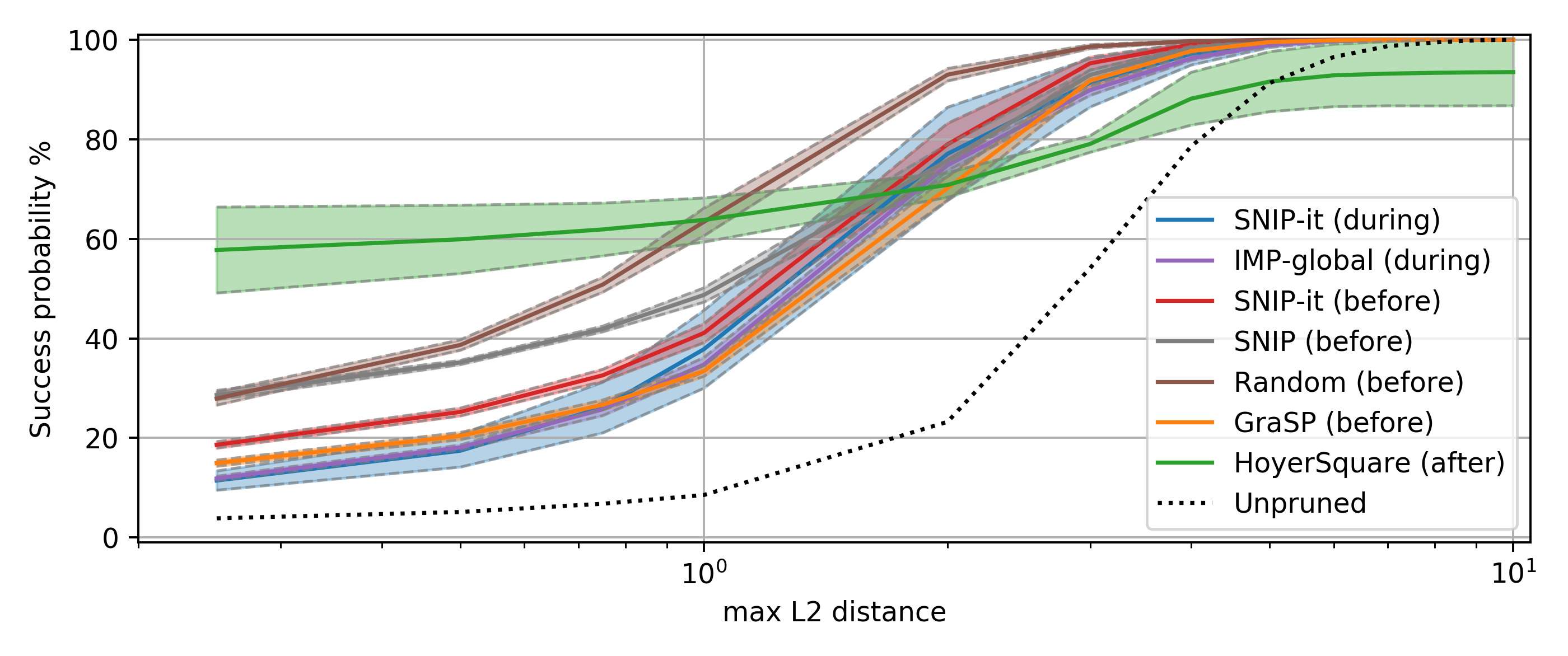} \label{fig:adversarial:sparsity_is_set}
}
  \caption{\textit{Robustness to $\ell_2$-CarliniWagner attacks \citep{carlini2017towards} as probability of success, for MLP5 on MNIST.}}
  \label{fig:adversarial}
\end{figure}

Consequently, we would also expect the methods that better conserve the input layer, to be more robust to adversarial attacks, since they don't over-fit on a few pixels \citep{modas2019sparsefool}. Therefore, we subject the same setup to $\ell_2$-CarliniWagner attacks \citep{carlini2017towards} - using open-source library Foolbox \citep{rauber2017foolbox}. Experiments have shown that these attacks, given unlimited $\ell_2$-distance between the adversarial and real picture, approach a perfect success-rate \citep{carlini2017towards, ren2020adversarial}. Therefore, we evaluate our setup at capped maximum distances. We discover that, for the low-distance regime where the $\ell_2$-norm $ \leq 1$, the pruning methods which maintain input layer are indeed more robust in the face of these attacks. This is displayed on the left of Figure \ref{fig:adversarial:sparsity_is_set}. However, when we get to higher distances this does not hold. Generally, \iterativesnip\ significantly improves in adversarial robustness from SNIP \citep{lee2019snip}, yet performs on-par with some other baselines in the high sparsity regime (Figure \ref{fig:adversarial:eps_is_set}) and also in the low $\ell_2$-distance regime (Figure \ref{fig:adversarial:sparsity_is_set}). 
\vspace{-5pt}
\vspace{-5pt}
\section{Conclusion}
In this work we demonstrated that applying the sensitivity criterion iteratively in a simple fashion creates a more robust and better performing algorithm in the high-sparsity regime, without adding supplemental complexity. With this, we achieved state-of-the-art sparsity-performance trade-offs. Within the unstructured domain, we further showed that our approach significantly improves overfitting and connectivity problems that were native to vanilla SNIP \citep{lee2019snip}. Moreover, we introduced a novel structured method, which operates \emph{before training} and achieves significant speedups that can also be exploited for the training process itself. Furthermore, we related structured and unstructured pruning using the sensitivity criterion, showing we can now compare them on an equal footing. However, the latter we leave for future work to research. 

\begin{ack}
This work was supported, funded and supervised by, \textit{(a)} the University of Amsterdam, in cooperation with \textit{(b)} the Amsterdam-based company BrainCreators B.V. 
\end{ack}

\section*{References}

\medskip
\small

\bibliographystyle{abbrv}

\renewcommand{\bibsection}{}
\bibliography{references}

\newpage
\Large{\textbf{Supplementary Material}}
\appendix
\normalsize

\section{Elasticity histograms}
\label{sec:append:histograms}
\begin{figure}[H]
    \centering
     \subfloat[\textit{LeNet5}]{
  \includegraphics[trim={0.1cm 0.1cm 0.1cm 1.35cm},clip,width=0.41\linewidth]{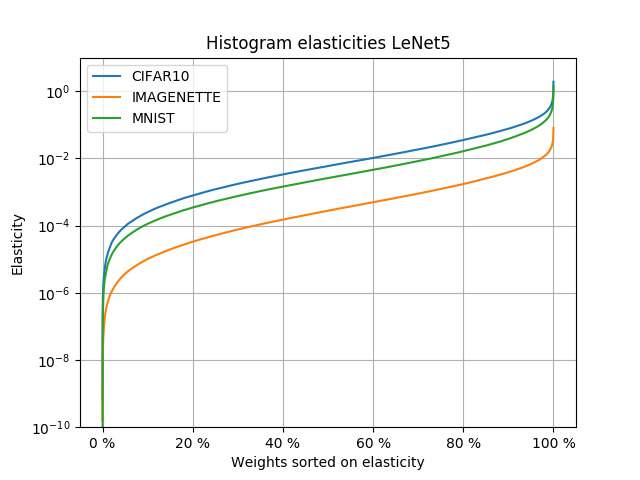}         \label{subfig:elasticities_under_loop:lenet5}
}
 \subfloat[\textit{VGG16}]{
  \includegraphics[trim={0.1cm 0.1cm 0.1cm 1.35cm},clip,width=0.41\linewidth]{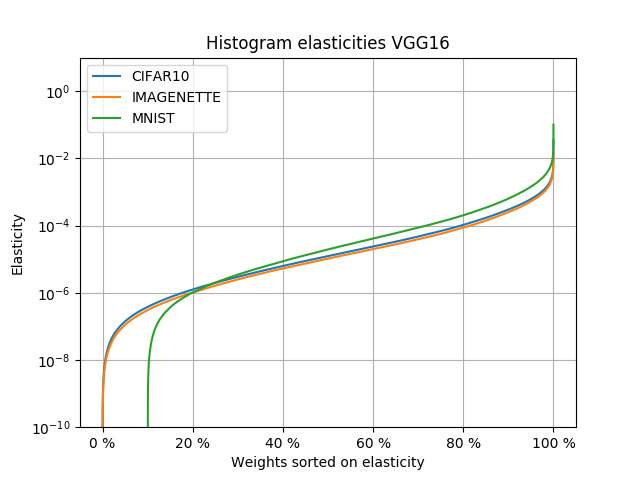} \label{subfig:elasticities_under_loop:vgg16}
}
    \caption{\textit{Elasticities of loss function w.r.t. the network-weights at initialisation for different datasets. Elasticities are sorted like a Lorentz-curve (yet don't sum up to 1). Note that elasticities are displayed in log-scale.}}
    \label{fig:elasticities_under_loop}
\end{figure}

When examining the distribution of the elasticity signal that all parameters in a network receive, it is empirically observed that virtually all weights are in the inelastic category; a great deal of them are nearly completely inelastic, some are at most a little elastic and only very few approach unit-elasticity or higher $\varepsilon_L[\theta_{ij}] \geq 10^0$. Examples of these elasticities are displayed in log-scale in Figure \ref{fig:elasticities_under_loop}. This is a pattern that is empirically found to re-emerge in every single network-dataset combination that has been evaluated in this research.

\section{Exploring combined structured and unstructured pruning}
\label{sec:append:cnip}

We explore a combined structured and unstructured approach. This would materialise as actually taking the union of both weights and nodes in Algorithm \ref{alg:snipit:drawing} on line 2 and subsequently evaluating and pruning them on an equal footing, with the elasticity framework we introduced. We will refer to this here as \lq\combinediterativesnip\rq.
\begin{figure}[H]
    \centering
     \subfloat[\textit{Weight-sparsity}]{
  \includegraphics[width=0.38\linewidth]{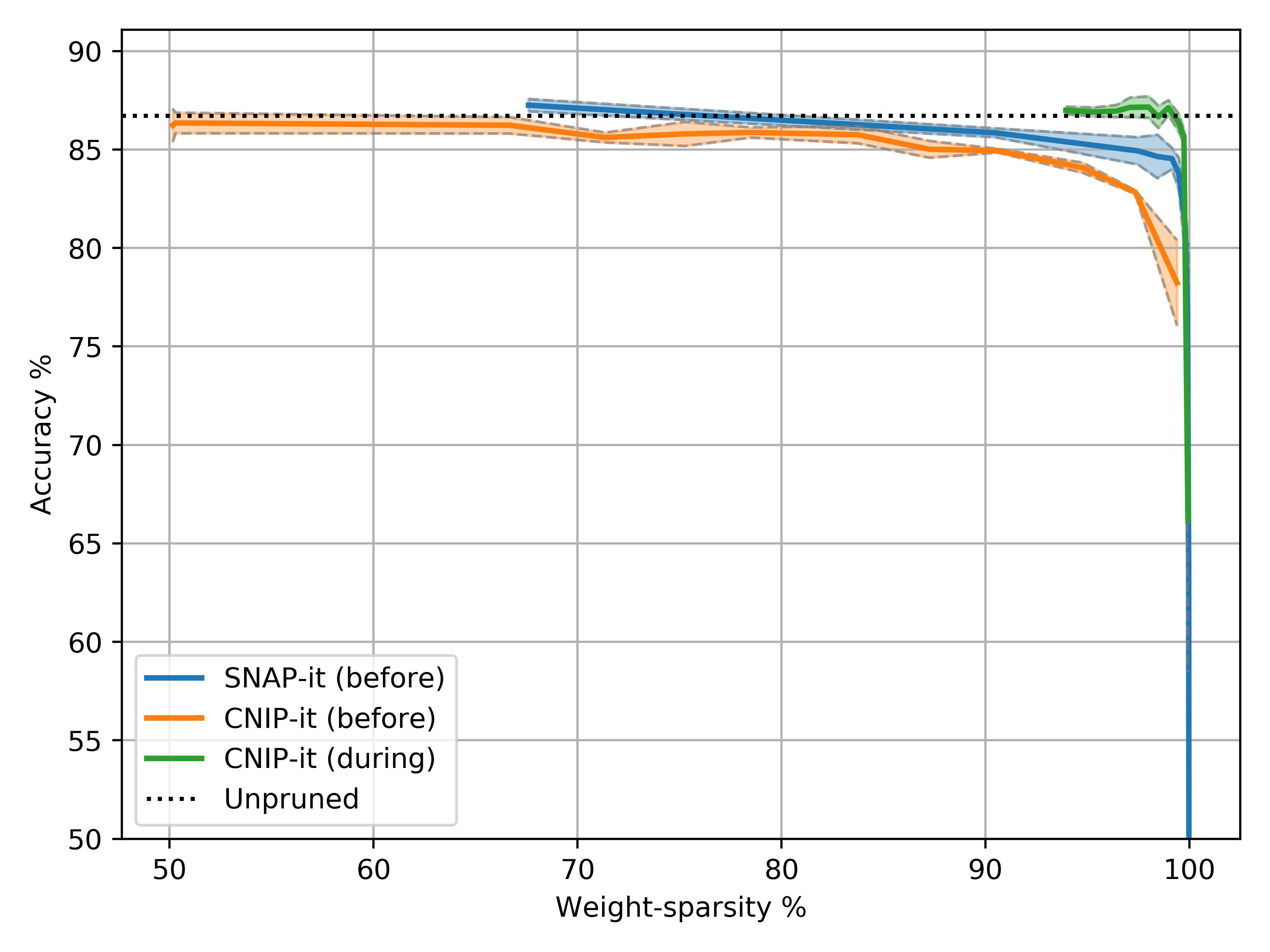}         \label{subfig:cnip:weight}
}
 \subfloat[\textit{Node-sparsity}]{
  \includegraphics[width=0.38\linewidth]{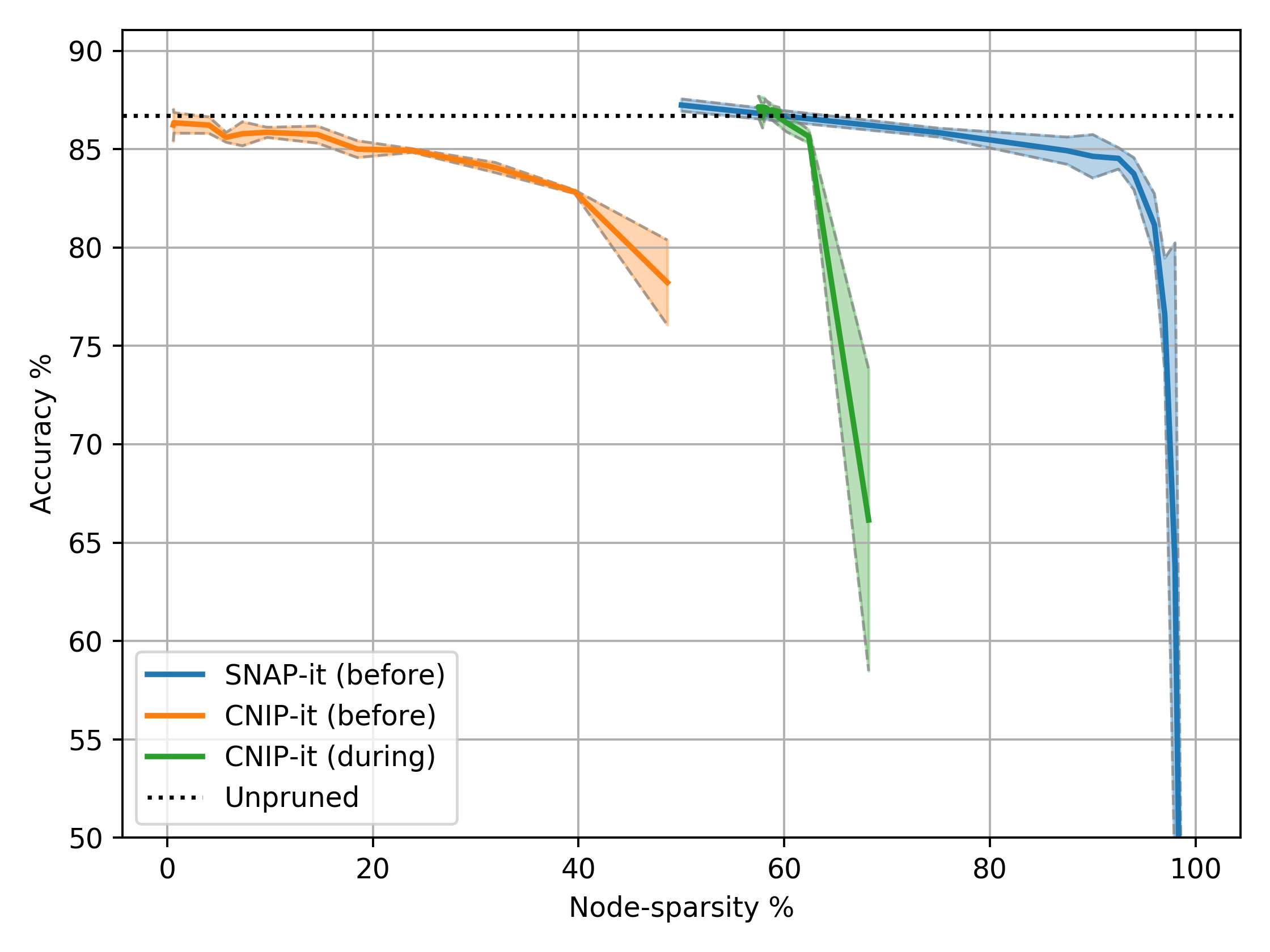} \label{subfig:cnip:node}
}
    \caption{\textit{Weight-sparsity (left) and node-sparsity (right), versus performance of \combinediterativesnip\ on VGG16 (CIFAR10) compared to our own method \iterativestructuredsnip.}}
    \label{fig:cnip}
\end{figure}

We observe that \combinediterativesnip\ reaches very high weight-sparsity levels without much loss in accuracy. However, the node sparsity is rather low. See Figure \ref{fig:cnip} for an illustration of this finding; \combinediterativesnip\  has higher weight-sparsity than \iterativestructuredsnip\ in Figure \ref{subfig:cnip:weight}, yet notable lower node-sparsity in Figure \ref{subfig:cnip:node}. This is not at all unexpected as we presume most nodes are more important to the final loss than most individual weights. This finding may be desirable for some practical applications but it does not lead to the same speedups as \iterativestructuredsnip\ does or to the same weight-sparsity as \iterativesnip\ does. Consequently, it is harder to compare to baselines, since combining the two structure-types was, to the best of our knowledge, not explicitly mentioned in literature. Therefore we concluded the limited applicational value was a ground to not make \combinediterativesnip\ the main focus of the paper. 

\section{Memory footprint prior to training}
\label{sec:append:memory}
\begin{figure}[H]
  \centering
 \includegraphics[width=0.4\linewidth,trim={0.1cm 0.1cm 0.1cm 0.3cm},clip]{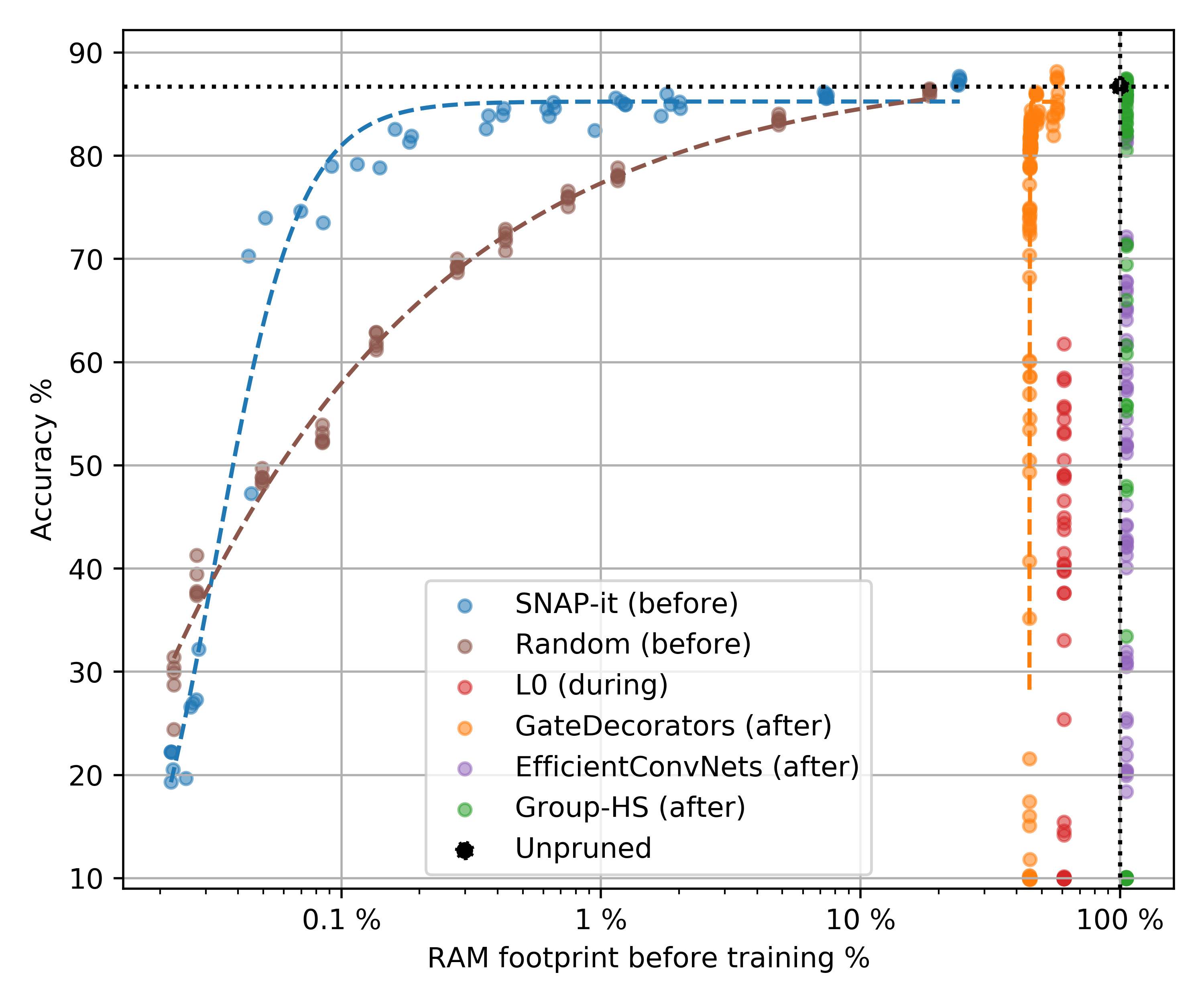} 
  \caption{\textit{Trade-off between RAM-footprint at the start of training and final performance for VGG16 on CIFAR10, with trend-line, showing the benefits on RAM for structured pruning before training.}}
  \label{fig:ram_accuracy}
\end{figure}

\FloatBarrier
When one prunes structurally before training one gains memory and computational benefits from the start. This is reflected in Figure \ref{fig:ram_accuracy}, where we unsurprisingly identify that the two methods that perform structured pruning before training, can substantially reduce their RAM-footprint before training, whereas the others still have the full network in memory at that point. As mentioned in the paper, this allows us to jointly increase the batch-size and learning rate in return \citep{goyal2017accurate}.
\section{Schematic illustration \iterativesnip}
\label{sec:diagram}
\FloatBarrier

\begin{figure}[H]
    \centering
    \captionsetup{width=.8\linewidth}
    \includegraphics[width=0.9\linewidth, trim={0.5cm 15cm 6cm 9cm},clip]{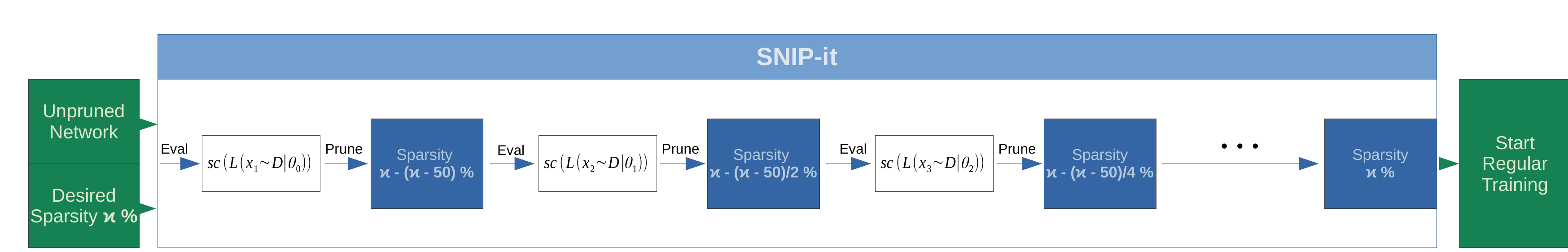}
    \caption{\textit{Schematic illustration of \iterativesnip, as discussed in Algorithm \ref{alg:snipit:drawing}.}}
    \label{fig:snipit:drawing}
\end{figure}

\FloatBarrier
\section{Documenting abandoned approaches}
\label{sec:append:abandon}
In order to prevent future research repeating work we already found to be ineffective, we will provide information on said findings. First of all, we pursued using GraSP's \citep{wang2020picking} criterion as a criterion for \iterativesnip. This failed because when applied iteratively it tried to reactivate already pruned weights in a far greater proportion than it pruned the remaining weights. On the other hand, when the pruned weights were omitted from consideration, it resulted in disconnection. Moreover, we tried applying GraSP to a structured setting and found that, although it works, it doesn't yield favourable results. Additionally, we undertook experiments with  \iterativestructuredsnip\ (during training), but found it performed on-par with \iterativestructuredsnip\ (before training), so we omitted the former as it delays the availability of the benefits of structured pruning to inference time, in contrast to the latter. Finally, we attempted to automatically estimate the optimal sparsity rate $\kappa_{final}$ without tuning this parameter, so that it doesn't have to be set by a user and still have close to no drop in performance. For this we utilised a sorted histogram of the elasticity signal (see Figure \ref{fig:elasticities_under_loop}). Although this was somewhat accurate, we couldn't find an exact enough point and had too many exceptions to justify discussing it in the paper. However, we encourage future research to examine this further.

\section{Hyperparameter tuning}
\label{sec:append:tuning}

All baselines were subject to considerable hyperparameter tuning. In its regard we will provide details in this section. All hyperparameters not discussed here are all kept  on the recommended values from their native papers. For all tuning experiments we used a random held-out validation set, consisting of 20\% of the training set. We don't perform tuning on non-pruning parameters that concern aspects like network or optimisation - e.g. learning rate or LeakyReLU-slope - but instead set these by hand.
\begin{itemize}
    \item Firstly, IMP \citep{frankle2019lottery} has three tuneable parameters: the interval between pruning events $\tau$, the rate of pruning $\kappa_i$ and the rewind-epoch $e_k$. For the rate of pruning $\kappa_i$ we employed the same rule of thumb as \iterativesnip\ (Algorithm \ref{alg:snipit:drawing}) to keep comparison as fair as possible. For the rewind-epoch $e_k$ we used the suggested range of the paper: between 0.1\% and 7\% of convergence \citep{frankle2019lottery}, which for us was an epoch in range $[0, 6]$ and was tuned by grid-search. We found the optimum to be either epochs 5 or 6, with 6 being more frequent. Hence we settled on 6 and froze it for all experiments. The interval $\tau$ was jointly tuned by grid-search also. The original paper \citep{frankle2018lottery} claimed that rewinding should occur early in training, as 'winning tickets' can be found early in training. That, together with the desire to have the final sparsity as early as possible to allow for fine-tuning, led us to aim tuning for a lower value. We tuned in the range $[1, 10]$ and found that it was optimal at 4 or 6. Hence, $4$ was the lowest value where results were stable, which is the value we then also applied to \iterativesnip\ ($\tau$ in Algorithm \ref{alg:snipit:drawing}) to keep things equal as much as possible.
    
    \item Secondly, the baselines of $\ell_0$-regularisation \citep{louizos2017learning} and HoyerSquare \citep{yang2019deephoyer} are both based on a regularisation parameter $\lambda$ which usually requires tuning for the best sparsity-performance trade-off, instead of setting the preset sparsity rate $\kappa$ like other methods - e.g. GraSP \citep{wang2020picking} or \iterativesnip. However, we're actually exploring a range of sparsities for our experiments. Practically, we used $10$ logarithmic intervals in the range $[10^{-6}, 5]$, which generated the data we worked with for these methods. So this is technically not tuning but part of the experiment. 
    \item Thirdly, we manually tuned the threshold for HoyerSquare and Group-HS \citep{yang2019deephoyer} with 7 logarithmic steps in the range $[10^{-5}, 10^{-2}]$ and finally settled on $10^{-3}$. Even though the original paper recommended $10^{-4}$, we found this worked better.
    \item Finally, for all methods that prune \emph{after training} we train for 70 epochs, prune and thereafter perform fine-tuning for another 10 epochs. This was tuned by hand. Methods that are not in this category are trained for the same number of total epochs (80), except $\ell_0$-regularisation \citep{louizos2017learning} and GateDecorators \citep{you2019gate}, which required twice that to converge. Again tuned by hand.
\end{itemize}

\section{Implementation}
\label{sec:append:impl}
All work was implemented in Python \citep{python_lang}, using automatic differentiation library PyTorch \citep{paszke2017automatic}. Code is freely available at \url{https://github.com/StijnVerdenius/SNIP-it}

\section{Metrics}
\label{sec:append:metrics}
This section is dedicated to how the specifics metrics are calculated with the aim of reproducibility. They consist of two metrics that are directly measurable and two that are estimations. Said metrics are always only reported in the paper as a reduction w.r.t. the unpruned baseline and are calculated as follows:

\begin{itemize}
    \item \textbf{FLOPS} are a proxy for computational effort. We have not directly measured these from the machines but instead estimated them. In literature multiple estimation methods coexist. We used the estimation technique from the implementation of \citep{liu2018rethinking}, yet slightly altered for integration in our implementation.
    \item \textbf{Time} is a metric that can be measured directly. However, its measurements can depend on the host machine's capabilities, external processes running on the same machine, temperature and/or other unknown factors. We measured inference time on the GPU with PyTorch's CUDA time measurement tool \citep{paszke2017automatic} and training time as the total time of a run in milliseconds.
    \item \textbf{RAM-footprint} is how much memory the process takes up on the GPU. This effectively dictates the maximum combined model-size and batch-size that can be run at the same time. However, its measurements can depend on external processes as well. Furthermore, the epoch that measuring takes place can play a role. Hence, we document both point-measurements and median values over longer periods of time. We measured RAM on the GPU directly with PyTorch's CUDA memory measurement tool \citep{paszke2017automatic}.
    \item \textbf{Disk storage} was estimated as the possible size when storing by CSR-format \citep{bulucc2009parallel, saad2003iterative}. Here, a sparse matrix is stored in three vectors: one with all nonzero entries and two with its indices. By considering only the nonzero entries and storing said indices in a lower precision we obtain compression. When we estimated the disk size we used a ratio of 16:1 for precision of the nonzero floats and the indices respectively. This is somewhat on the high-end, for more conservative estimations one could consider the ratios 8:1 or 4:1 as well.
\end{itemize}

\section{Infrastructure}
\label{sec:append:infr}

All experiments were run on a distributed scheduling system with an array of different computers with different GPUs. The GPUs available to this system were GeForce RTX 2080 Ti, TITAN X, GeForce GTX 1080 Ti and Tesla V100-DGXS-16GB. Furthermore, the operating system for all was Ubuntu 18.04. All run-configurations were equally likely to end up at each machine. We are aware that this may cause higher variance in raw time measurement and therefore assert those are taken with reservation. However, we empirically observed it depended disproportionately more so on the network architecture and dataset used. Overall, experiments' run-times were observed to be between roughly a few minutes and five hours - depending on its setup. 

\section{Datasets}
\label{sec:append:data}

Three different datasets were adopted to demonstrate the effectiveness of the methods described. A short description will follow of the dataset and how it was handled.
\begin{itemize}
    \item \textbf{MNIST} \citep{lecun1998gradient}  is employed as a baseline dataset for pruning algorithms, for exploratory parts of the research. It contains 60k training images of hand-written digits as well as 10k test images, all of 10 different classes. Images are of input dimensions $1\times28\times28$. The images were normalised, yet no additional data augmentation was performed.
    \item \textbf{CIFAR10} \citep{krizhevsky2009learning} is a classic dataset for pruning algorithms and was chosen as a default dataset to evaluate the algorithms on. It contains 50k images of different object-classes, as well as 10k test images. Images are of input dimensions $3\times32\times32$. The images were normalised. Moreover, data augmentation was applied to the training set, in the form of random horizontal flips with a probability of $20 \%$.
    \item \textbf{Imagenette} \citep{imagenette} is a subset of images from the Imagenet challenge \citep{deng2009imagenet}, freely available online \citep{imagenette}. It was chosen to additionally evaluate on a more realistic benchmark with larger pictures and less available data. It contains 10 classes with 13k training images and 500 test images. Images are preprocessed to have 3 valid channels and its shortest size resized to 320 and aspect ratio maintained. Thereafter, we scale them down to $3\times128\times128$ for computational reasons. The images were normalised. Moreover, training data augmentation was applied, again in the form of random horizontal flips with a probability of $20 \%$.
\end{itemize}{}
\section{Supplementary results}
\label{sec:append:results}

Here we will report and display results that did not make it in the main paper due to spatial constraints. Sections are organised by the dataset or metric that was used to obtained the results that are displayed.

\subsection{MNIST}
\FloatBarrier
\begin{figure}[h!]
 \centering
 \includegraphics[trim={0.1cm 0.1cm 0.1cm 0.3cm},clip,width=0.35\linewidth]{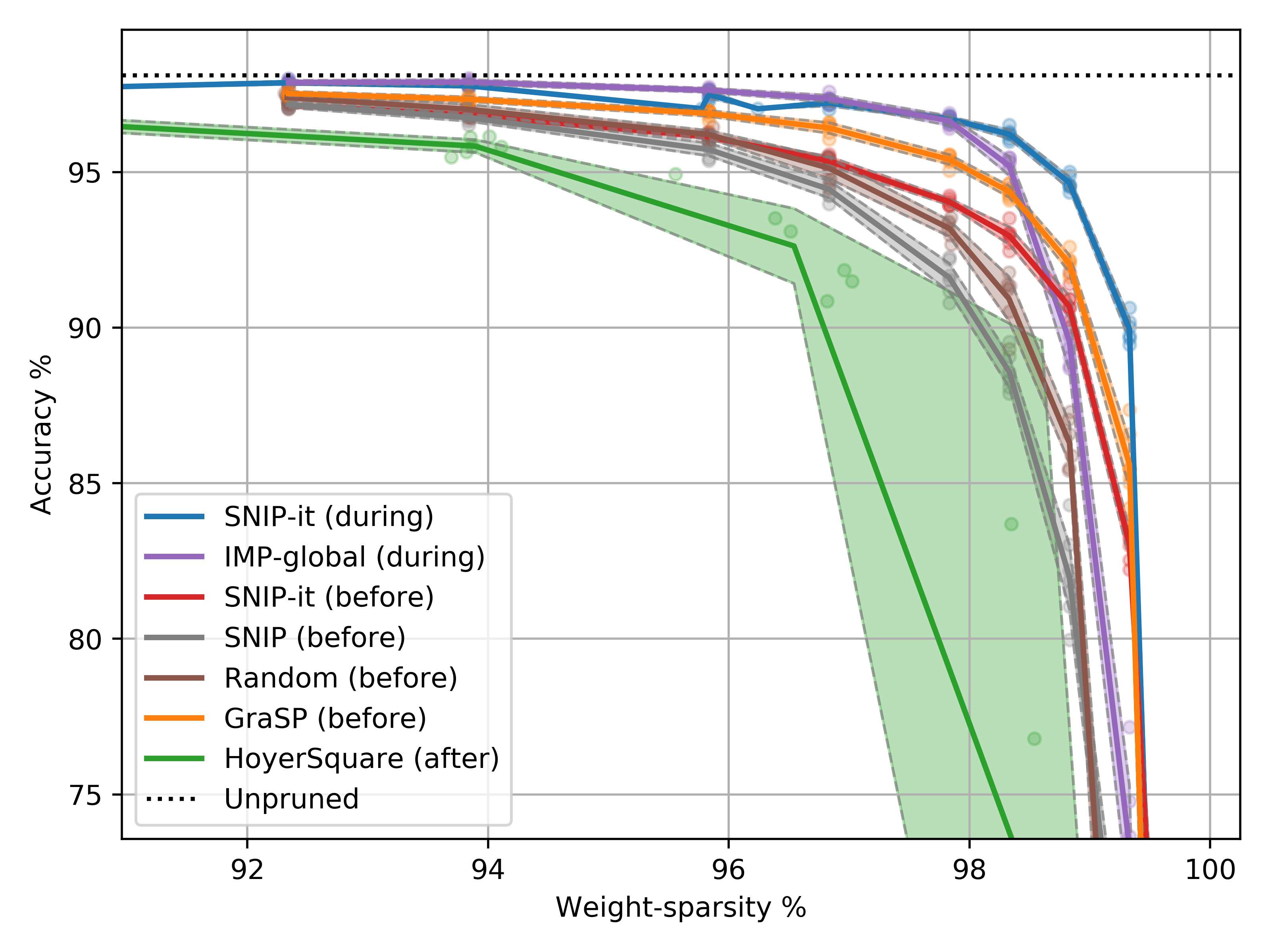} 
 \caption{\textit{Unstructured sparsity-performance trade-off with confidence bound (MLP5). This figure reports points corresponding to Figures \ref{fig:saliencies},  \ref{fig:weights-remaining} \& \ref{fig:adversarial} in the main paper.}}
\end{figure}
\begin{table}
\centering
\caption{\textit{Unstructured sparsity-performance trade-off. Points are picked using the highest harmonic mean (HM) between sparsity and accuracy. This table reports points corresponding to Figures \ref{fig:saliencies},  \ref{fig:weights-remaining} \& \ref{fig:adversarial} in the main paper.}}
\vspace{10pt}
\scalebox{0.7}{ 
\begin{tabular}{@{}r@{\hskip 1.1cm}c@{\hskip 0.3cm}c@{\hskip 0.3cm}c@{}}
\toprule

 \multicolumn{4}{c}{\textbf{MNIST}}  \\ 
& \multicolumn{3}{c}{MLP5} \\
								& \small Accuracy  			 &\small Sparsity 	& \small HM 		\\
\midrule

Baseline 						&  98\%  		         &			&		   \\
 & \multicolumn{3}{c}{\small Prunes after training}   \\ 
HoyerSquare						&  96\%  $\pm$ 0.3  &     94\%     &     95     \\
 & \multicolumn{3}{c}{\small Prunes during training}   \\ 
IMP-global         				&  \bb{97\%  $\pm$ 0.2}  &     \bb{98\%}     &     \bb{97}     \\
SNIP-it \footnotesize{(ours)}   &  \bb{97\%  $\pm$ 0.1}  &    \bb{98\%}     &    \bb{97}     \\
& \multicolumn{3}{c}{\small Prunes before training}   \\ 
GraSP              				&  96\%  $\pm$ 0.2  &     97\%     &     96     \\
Random             				&  96\%  $\pm$ 0.2  &     96\%     &     96     \\
SNIP      						&  96\%  $\pm$ 0.3  &     96\%     &     96     \\
SNIP-it \footnotesize{(ours)}   &  95\%  $\pm$ 0.2  &     97\%     &     96     \\
\bottomrule

\end{tabular}
}
\end{table}
\FloatBarrier
\subsection{CIFAR10}
\label{sec:append:results:cifar}

\FloatBarrier
\begin{figure}[h!]
  \centering
 \includegraphics[width=0.5\linewidth]{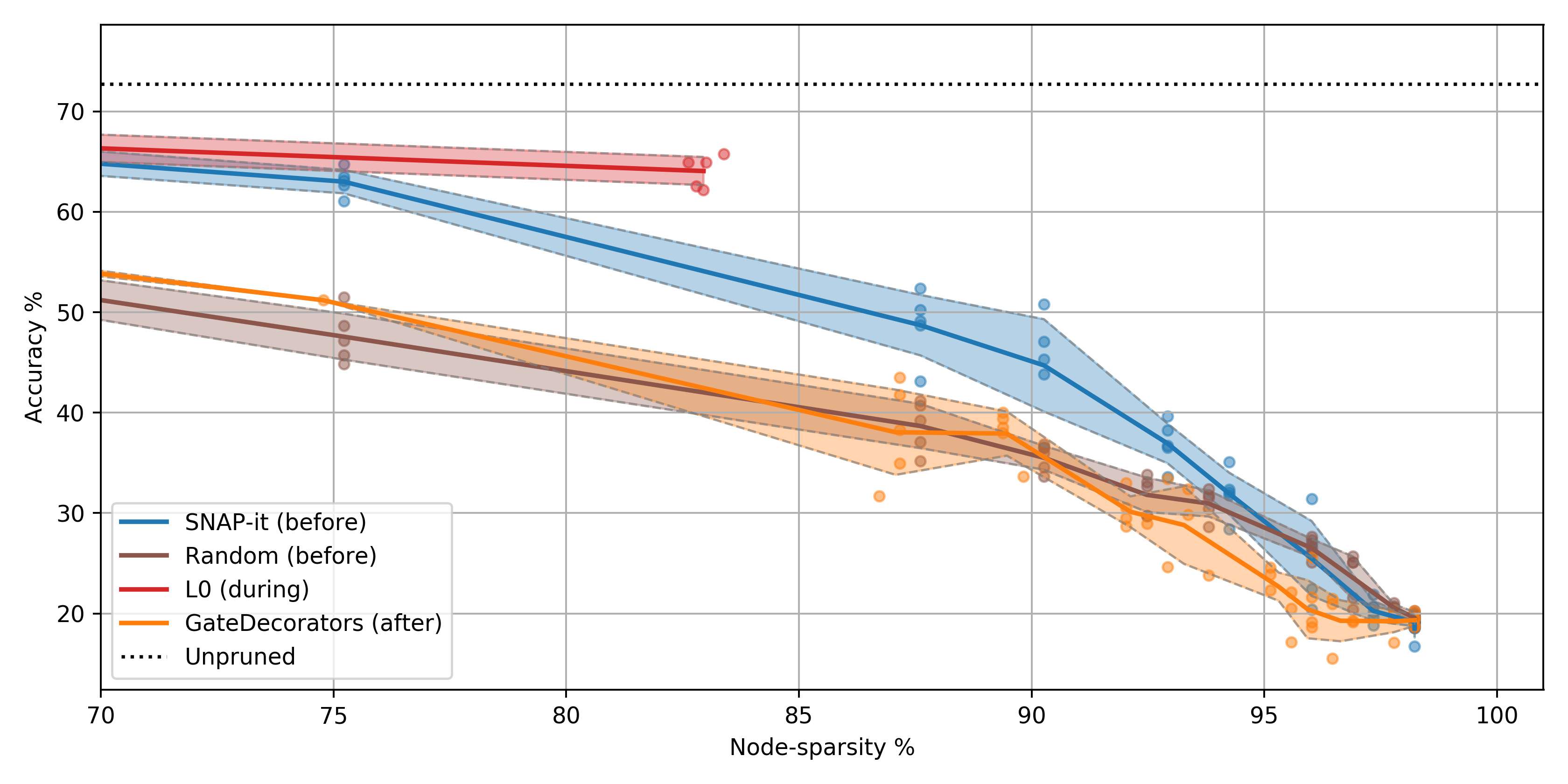}
  \caption{\textit{Structured sparsity-performance trade-off with confidence bound (LeNet5), showing $\ell_0$-regularisation \citep{louizos2017learning} works well with recommended setup from their paper, unlike is observed in other setup-instances.}}
  \label{fig:l0_works_sometimes}
\end{figure}

\begin{table}[h!]
\centering
\caption{\textit{Structured results on CIFAR10. Points are picked using the highest harmonic mean between sparsity and accuracy. This table reports points corresponding to Figure \ref{fig:node_spars} in the main paper.} }
\vspace{10pt}
\scalebox{0.7}{
\begin{tabular}{@{}c@{\hskip 0.75cm}c@{\hskip 1.25cm}c@{\hskip 0.2cm}c@{\hskip 1.25cm}c@{\hskip 0.1cm}c@{\hskip 1.25cm}c@{\hskip 0.1cm}c@{\hskip 1.25cm}c@{\hskip 0.1cm}c@{}}
\toprule
\multicolumn{10}{c}{\textbf{VGG16}}  \\ 

          Method              & Accuracy &  \multicolumn{2}{@{}c@{\hskip 1.25cm}}{Sparsity} & \multicolumn{2}{@{}c@{ \hskip1.25cm}}{FLOPS} 					& \multicolumn{2}{@{}c@{\hskip 1.25cm}}{Time} 					& \multicolumn{2}{@{}c@{}}{Storage} \\
          					  &  	 & \scriptsize Weight & \scriptsize Node    & \scriptsize Inference & \scriptsize Train  	& \scriptsize Inference & \scriptsize Train & \scriptsize Disk & \scriptsize RAM  	\\  	

          \midrule

Baseline                       & 87\%            &&         &&    &&  &&   \\
\multicolumn{10}{c}{ \small Prunes after training}\\ 
Effic.ConvN.				   &      82 \%$\pm$ 0.3       &          94\% & 75\%           &          15\x & 1 \x           &          3 \x & 1 \x           &        6e+02 \x & 1 \x         \\
GateDecorators                 &      83 \%$\pm$ 1.0       &       \bb{99\%} & 88\%         &      \bb{30\x} & 1 \x          &          5 \x & 1 \x           &        3e+04 \x & 2 \x         \\
Group-HS					   &    \bb{85 \%$\pm$ 0.3}    &          96\% & 82\%           &          3 \x & 1 \x           &          3 \x & 1 \x           &        2e+03 \x & 1 \x         \\
\multicolumn{10}{c}{ \small Prunes during training}\\ 
L0                             &      57 \%$\pm$ 4.2       &          86\% & 70\%           &          2 \x & 1 \x           &          1 \x & 1 \x           &        9e+01 \x & 2 \x         \\
 \multicolumn{10}{c}{ \small Prunes before training}\\ 
Random						   &      83 \%$\pm$ 0.4       &          94\% & 75\%           &          15\x & \bb{15\x}      &     \bb{6 \x} & \bb{2 \x}      &        6e+02 \x & 21\x         \\
SNAP-it \footnotesize{(ours)}  & \bb{85 \%$\pm$ 0.6}  	   &     \bb{99\%} & \bb{93\%}      &          11\x & 11\x           &          5 \x & \bb{2 \x}      &   \bb{6e+04} \x & \bb{156\x}   \\
\toprule
 \multicolumn{10}{c}{\textbf{AlexNet}}  \\ 
\midrule
Baseline                       & 83\%		         &&         &&    &&  &&    \\
\multicolumn{10}{c}{ \small Prunes after training} \\
Effic.ConvN.				   &      78 \%$\pm$ 0.6       &          94\% & 75\%           &          14\x & 1 \x           &          3 \x & 1 \x           &        6e+02 \x & 1 \x         \\
GateDecorators                 &      80 \%$\pm$ 0.9       &     \bb{99\%} & \bb{90\%}      &          25\x & 1 \x           &          6 \x & 1 \x           &   \bb{6e+04} \x & \bb{205\x}   \\
Group-HS					   &      80 \%$\pm$ 1.7       &          95\% & 87\%           &          2 \x & 1 \x           &          2 \x & 1 \x           &        1e+03 \x & 1 \x         \\
\multicolumn{10}{c}{ \small Prunes during training}\\ 
L0                             &      72 \%$\pm$ 1.2       &          88\% & 67\%           &          2 \x & 1 \x           &          1 \x & 1 \x           &        1e+02 \x & 2 \x         \\
 \multicolumn{10}{c}{ \small Prunes before training}\\ 
Random						   &      79 \%$\pm$ 0.2       &          98\% & 88\%           &     \bb{50\x} & \bb{50\x}      &     \bb{8 \x} & \bb{2 \x}      &        1e+04 \x & 109\x        \\
SNAP-it \footnotesize{(ours)}  &    \bb{83 \%$\pm$ 0.6 }   &   \bb{99\%} & \bb{90\%}        &          9 \x & 9 \x           &          5 \x & \bb{2 \x}      &        2e+04 \x & 121\x        \\
\end{tabular}
}
\end{table}
\begin{figure}[h!]
  \centering
 \includegraphics[width=0.35\linewidth]{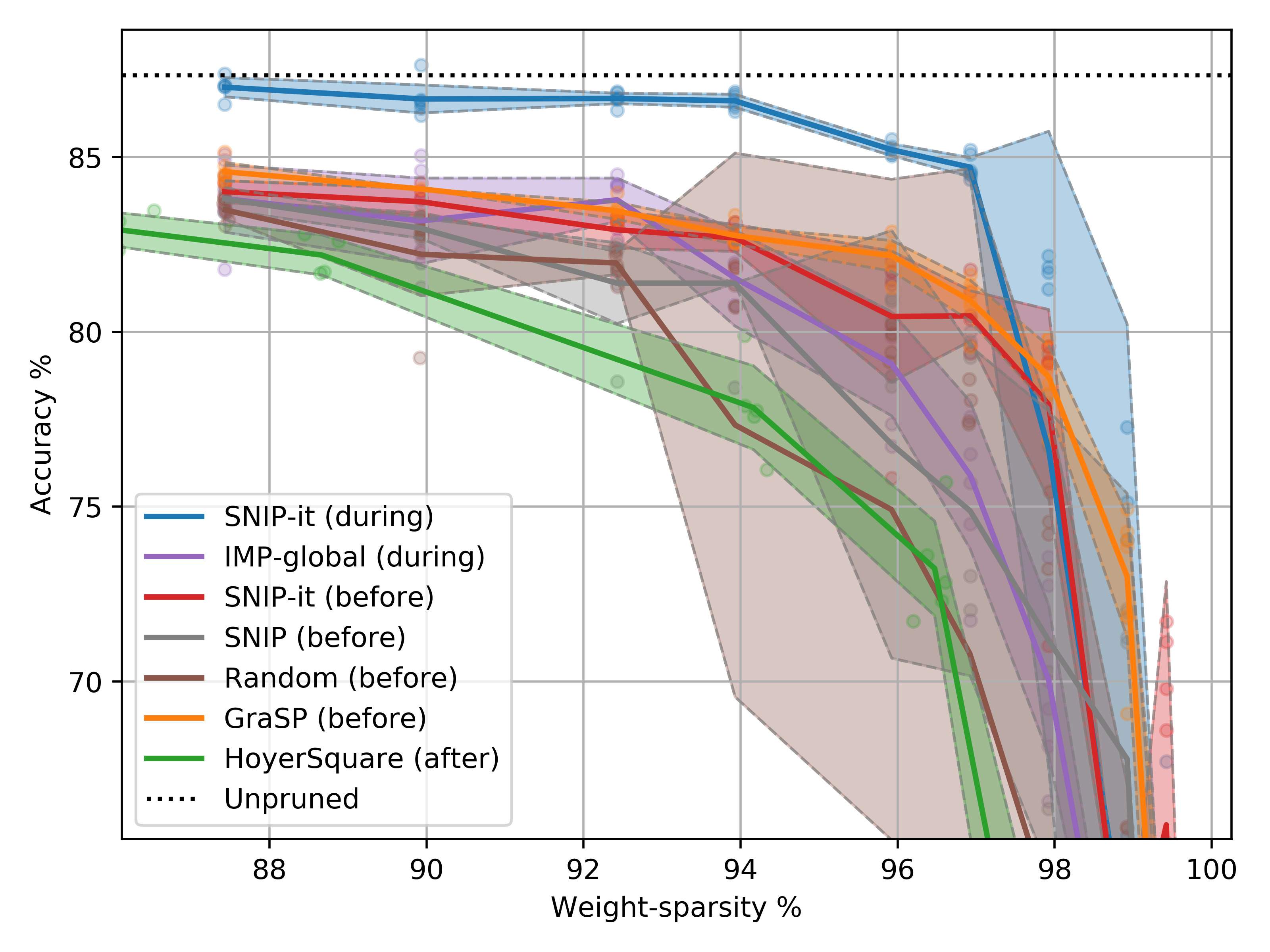}
  \caption{\textit{Unstructured sparsity-performance trade-off with confidence bound (Conv6). This figure reports points corresponding to Table \ref{tab:sparsity} in the main paper.}}
\end{figure}
\FloatBarrier
\subsection{Imagenette}
\FloatBarrier
\begin{figure}[h!]
  \centering
 \includegraphics[width=0.5\linewidth]{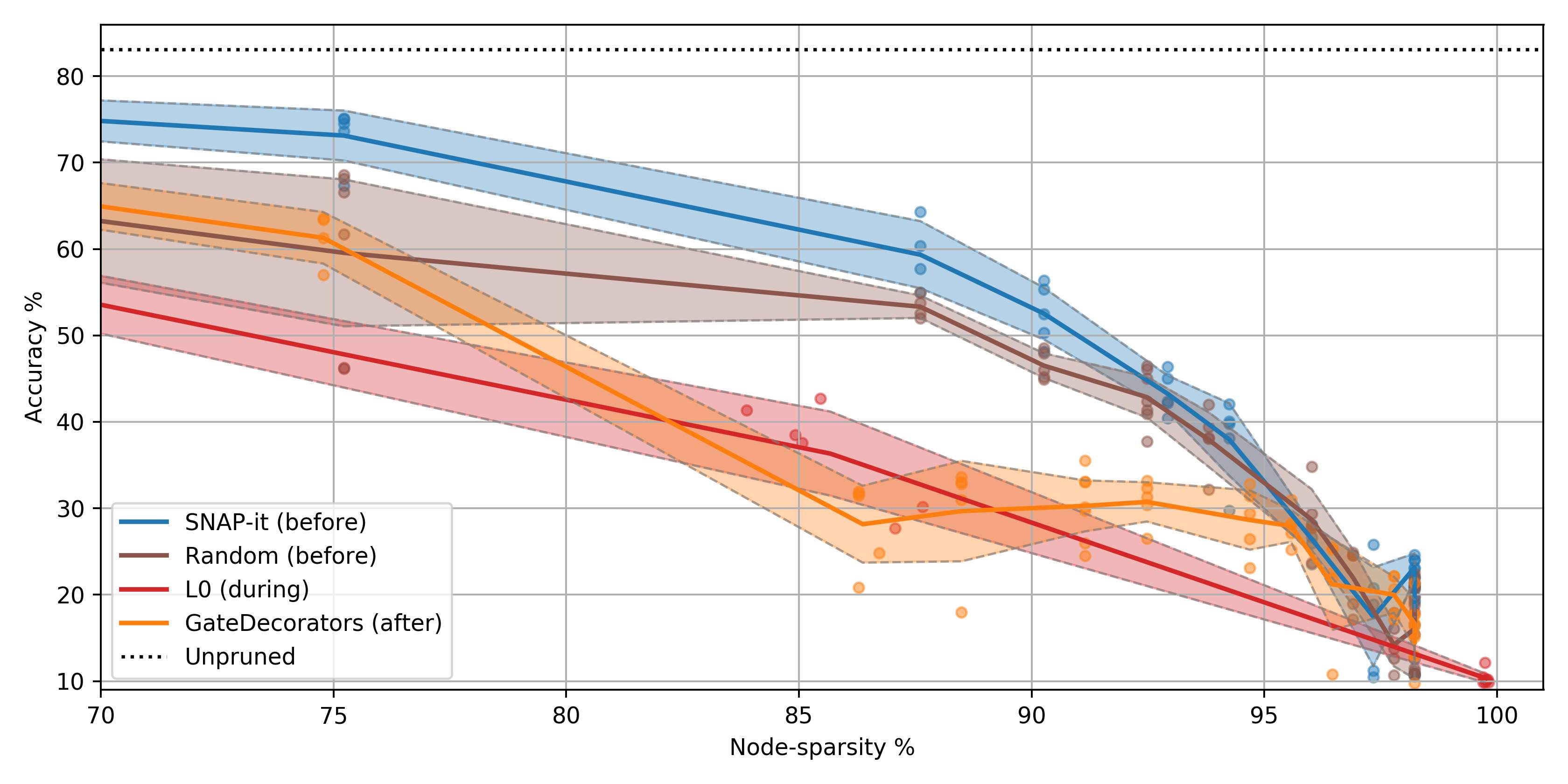}
  \caption{\textit{Structured sparsity-performance trade-off with confidence bound (LeNet5).}}
\end{figure}
\begin{figure}[h!]
  \centering
    \captionsetup[subfigure]{justification=centering}
  \subfloat[\textit{\small LeNet5}]{
  \includegraphics[width=0.3\linewidth]{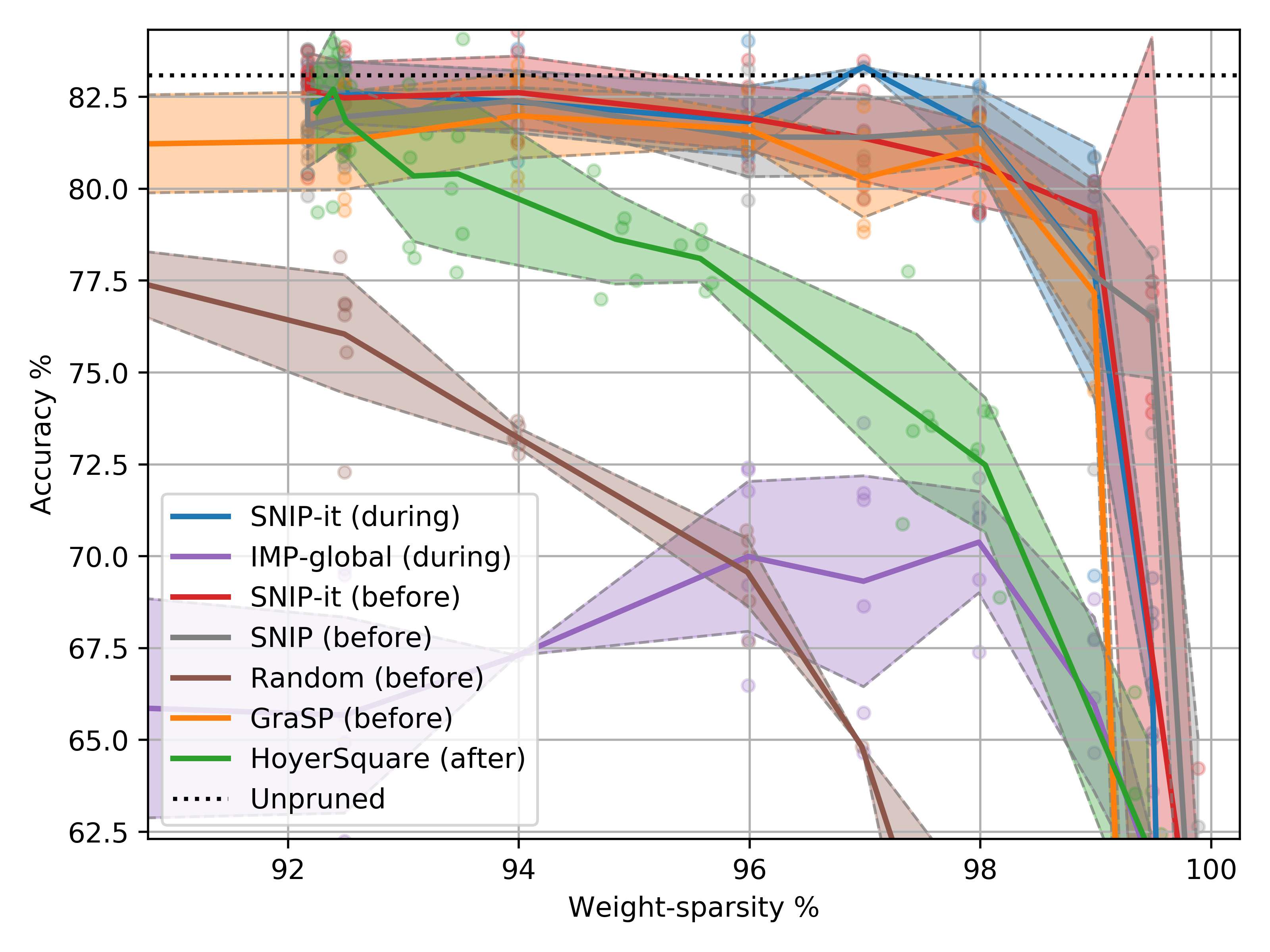} 
}
  \subfloat[\textit{\small MLP5}]{
  \includegraphics[width=0.3\linewidth]{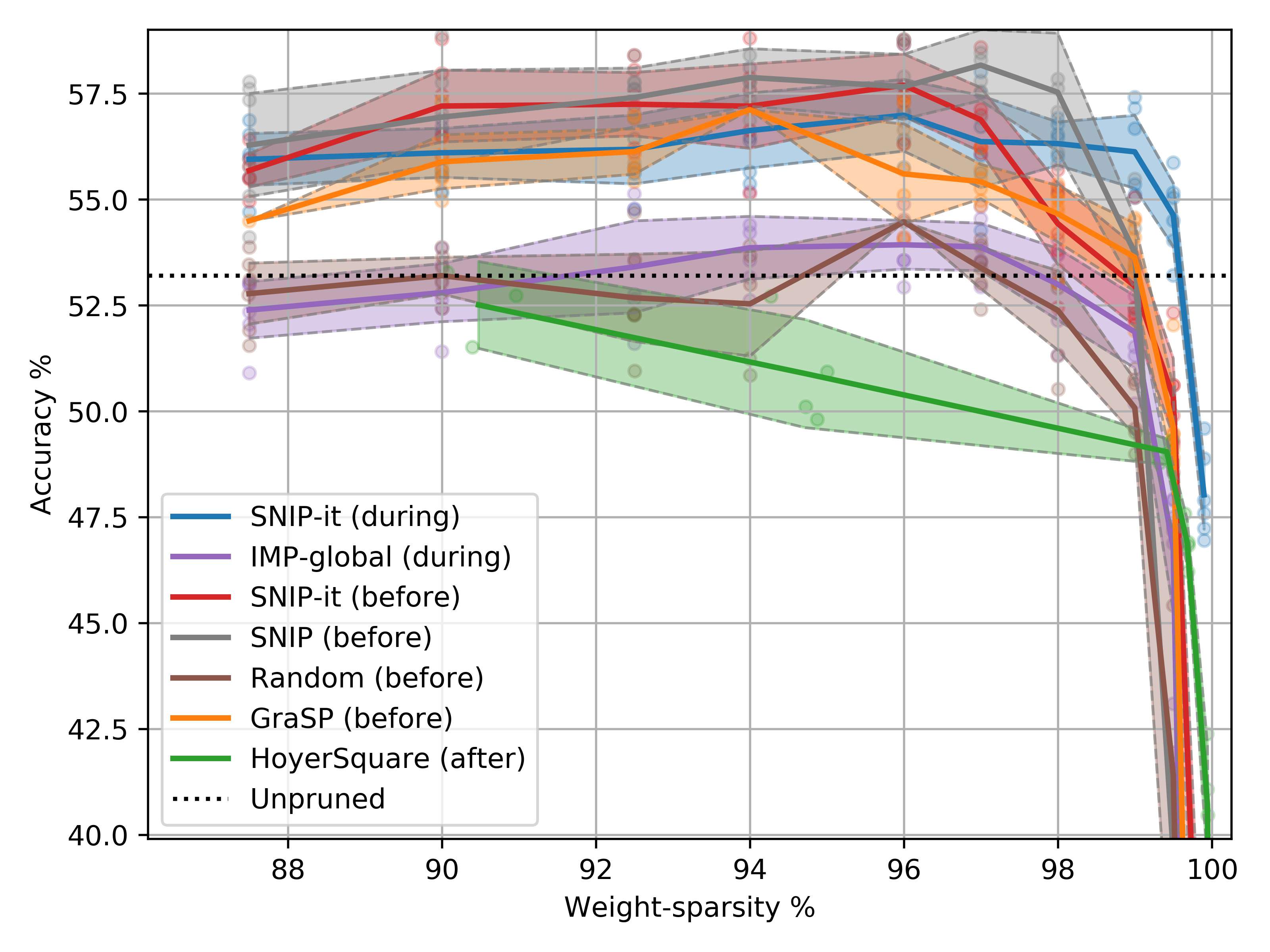} 
}
\hspace{0mm}
  \subfloat[\textit{\small ResNet18}]{
  \includegraphics[width=0.3\linewidth]{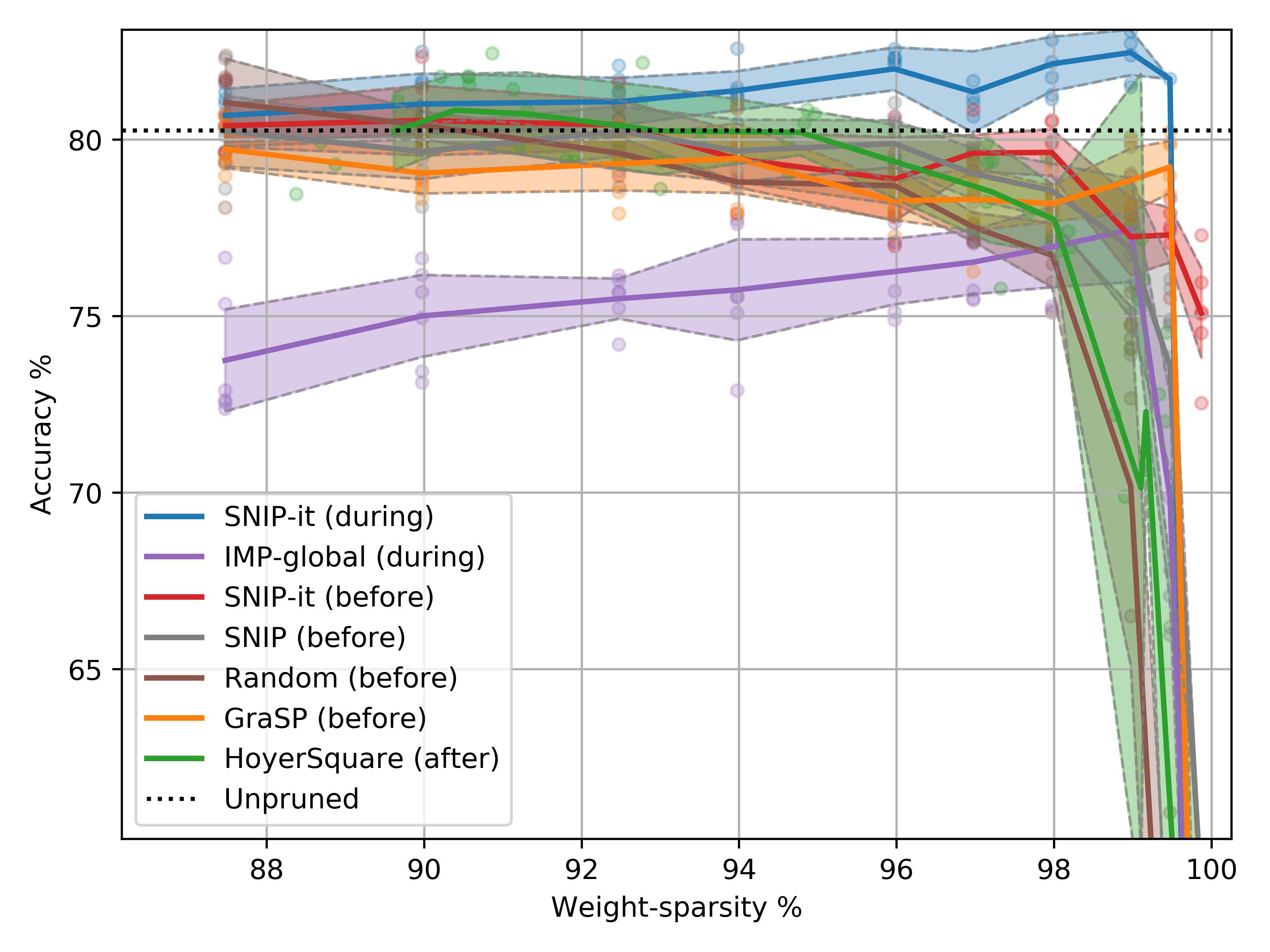} 
}
  \subfloat[\textit{\small Conv6}]{
  \includegraphics[width=0.3\linewidth]{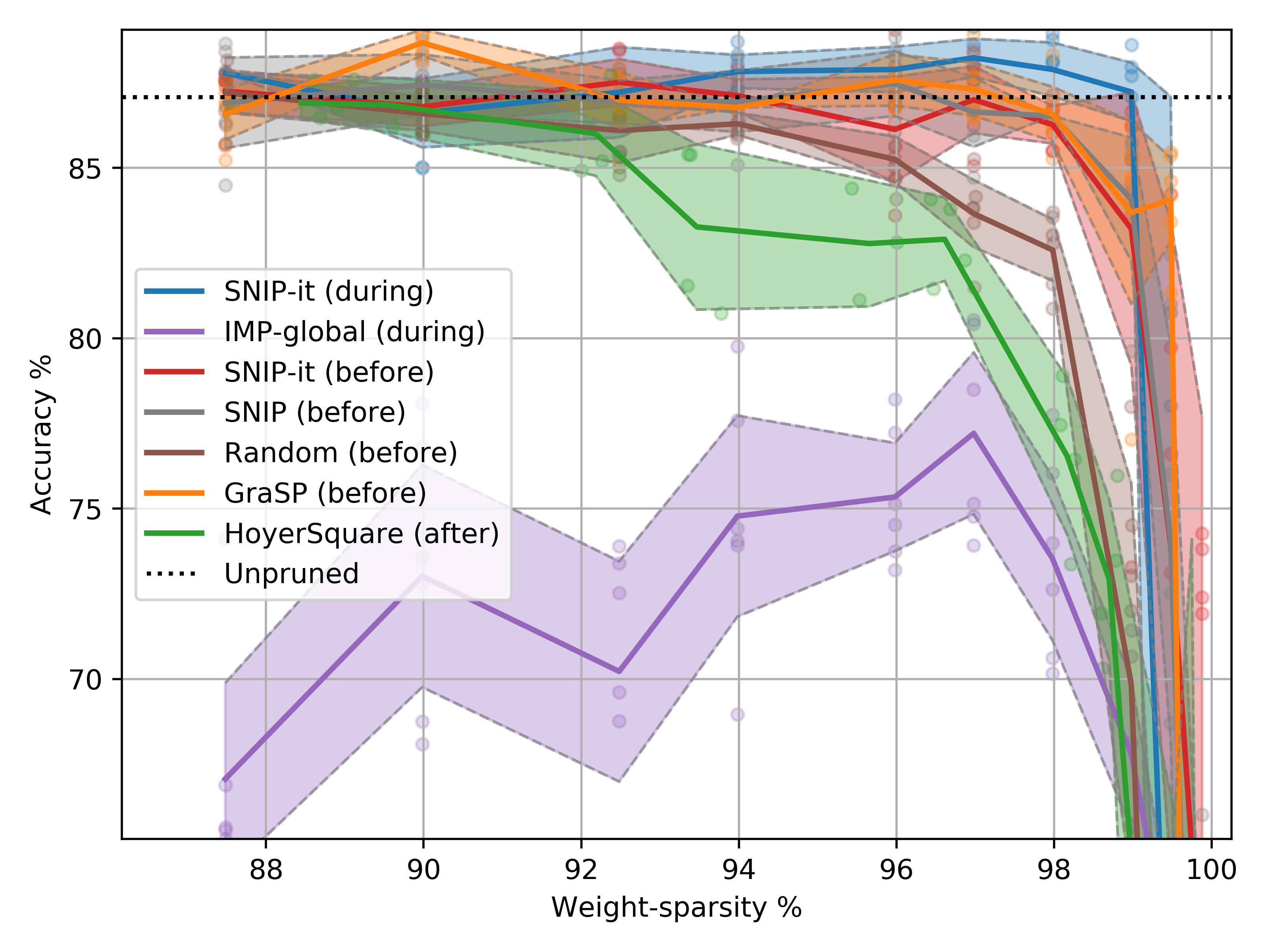} 
}
  \caption{\textit{Unstructured sparsity-performance trade-off with confidence bound. This figure reports points corresponding to Table \ref{tab:sparsity} in the main paper.}}
\end{figure}

\begin{figure}
  \centering
  \subfloat[\textit{\small AlexNet}]{
  \includegraphics[width=0.4\linewidth]{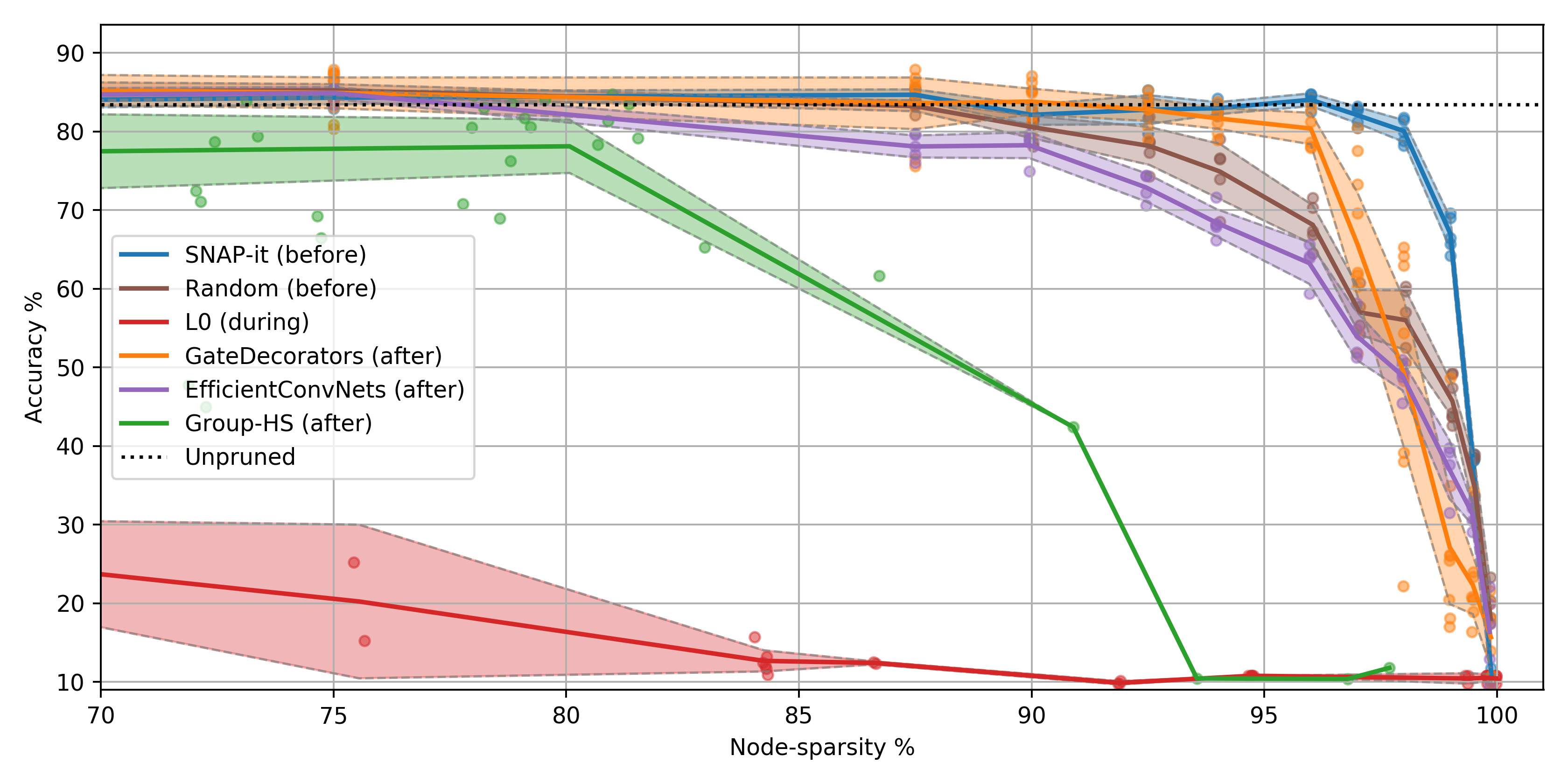} 
}
  \subfloat[\textit{\small VGG16}]{
  \includegraphics[width=0.4\linewidth]{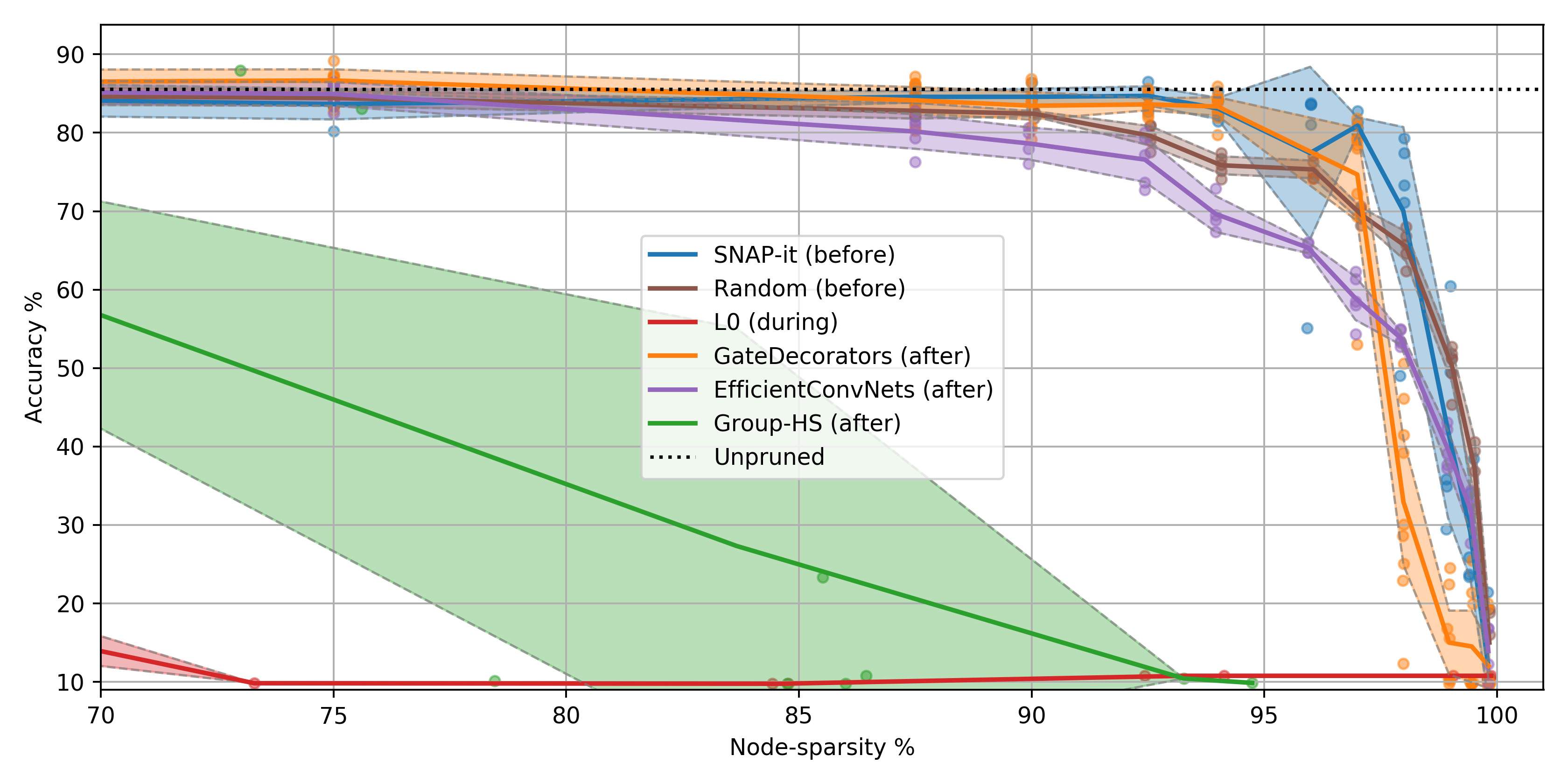} 
}
  \caption{\textit{Structured sparsity-performance trade-off with confidence bound. This figure reports points corresponding to Table \ref{tab:node:imagenette} in the main paper.}}
\end{figure}

\FloatBarrier

\subsection{FLOPS}
\FloatBarrier
\begin{figure}[h!]
\centering
\subfloat[\textit{AlexNet-CIFAR10}]{
  \includegraphics[width=30mm]{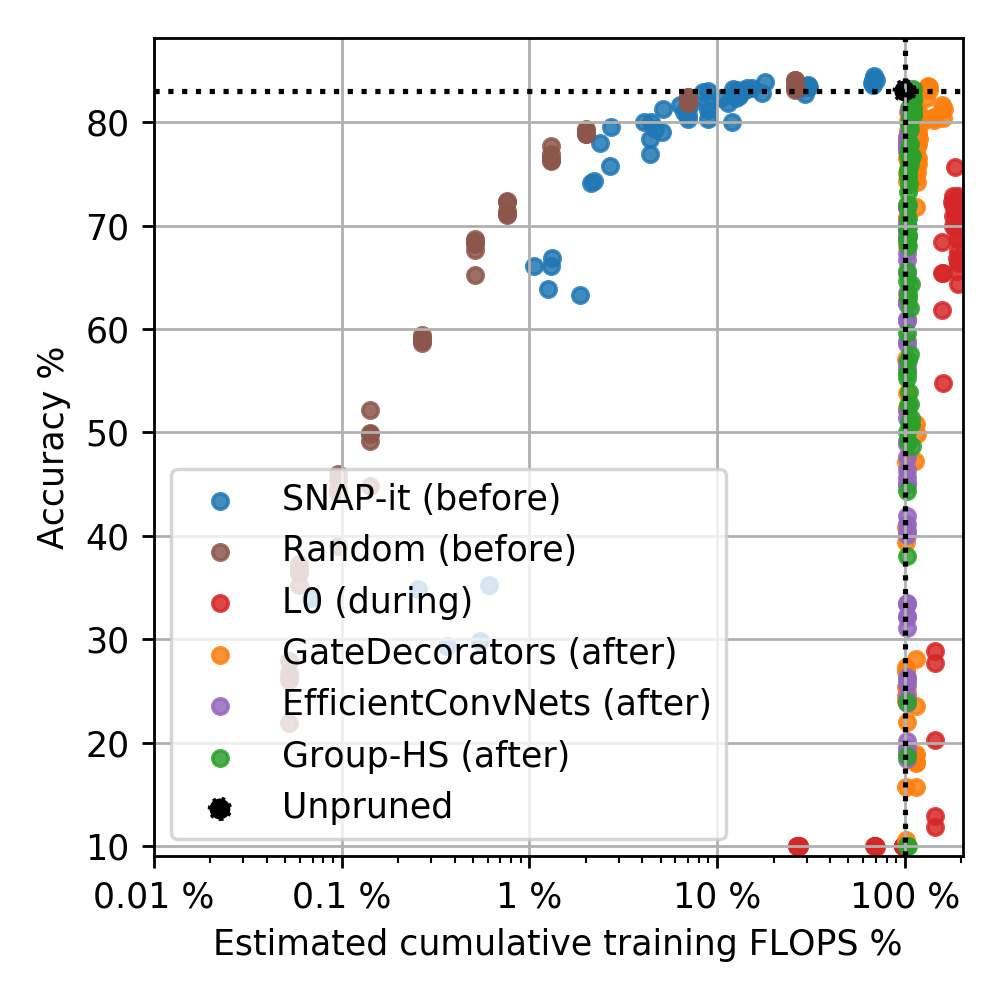}
}
\subfloat[\textit{AlexNet-Imagnette}]{
  \includegraphics[width=30mm]{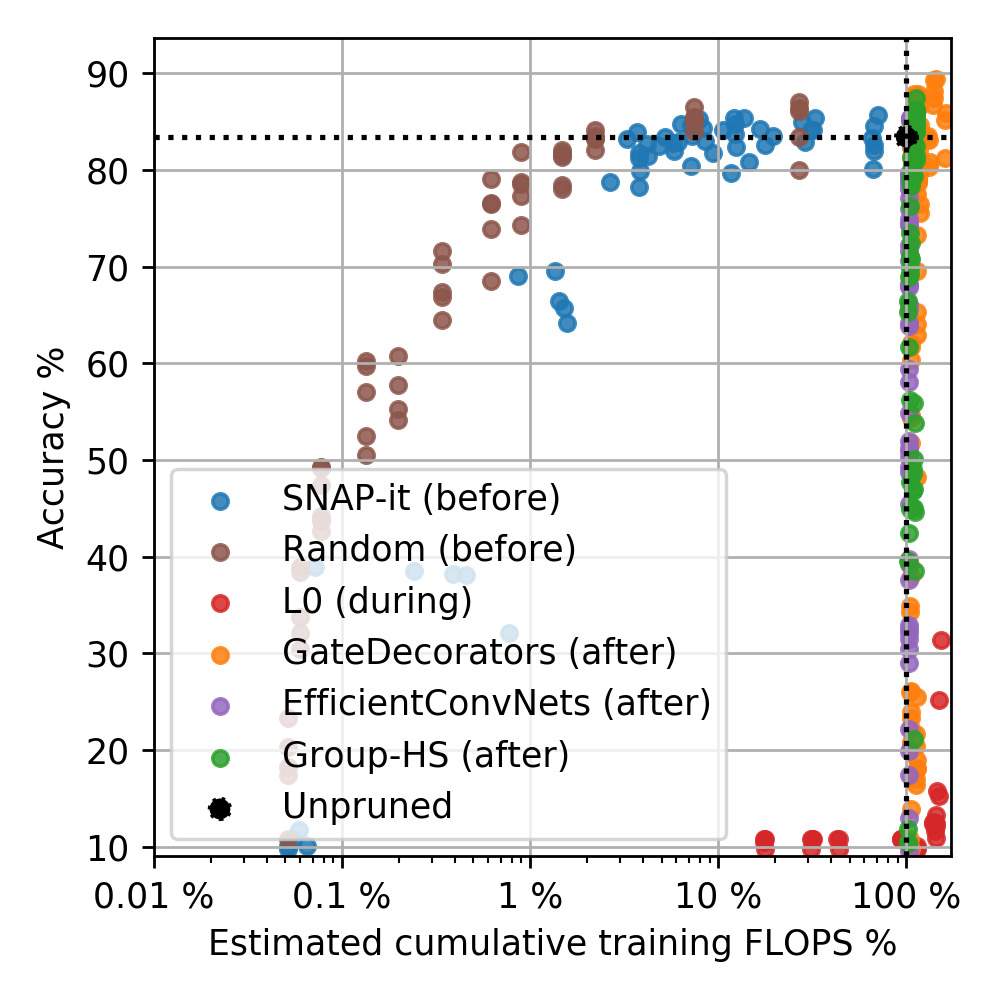}
}
\subfloat[\textit{VGG16-CIFAR10}]{
  \includegraphics[width=30mm]{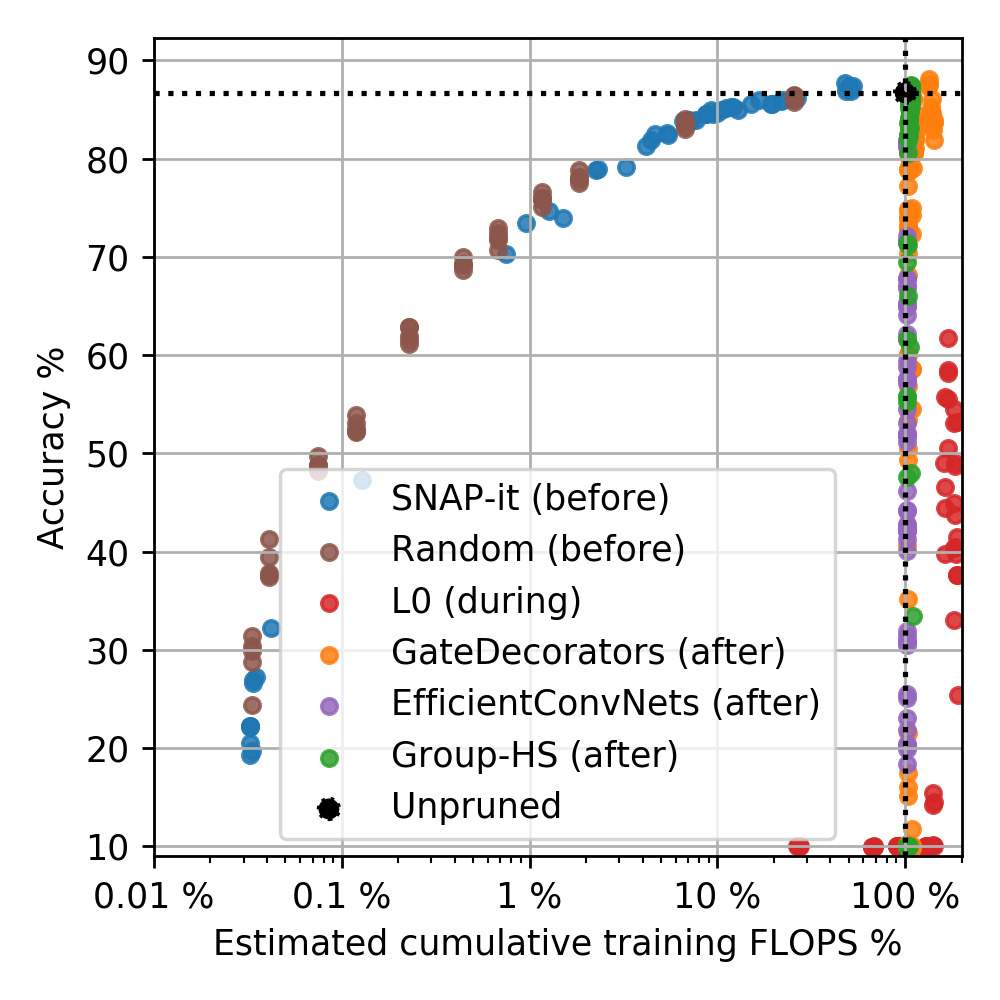}
}
\subfloat[\textit{VGG16-Imagnette}]{
  \includegraphics[width=30mm]{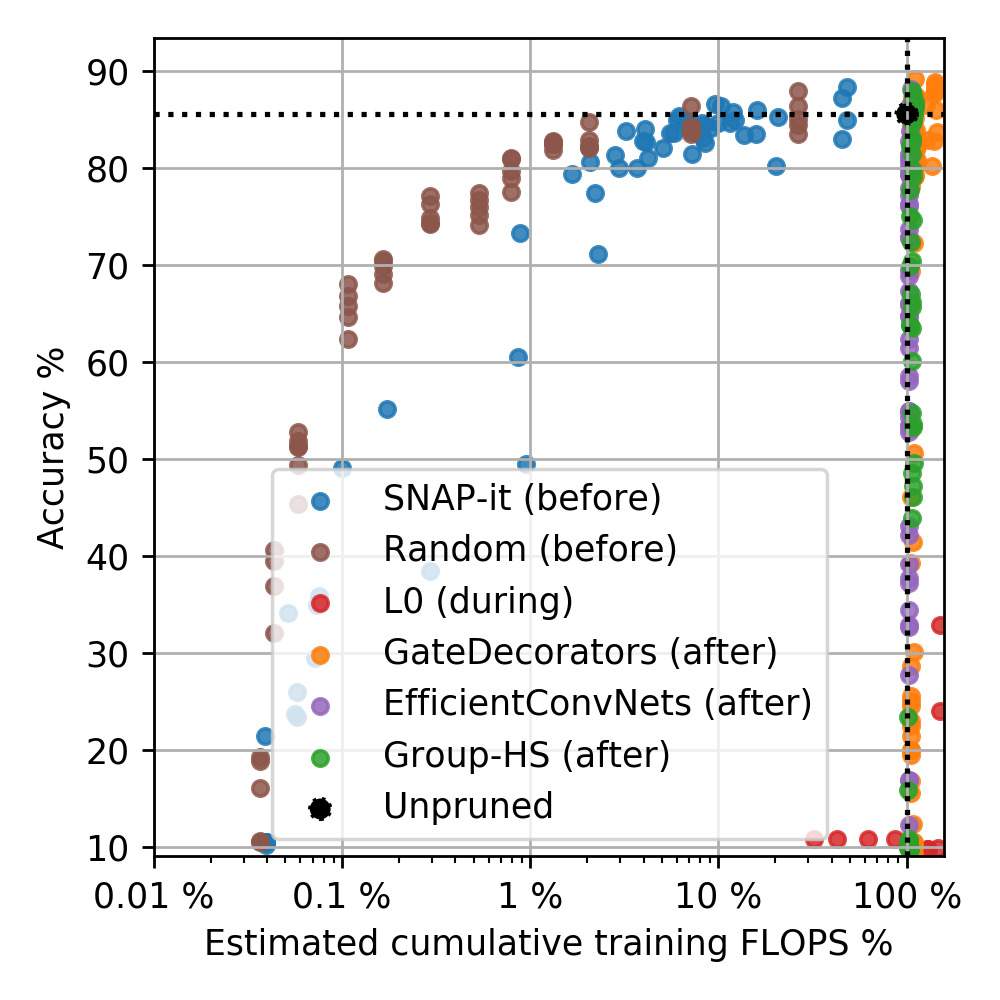}
}
\caption{\textit{Estimations of total cumulative FLOPS of the training process for a set number of epochs for multiple network-dataset combinations w.r.t. unpruned baseline. GateDecorators \citep{you2019gate} and $\ell_0$-regularisation \citep{louizos2017learning} required a higher number of epochs and naturally have a higher total because of it.}}
\end{figure}
\begin{figure}[h!]
\centering
\subfloat[\textit{AlexNet-CIFAR10}]{
  \includegraphics[width=30mm]{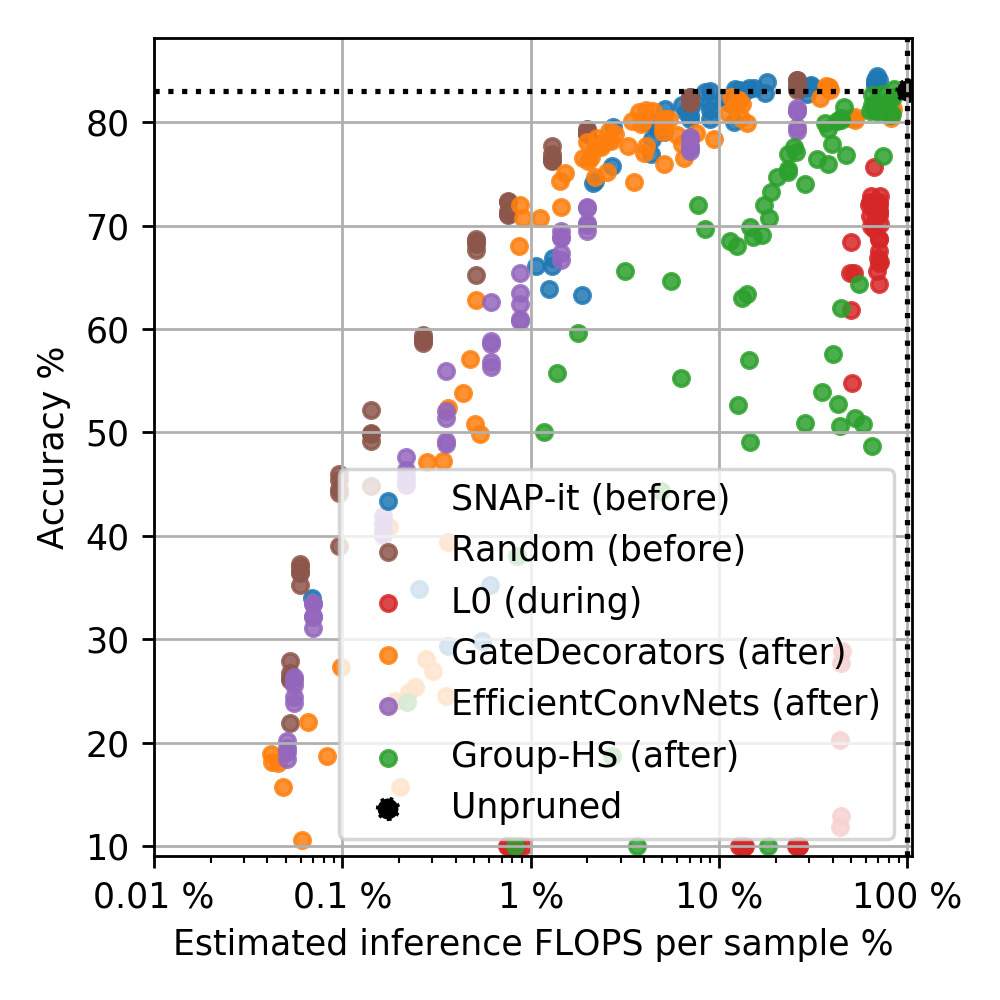}
}
\subfloat[\textit{AlexNet-Imagnette}]{
  \includegraphics[width=30mm]{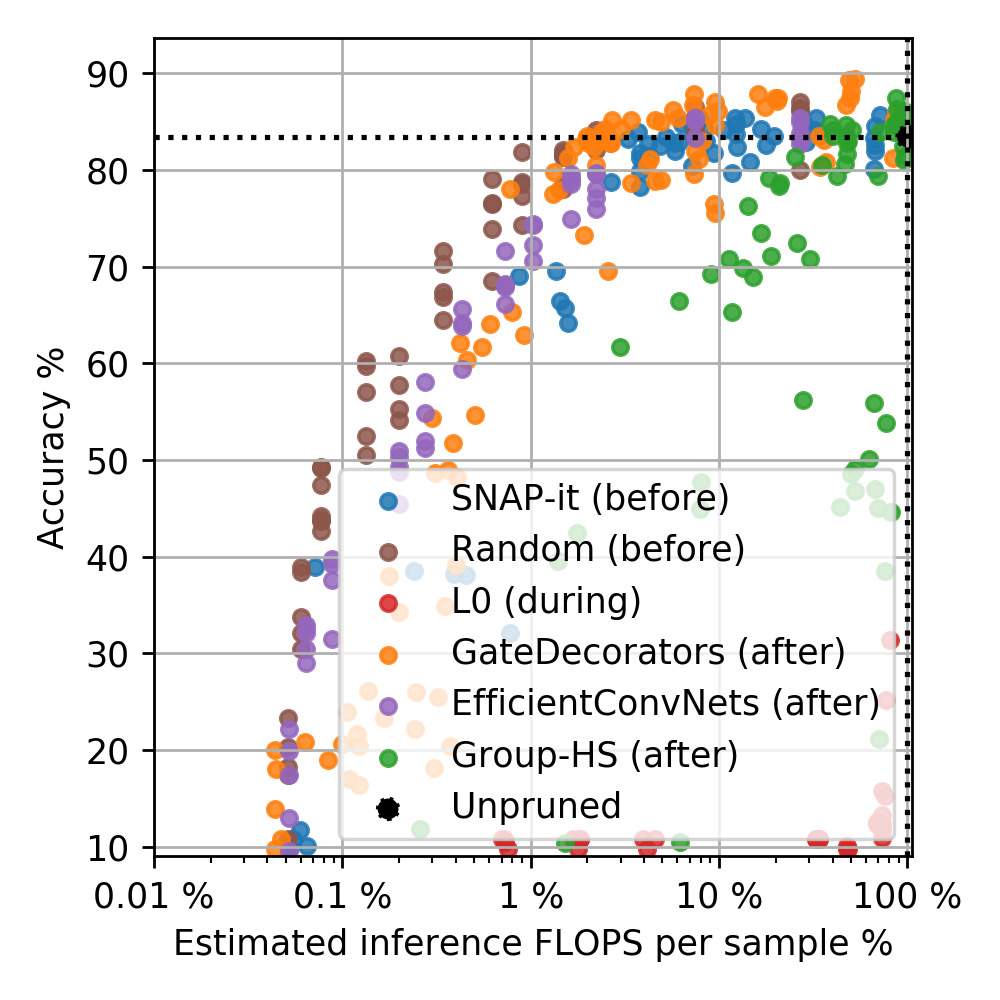}
}
\subfloat[\textit{VGG16-CIFAR10}]{
  \includegraphics[width=30mm]{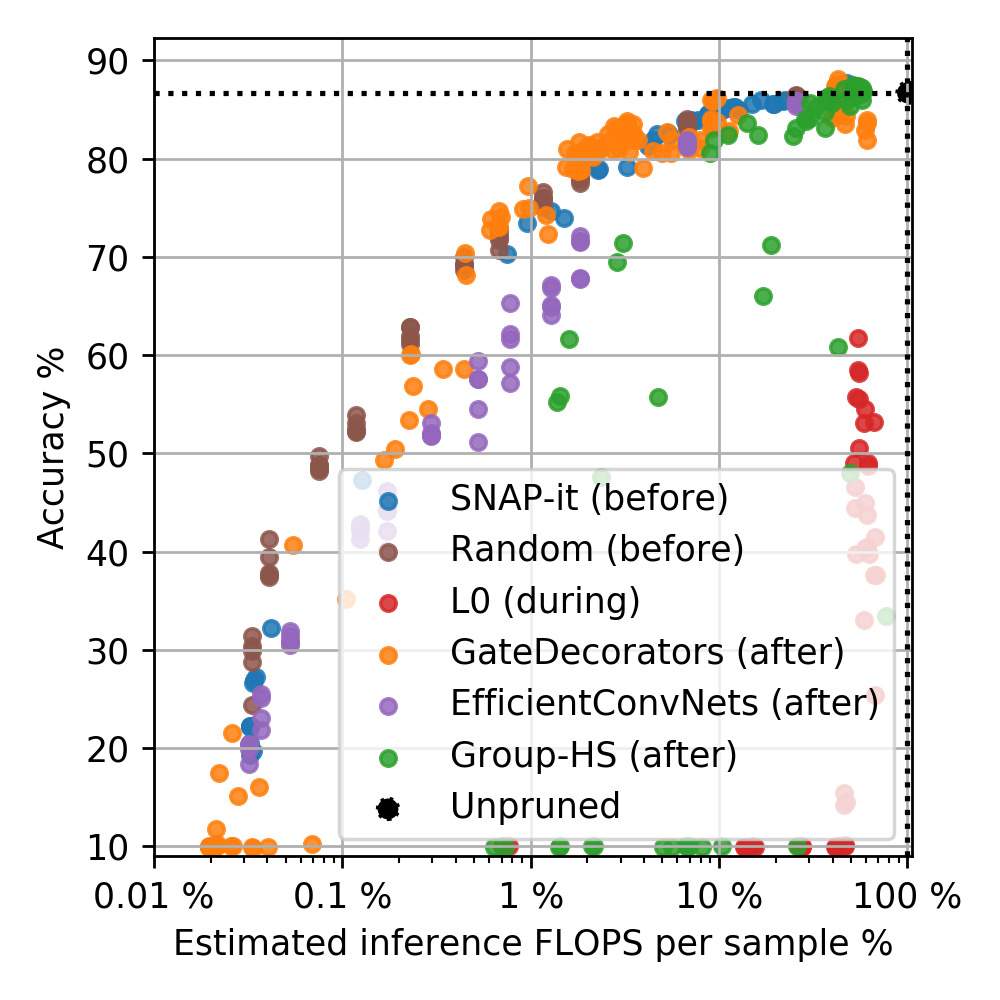}
}
\subfloat[\textit{VGG16-Imagnette}]{
  \includegraphics[width=30mm]{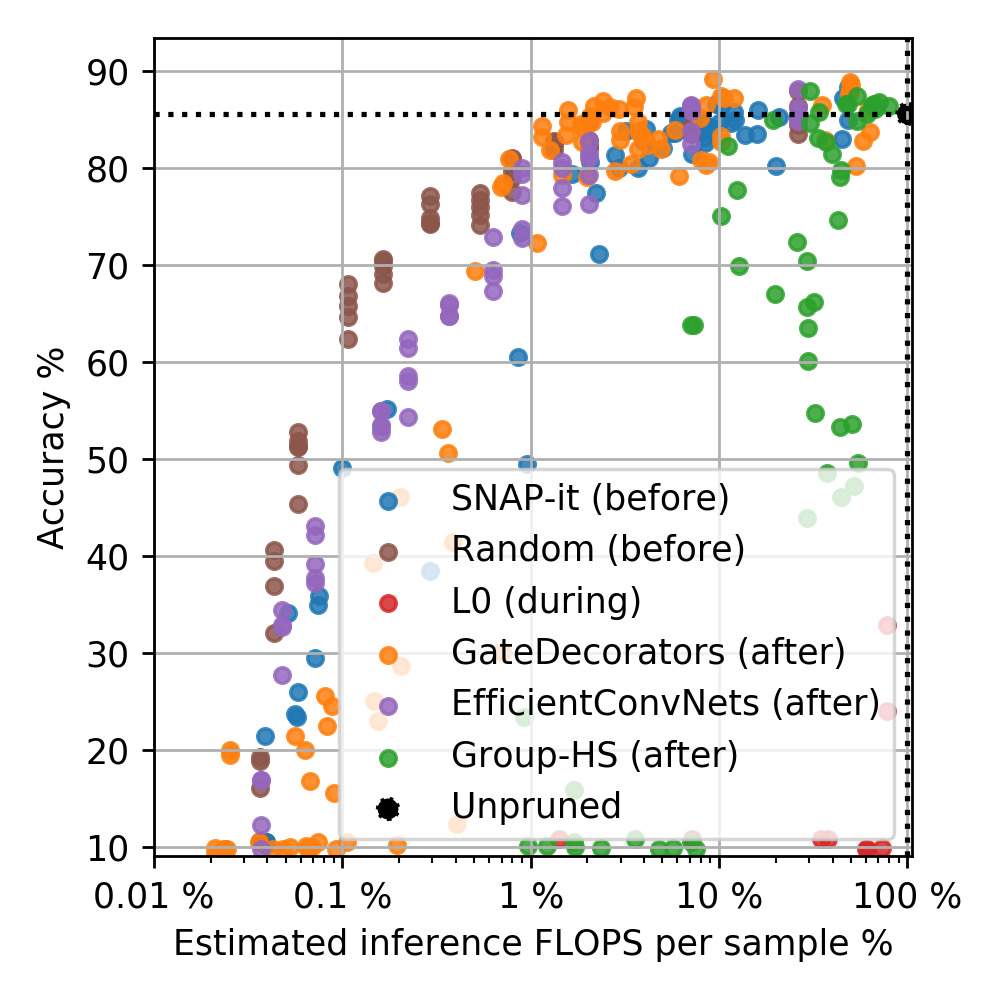}
}
\caption{\textit{Estimation of inference FLOPS after training for multiple network-dataset combinations w.r.t. unpruned baseline.}}
\end{figure}
\FloatBarrier
\newpage
\subsection{Disk}
\FloatBarrier
\begin{figure}[h!]
\centering
\subfloat[\textit{AlexNet-CIFAR10}]{
  \includegraphics[width=30mm]{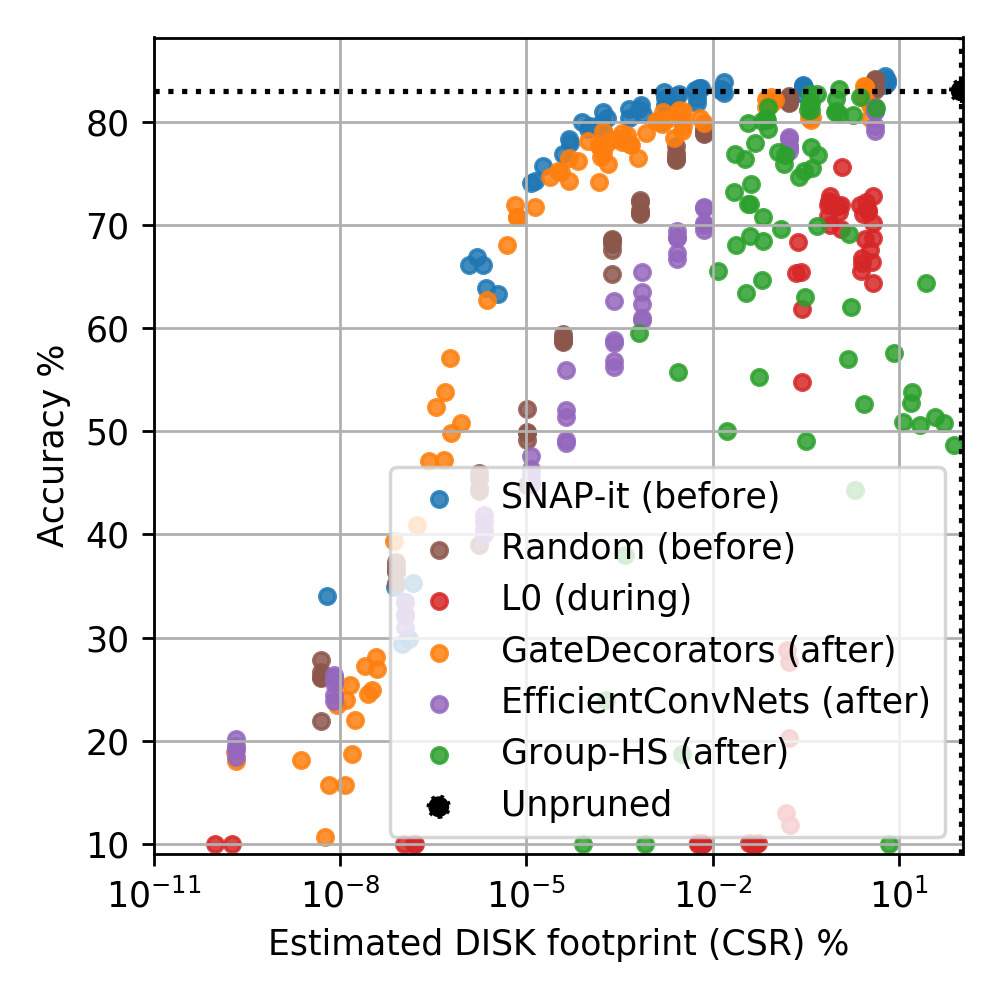}
}
\subfloat[\textit{AlexNet-Imagnette}]{
  \includegraphics[width=30mm]{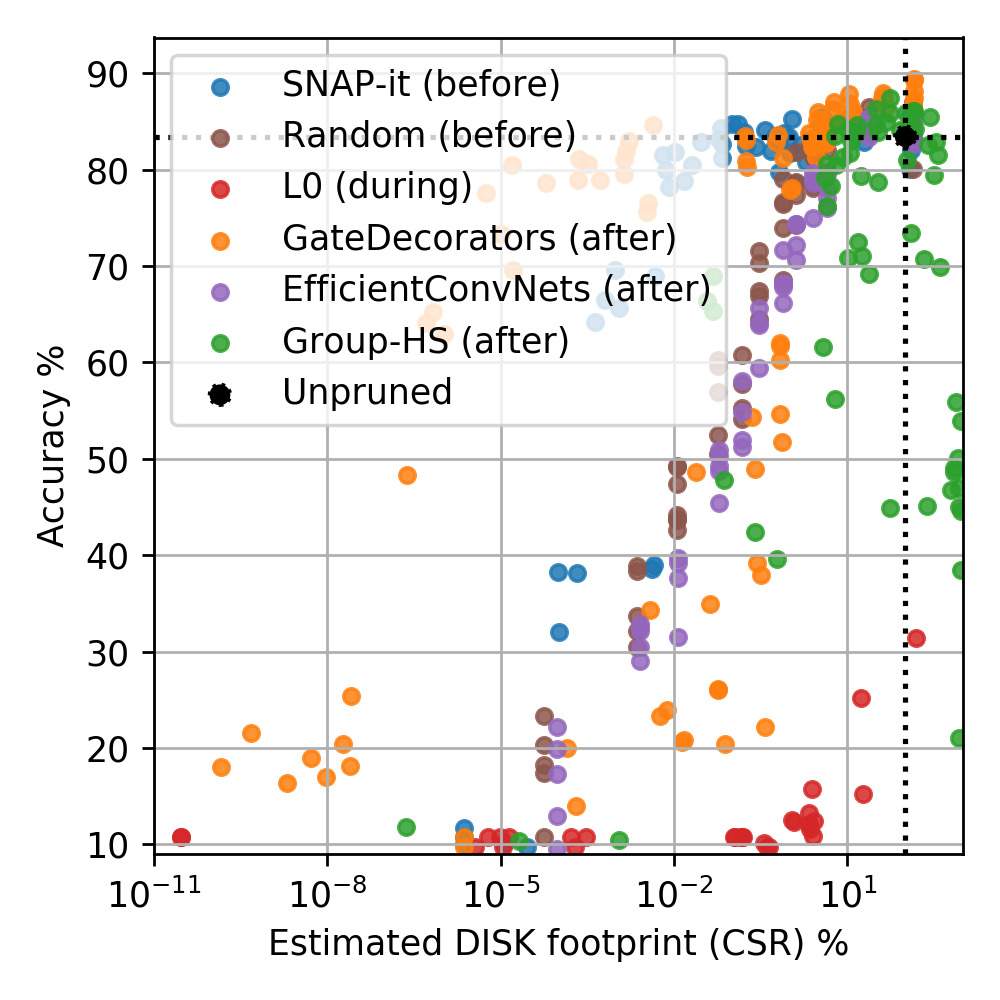}
}
\subfloat[\textit{VGG16-CIFAR10}]{
  \includegraphics[width=30mm]{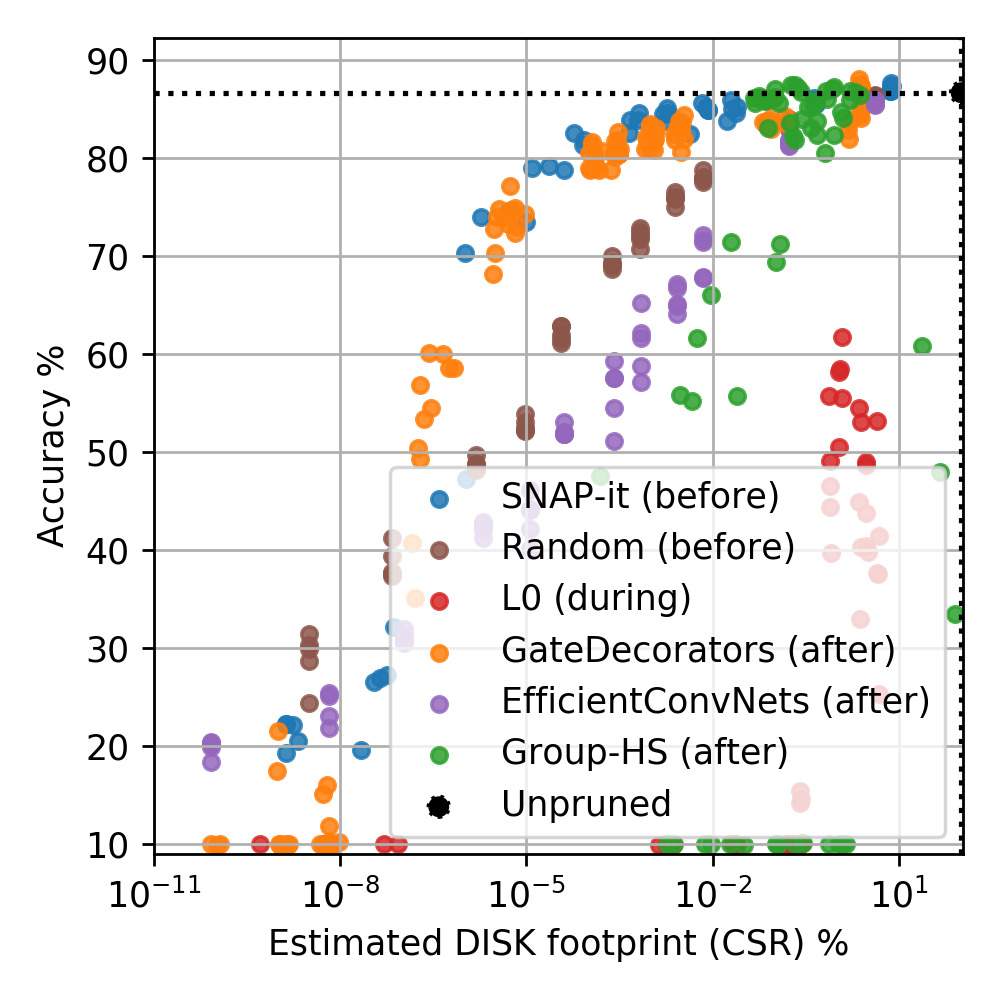}
}
\subfloat[\textit{VGG16-Imagnette}]{
  \includegraphics[width=30mm]{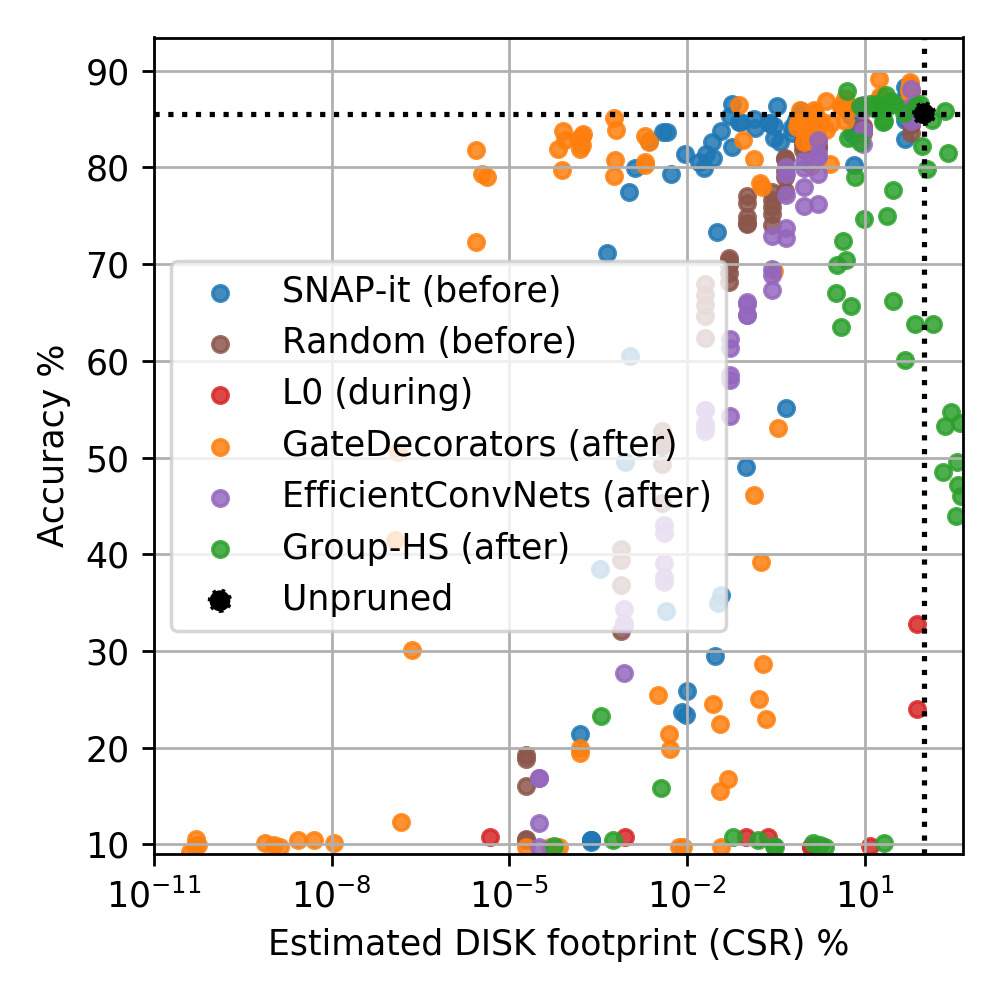}
}
\hspace{0mm}
\subfloat[\textit{Conv6-CIFAR10}]{
  \includegraphics[width=30mm]{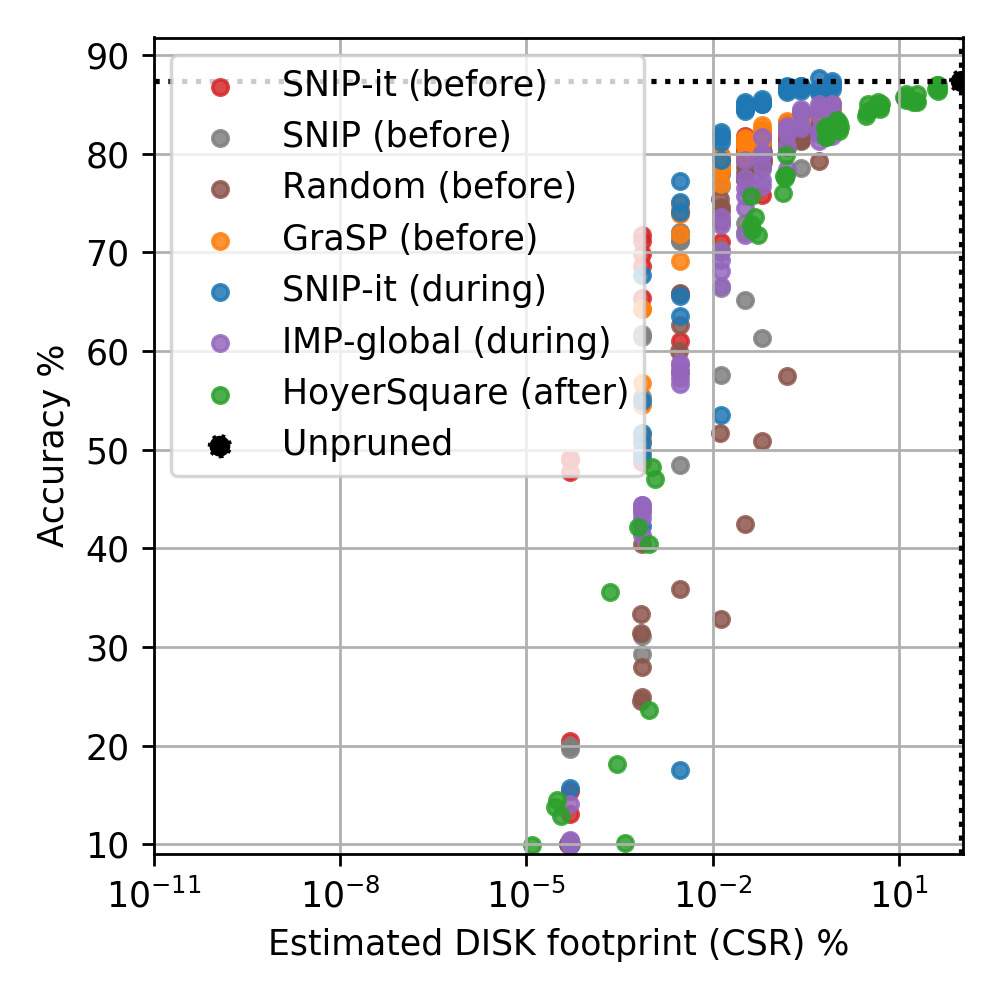}
}
\subfloat[\textit{Conv6-Imagenette}]{
  \includegraphics[width=30mm]{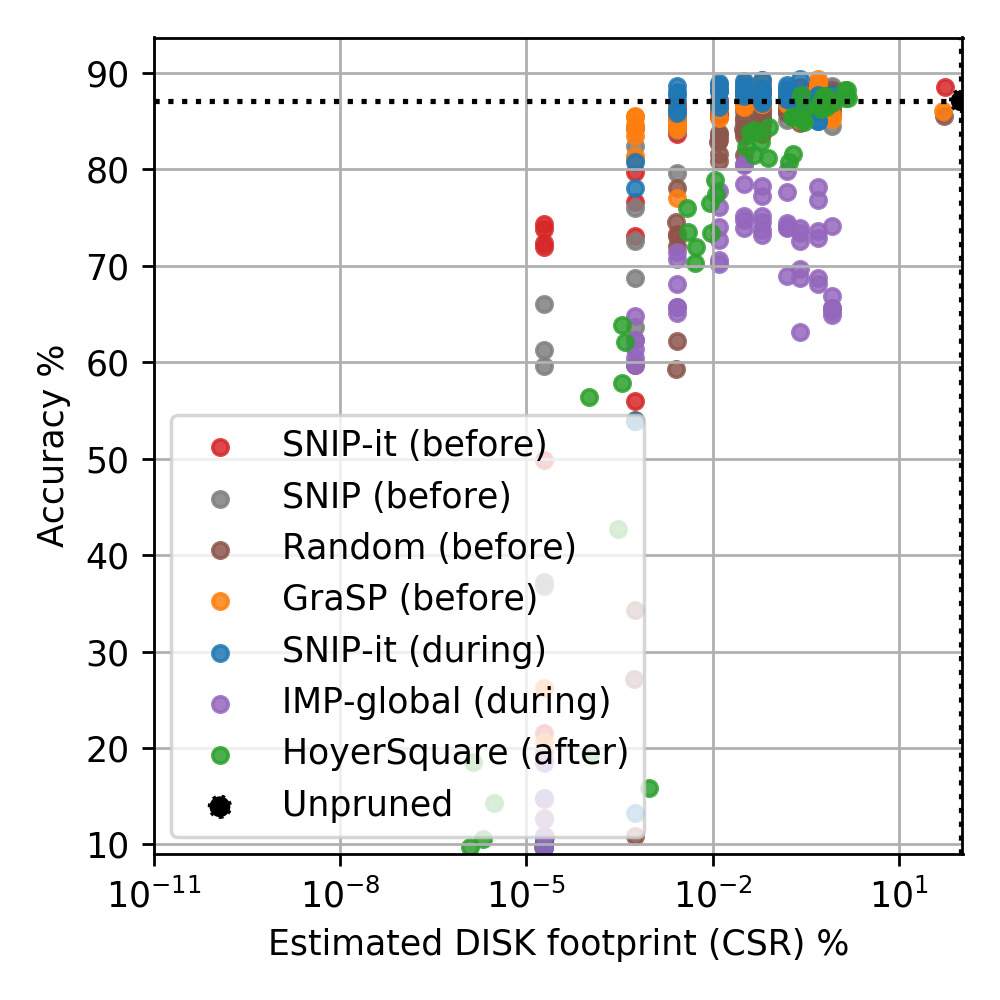}
}
\subfloat[\textit{LeNet5-CIFAR10}]{
  \includegraphics[width=30mm]{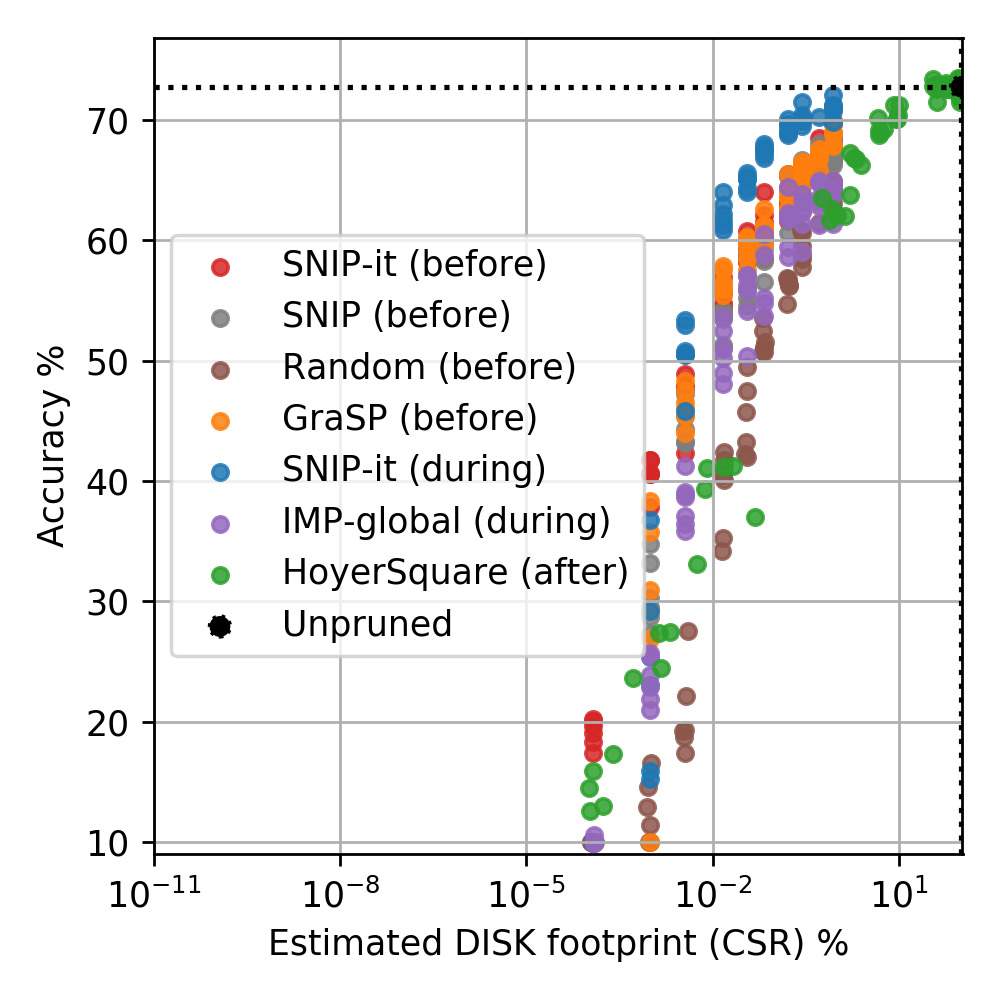}
}
\subfloat[\textit{LeNet5-Imagenette}]{
  \includegraphics[width=30mm]{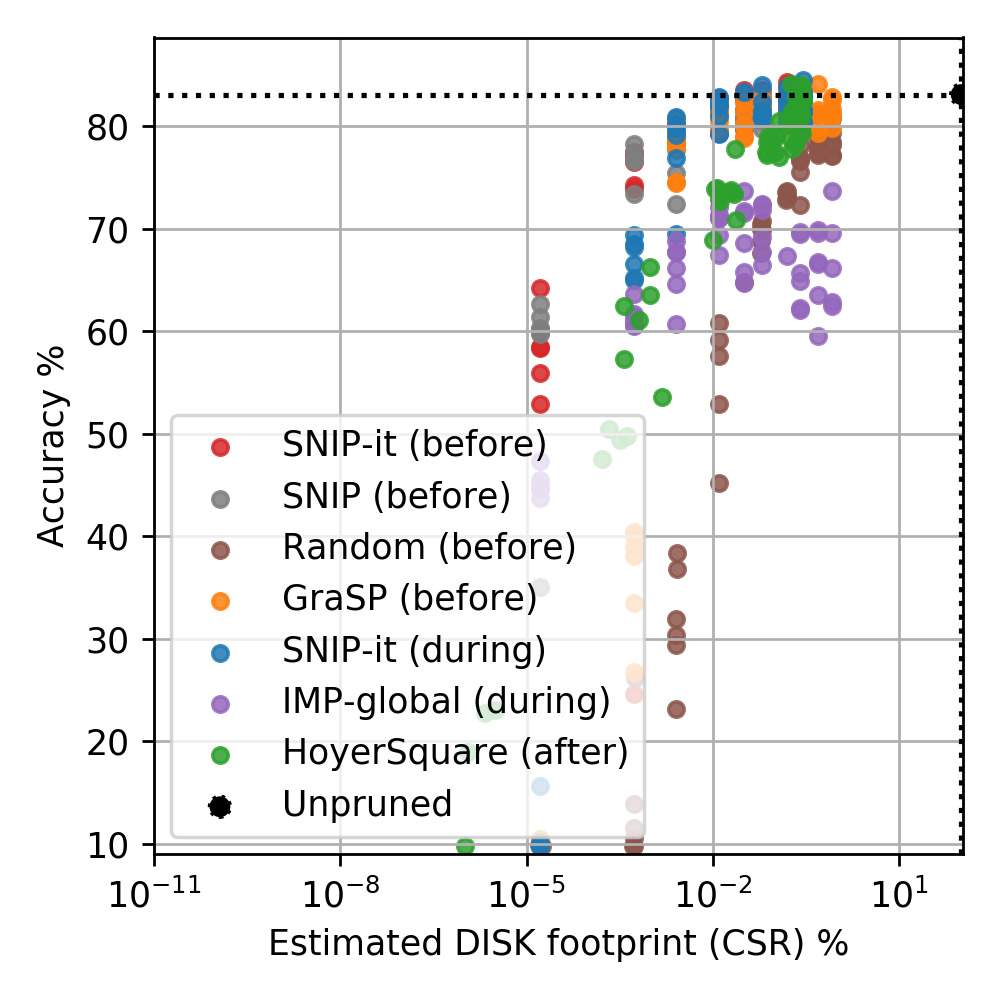}
}
\hspace{0mm}
\subfloat[\textit{MLP5-CIFAR10}]{   
  \includegraphics[width=30mm]{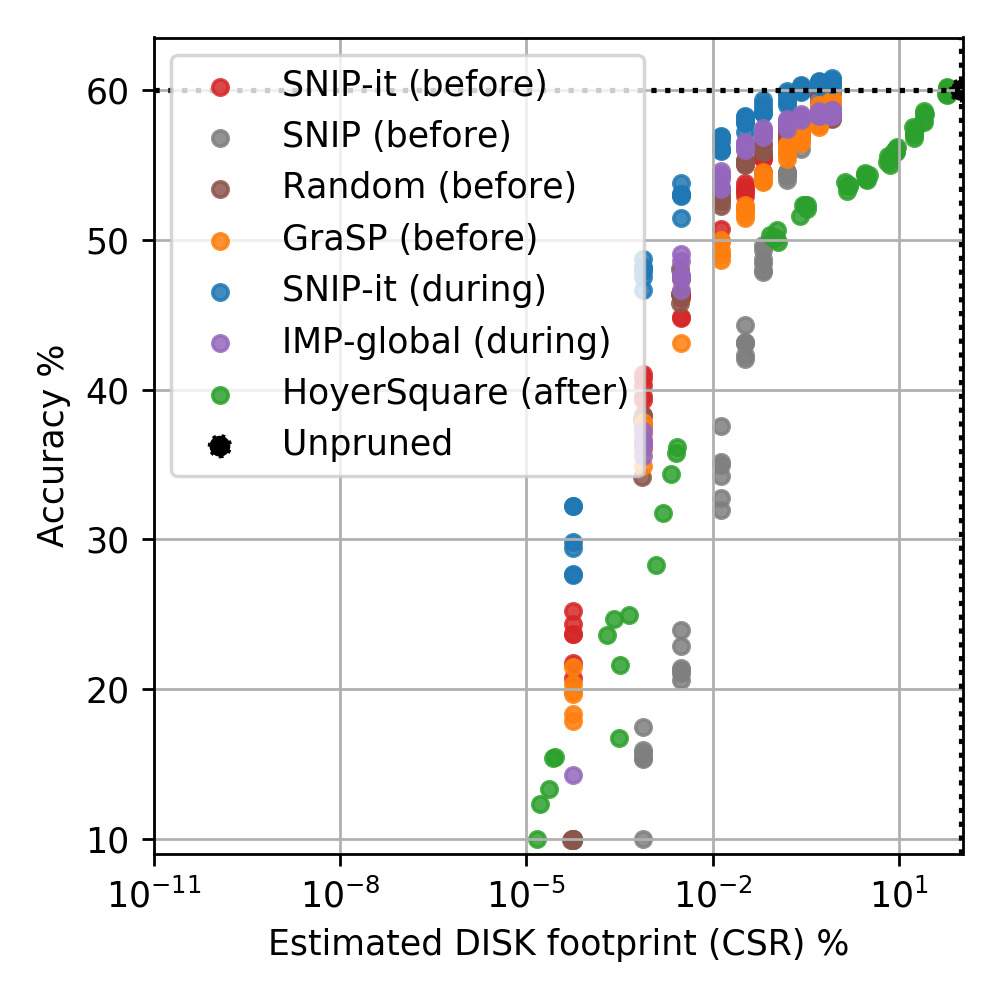}
}
\subfloat[\textit{MLP5-Imagenette}]{
  \includegraphics[width=30mm]{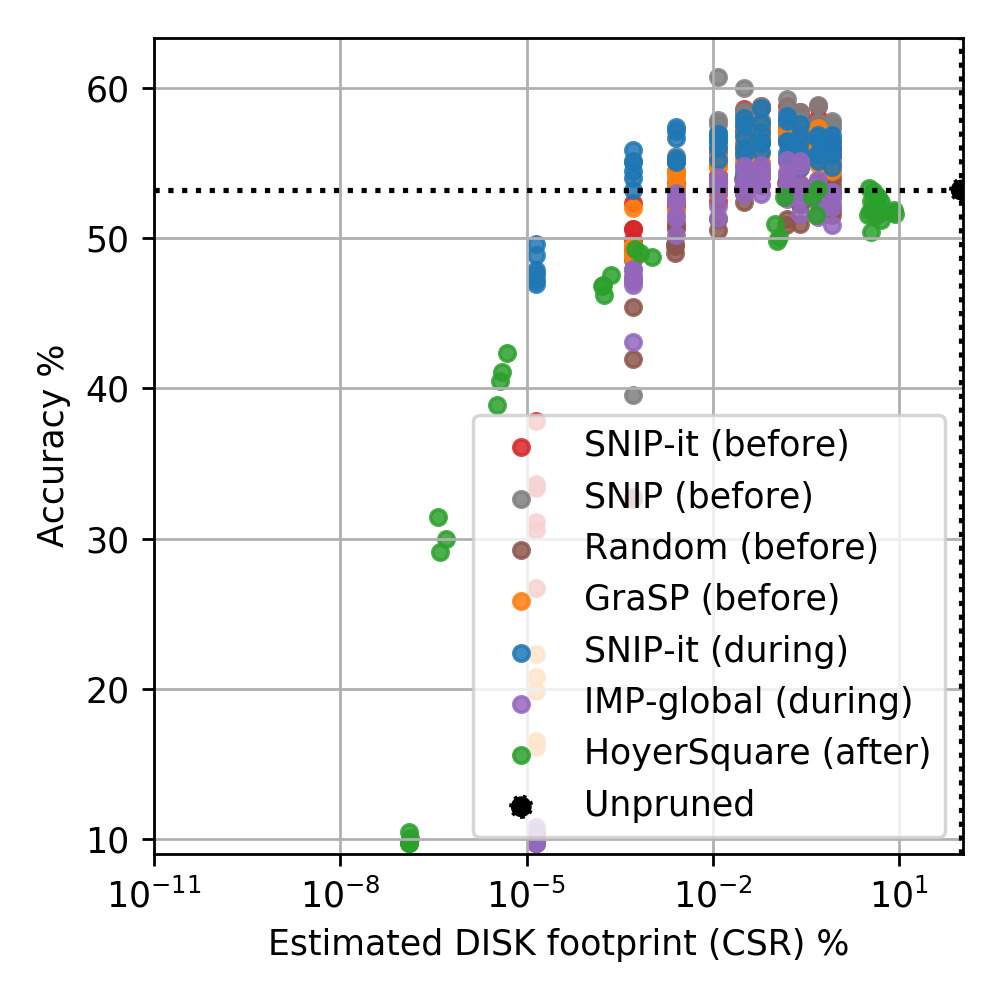}
}
\subfloat[\textit{ResNet18-CIFAR10}]{
  \includegraphics[width=30mm]{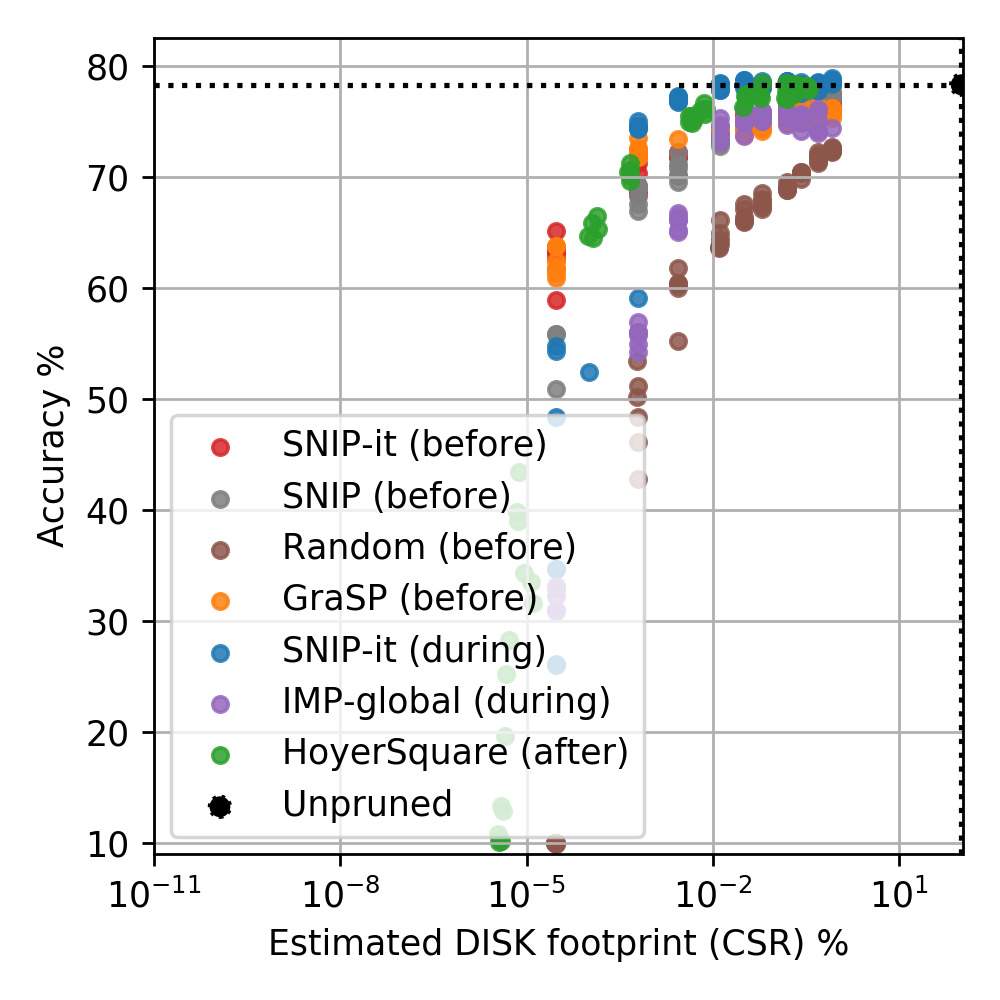}
}
\subfloat[\textit{ResNet18-Imagenette}]{
  \includegraphics[width=30mm]{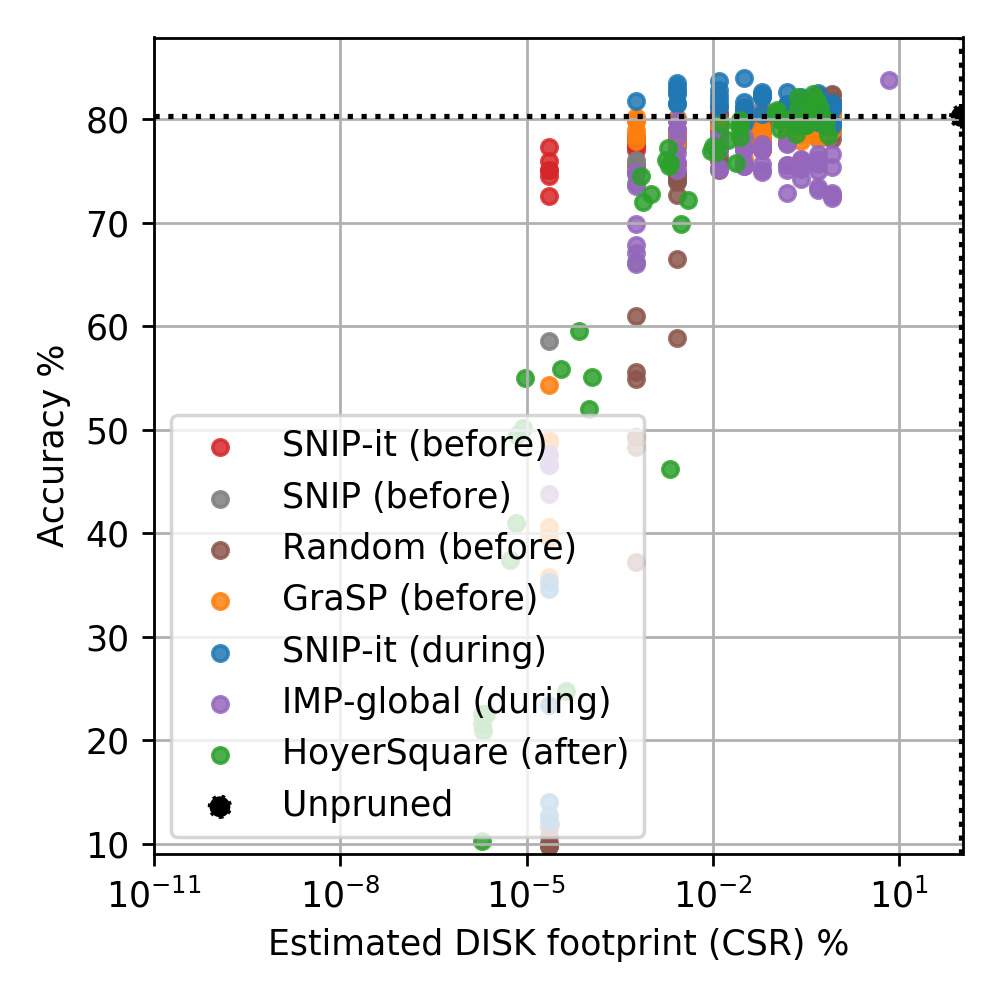}
}
\caption{\textit{Estimations of Disk storage for multiple network-dataset combinations as reduction in CSR-format \citep{bulucc2009parallel, saad2003iterative} w.r.t. unpruned baseline.}}
\end{figure}
\FloatBarrier

\subsection{Time}
\FloatBarrier
\begin{figure}[h!]
\centering
\subfloat[\textit{AlexNet-CIFAR10}]{
  \includegraphics[width=30mm]{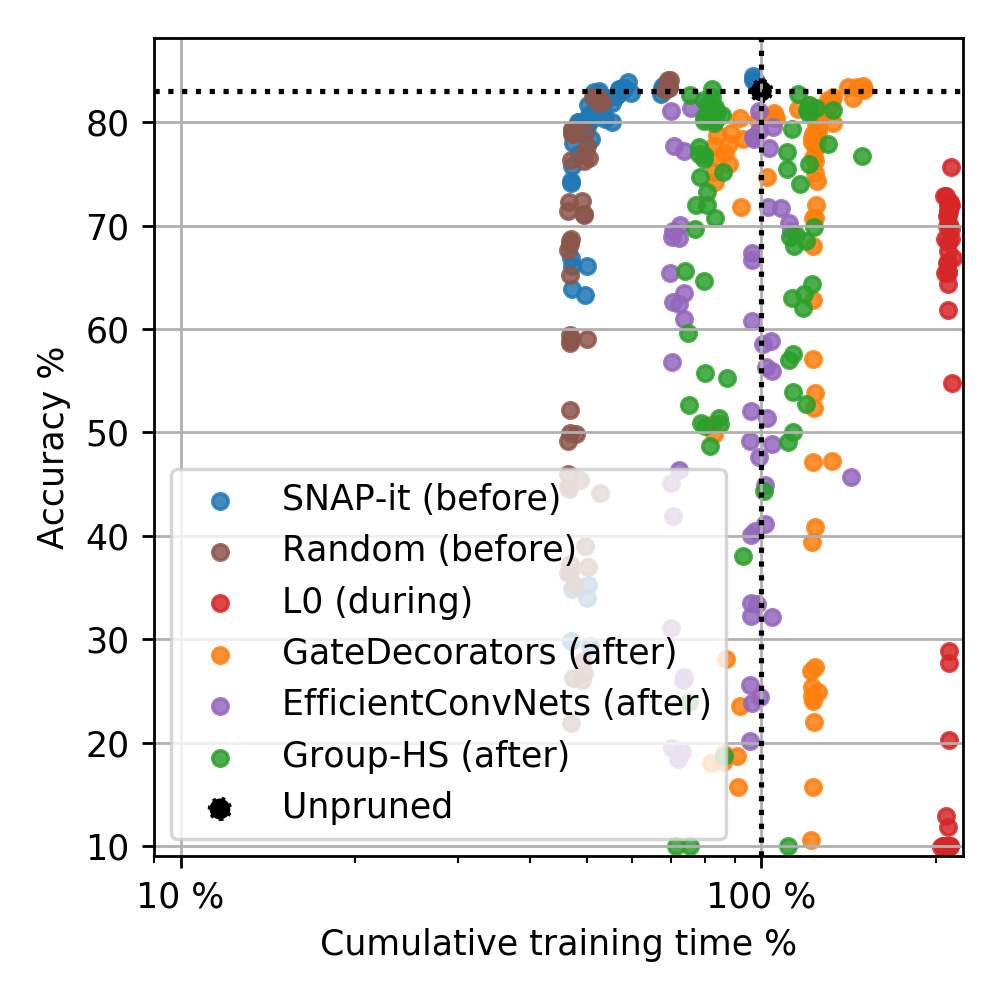}
}
\subfloat[\textit{AlexNet-Imagnette}]{
  \includegraphics[width=30mm]{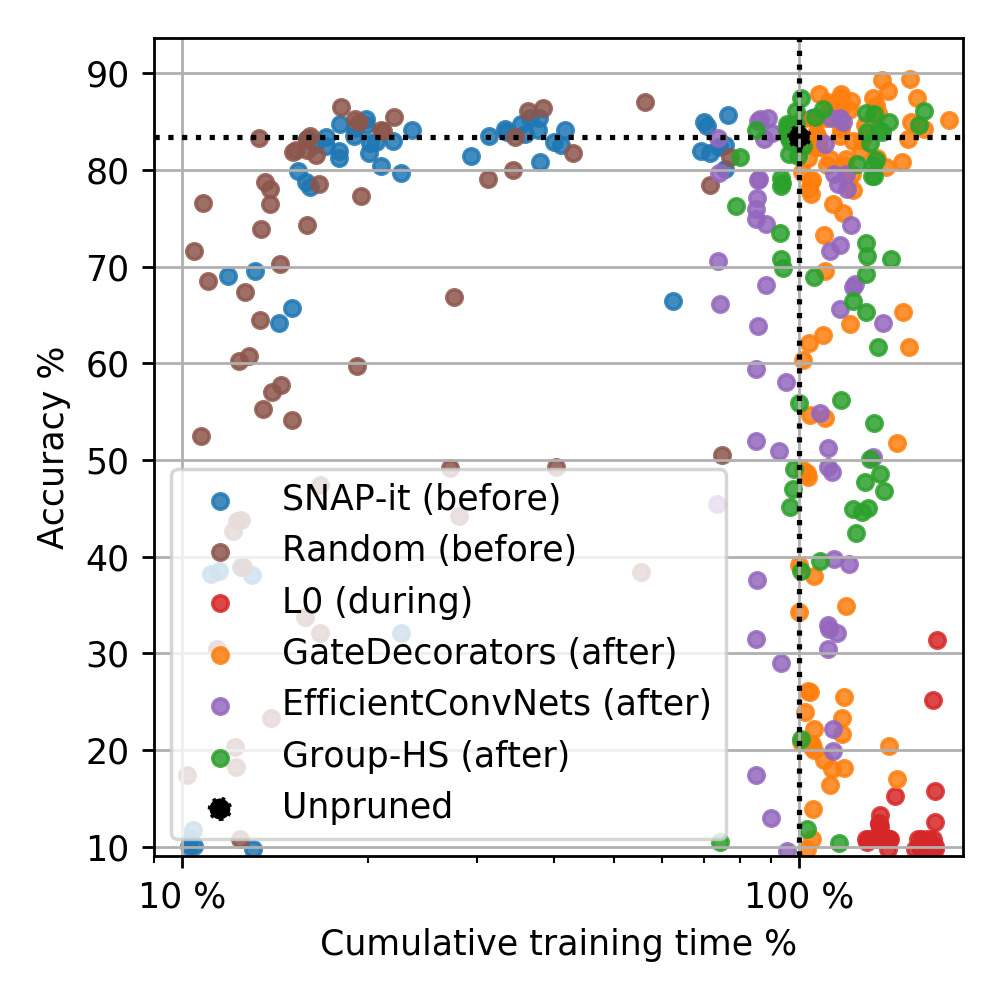}
}
\subfloat[\textit{VGG16-CIFAR10}]{
  \includegraphics[width=30mm]{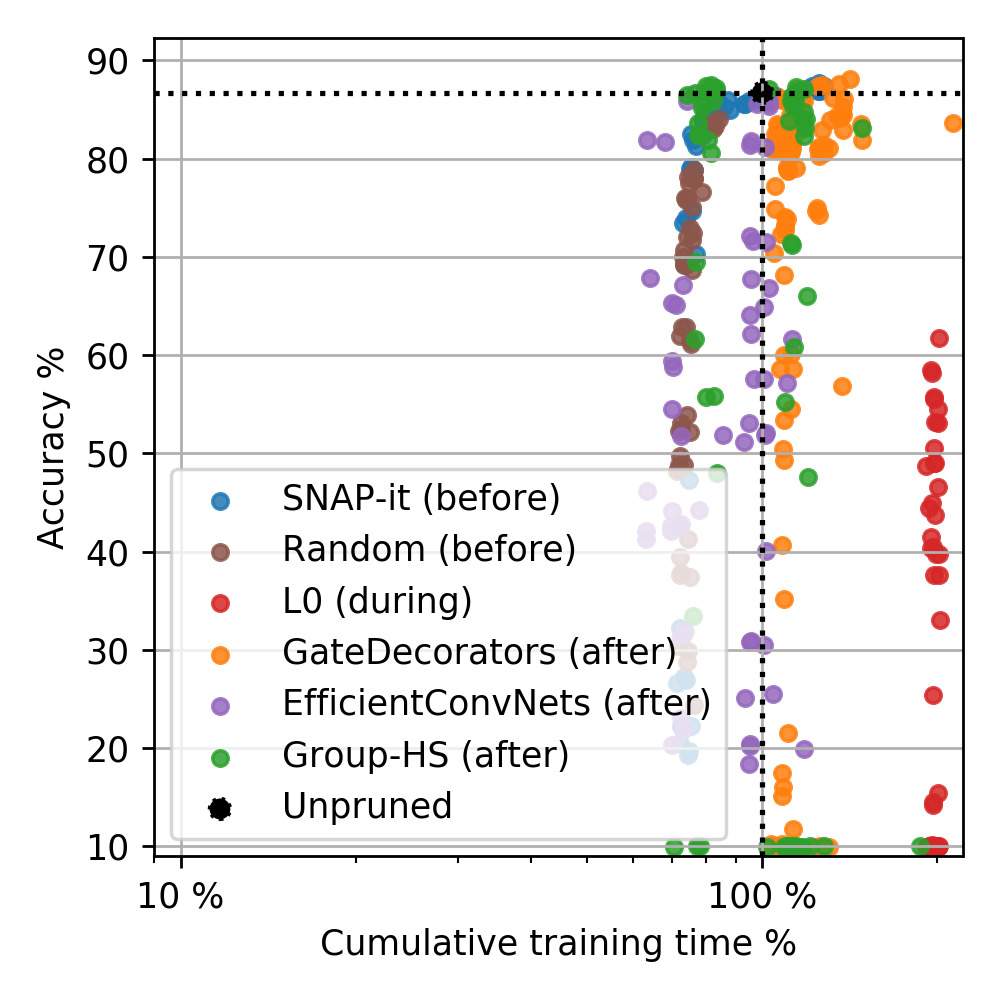}
}
\subfloat[\textit{VGG16-Imagnette}]{
  \includegraphics[width=30mm]{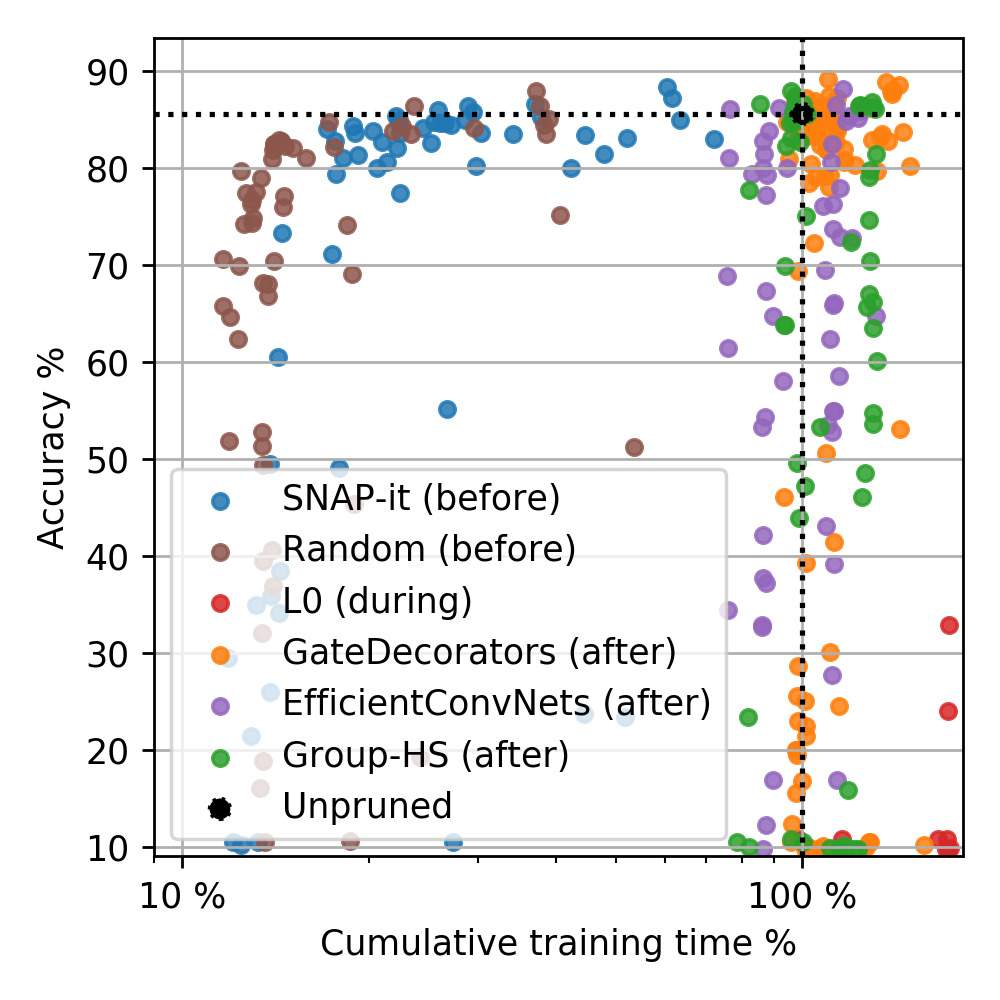}
}
\caption{\textit{Measurements of total run-time for a set number of epochs for multiple network-dataset combinations w.r.t. unpruned baseline. GateDecorators \citep{you2019gate} and $\ell_0$-regularisation \citep{louizos2017learning} required a higher number of epochs and naturally have higher measurements because of it.}}
\end{figure}
\begin{figure}[h!]
\centering
\subfloat[\textit{AlexNet-CIFAR10}]{
  \includegraphics[width=30mm]{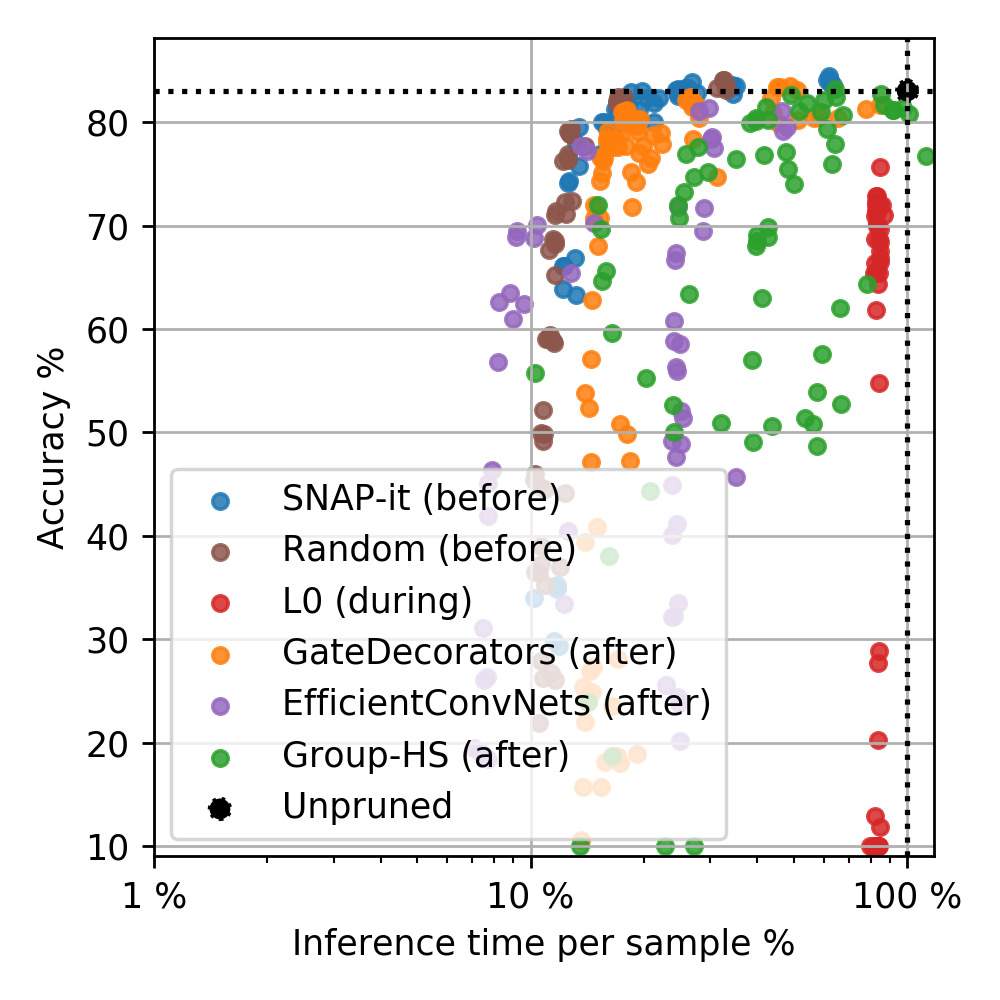}
}
\subfloat[\textit{AlexNet-Imagnette}]{
  \includegraphics[width=30mm]{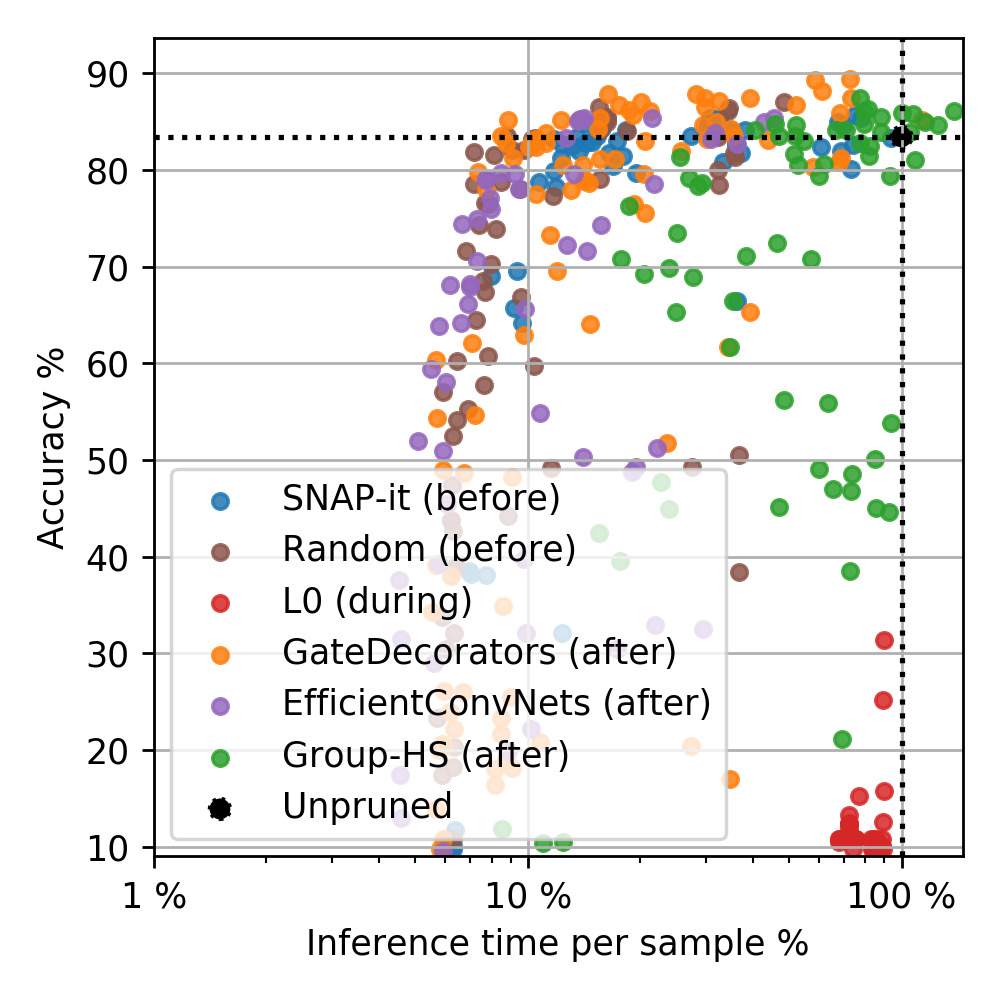}
}
\subfloat[\textit{VGG16-CIFAR10}]{
  \includegraphics[width=30mm]{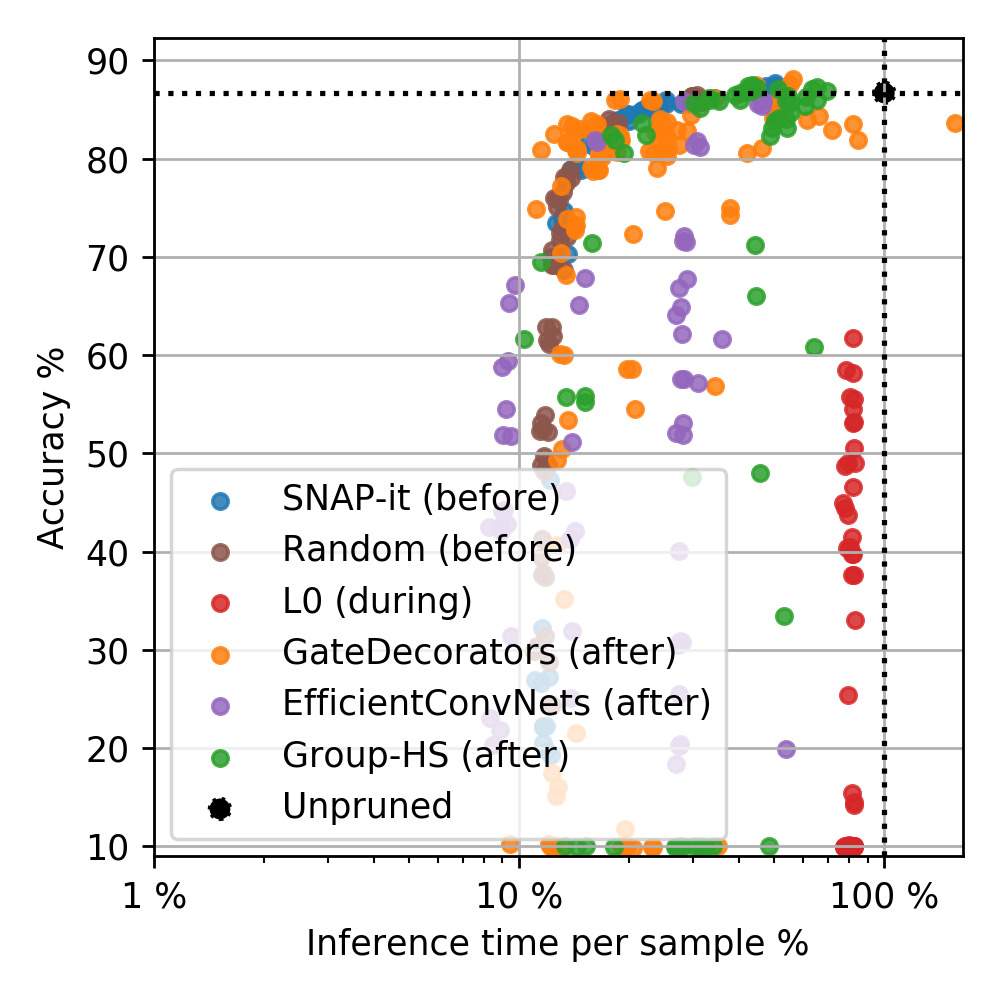}
}
\subfloat[\textit{VGG16-Imagnette}]{
  \includegraphics[width=30mm]{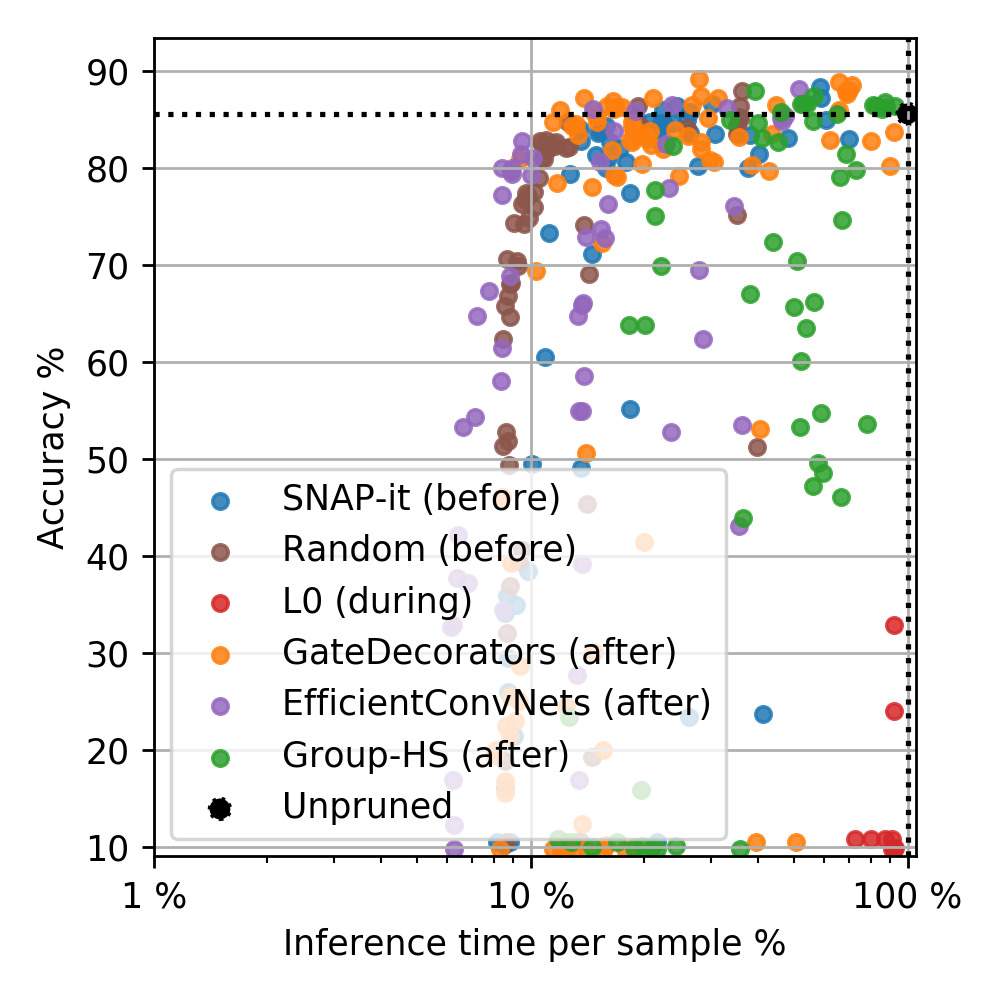}
}
\caption{\textit{Measurements of inference time after training for multiple network-dataset combinations w.r.t. unpruned baseline.}}
\end{figure}
\FloatBarrier
\subsection{RAM-footprint}
\FloatBarrier
\begin{figure}[h!]
\centering
\subfloat[\textit{AlexNet-CIFAR10}]{
  \includegraphics[width=30mm]{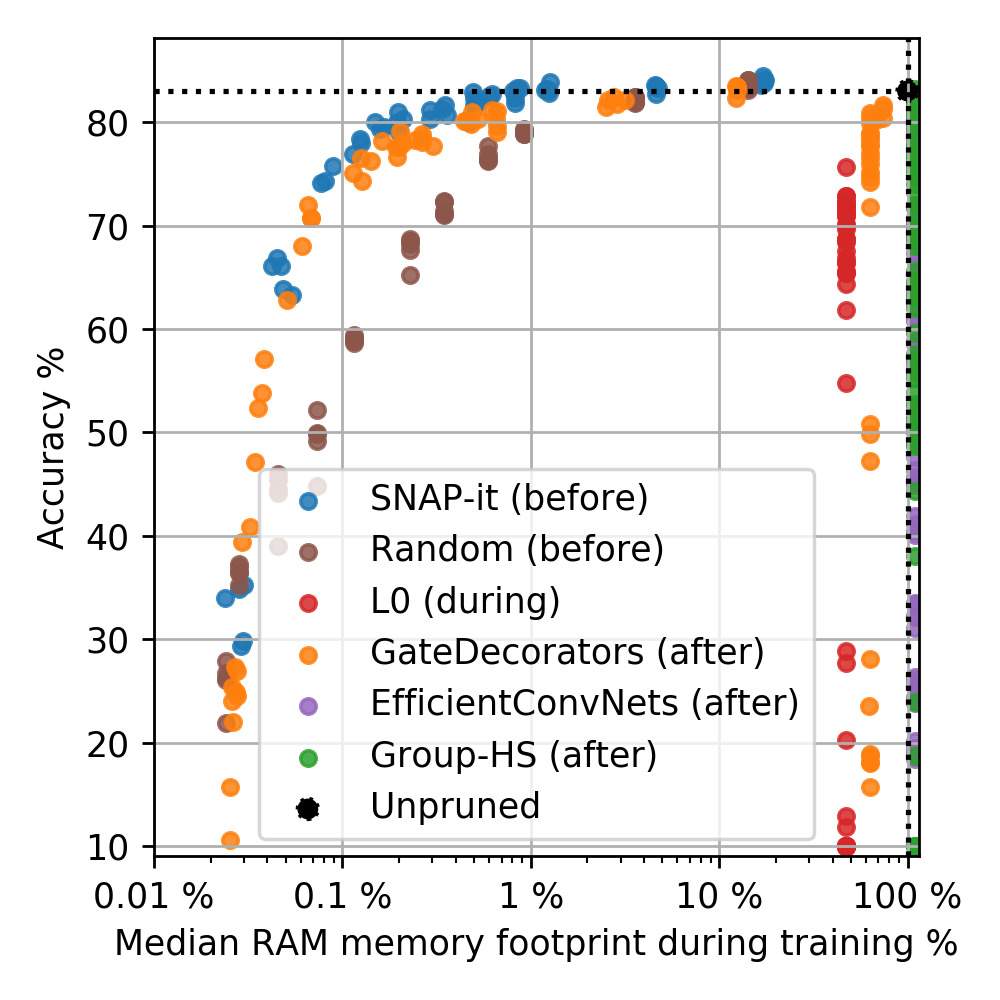}
}
\subfloat[\textit{AlexNet-Imagnette}]{
  \includegraphics[width=30mm]{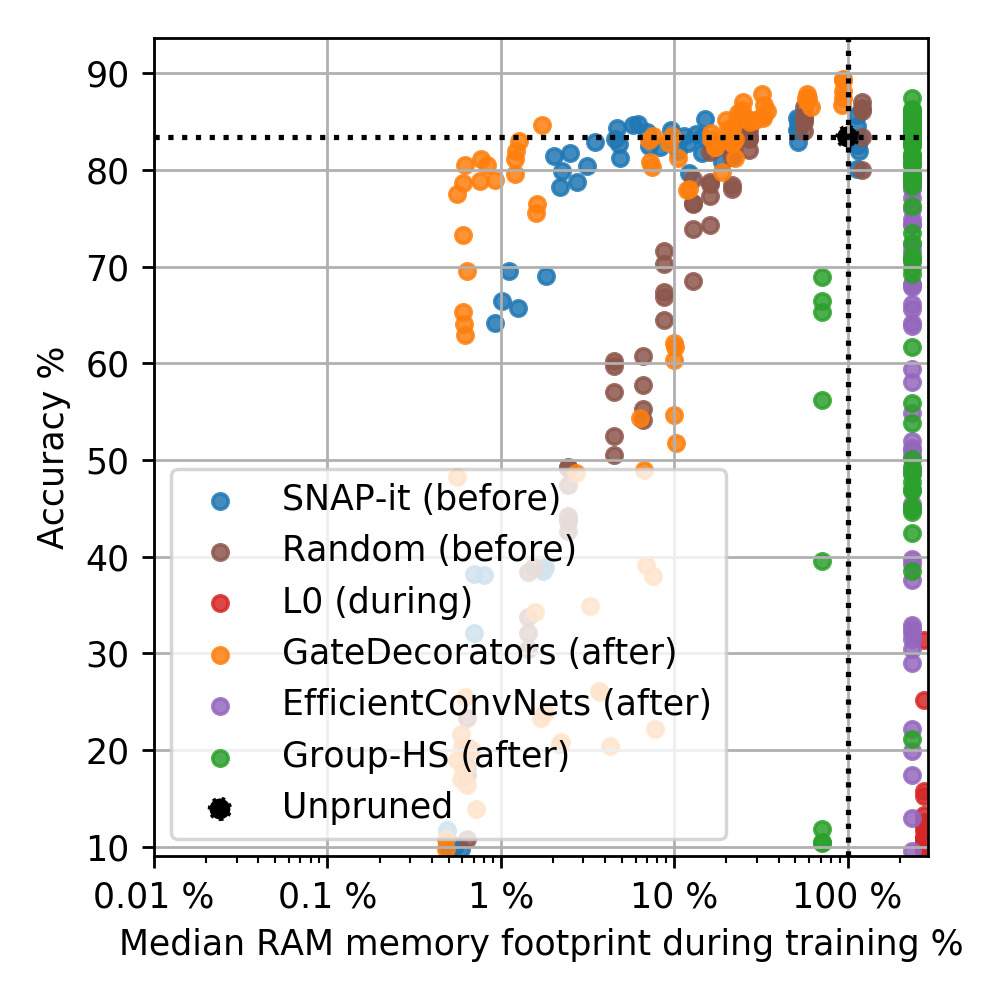}
}
\subfloat[\textit{VGG16-CIFAR10}]{
  \includegraphics[width=30mm]{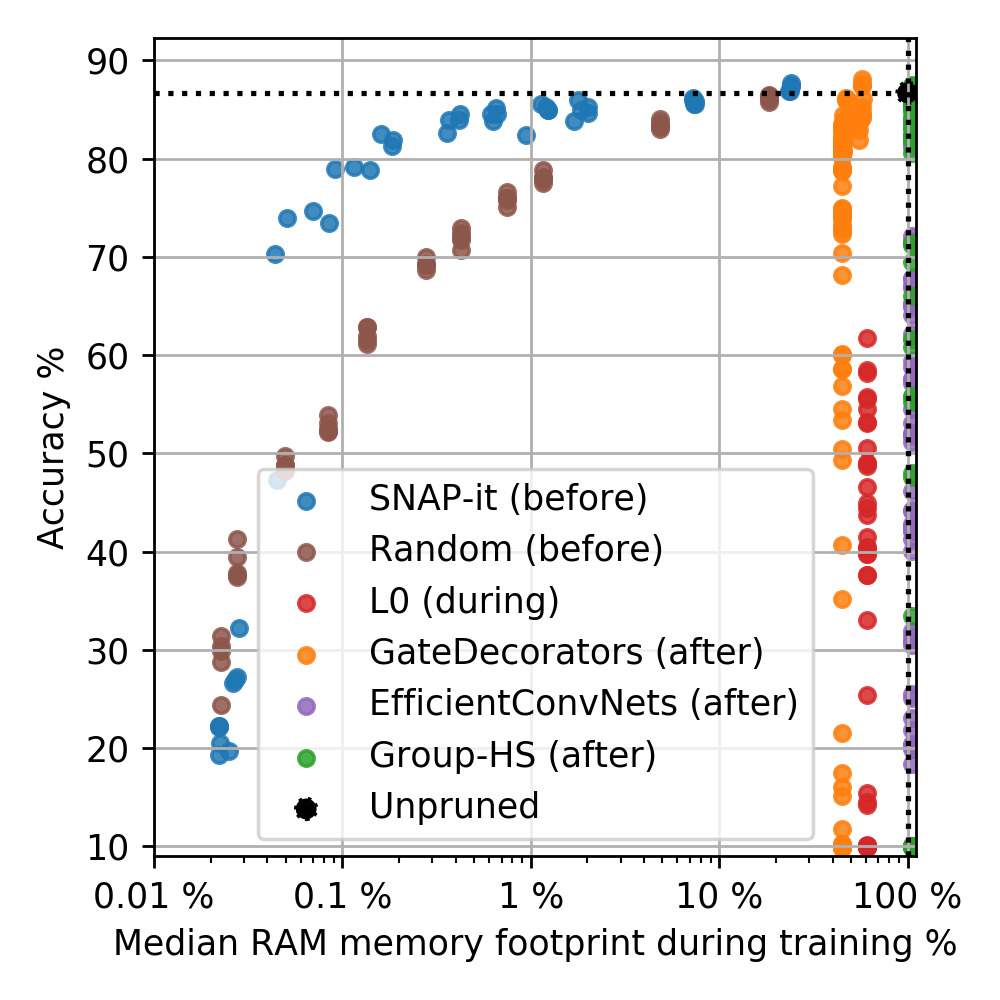}
}
\subfloat[\textit{VGG16-Imagnette}]{
  \includegraphics[width=30mm]{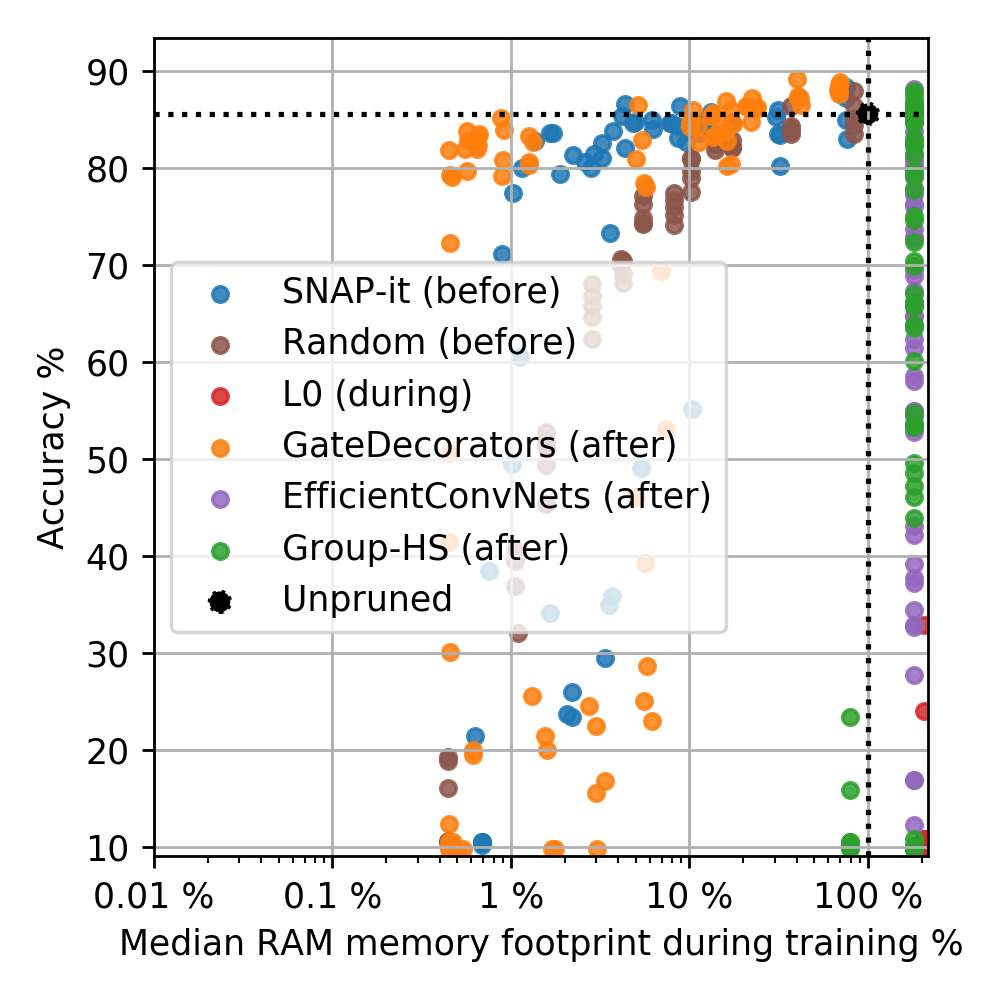}
}
\caption{\textit{Median measurements of RAM-footprint during training. Note that for GateDecorators \citep{you2019gate}, since it still trains rather long after the pre-train phase, sometimes a sparser measurement ended up being the median and sometimes one from the the pre-train phase, causing two separate patterns to emerge.}}
\end{figure}
\FloatBarrier

\section{Network architectures}
\label{sec:append:networks}

Here different network architecture details are documented. $I$ will denote the input dimensions and $K$ the number of classes. We alter most networks to use the LeakyReLU activation function with a slope of 5\%, dropout \citep{srivastava2014dropout} with a probability of 30\%, batch normalisation and pooling layers. \\

\footnotesize

\begin{itemize}
    \item \textbf{MLP5} \footnote{Own creation} \newline
   Linear(in features=I, out features=512, bias=True) \newline BatchNorm1d(512, eps=1e-05, momentum=0.1, affine=True, track running stats=True) \newline LeakyReLU(negative slope=0.05) \newline Dropout(p=0.3, inplace=False) \newline Linear(in features=512, out features=512, bias=True) \newline BatchNorm1d(512, eps=1e-05, momentum=0.1, affine=True, track running stats=True) \newline LeakyReLU(negative slope=0.05) \newline Dropout(p=0.3, inplace=False) \newline Linear(in features=512, out features=512, bias=True) \newline BatchNorm1d(512, eps=1e-05, momentum=0.1, affine=True, track running stats=True) \newline LeakyReLU(negative slope=0.05) \newline Dropout(p=0.3, inplace=False) \newline Linear(in features=512, out features=512, bias=True) \newline BatchNorm1d(512, eps=1e-05, momentum=0.1, affine=True, track running stats=True) \newline LeakyReLU(negative slope=0.05) \newline Dropout(p=0.3, inplace=False) \newline Linear(in features=512, out features=K, bias=True)  

    \item \textbf{LeNet5} \newline 
Conv2d(I, 6, kernel size=(5, 5), stride=(1, 1), padding=(2, 2)) \newline BatchNorm2d(6, eps=1e-05, momentum=0.1, affine=True, track running stats=True) \newline LeakyReLU(negative slope=0.05) \newline MaxPool2d(kernel size=2, stride=2, padding=0, dilation=1, ceil mode=False) \newline Conv2d(6, 16, kernel size=(5, 5), stride=(1, 1)) \newline BatchNorm2d(16, eps=1e-05, momentum=0.1, affine=True, track running stats=True) \newline LeakyReLU(negative slope=0.05) \newline MaxPool2d(kernel size=2, stride=2, padding=0, dilation=1, ceil mode=False) \newline Conv2d(16, 120, kernel size=(5, 5), stride=(1, 1)) \newline BatchNorm2d(120, eps=1e-05, momentum=0.1, affine=True, track running stats=True) \newline LeakyReLU(negative slope=0.05) \newline AdaptiveAvgPool2d(output size=(3, 3)) \newline Dropout(p=0.3, inplace=False) \newline Linear(in features=1080, out features=84, bias=True) \newline BatchNorm1d(84, eps=1e-05, momentum=0.1, affine=True, track running stats=True) \newline LeakyReLU(negative slope=0.05) \newline Dropout(p=0.3, inplace=False) \newline Linear(in features=84, out features=K, bias=True)  
    
    \item \textbf{Conv6} \newline
Conv2d(I, 64, kernel size=(3, 3), stride=(1, 1), padding=(1, 1)) \newline BatchNorm2d(64, eps=1e-05, momentum=0.1, affine=True, track running stats=True) \newline LeakyReLU(negative slope=0.05) \newline Conv2d(64, 64, kernel size=(3, 3), stride=(1, 1), padding=(1, 1)) \newline BatchNorm2d(64, eps=1e-05, momentum=0.1, affine=True, track running stats=True) \newline LeakyReLU(negative slope=0.05) \newline MaxPool2d(kernel size=2, stride=2, padding=0, dilation=1, ceil mode=False) \newline Conv2d(64, 128, kernel size=(3, 3), stride=(1, 1), padding=(1, 1)) \newline BatchNorm2d(128, eps=1e-05, momentum=0.1, affine=True, track running stats=True) \newline LeakyReLU(negative slope=0.05) \newline Conv2d(128, 128, kernel size=(3, 3), stride=(1, 1), padding=(1, 1)) \newline BatchNorm2d(128, eps=1e-05, momentum=0.1, affine=True, track running stats=True) \newline LeakyReLU(negative slope=0.05) \newline MaxPool2d(kernel size=2, stride=2, padding=0, dilation=1, ceil mode=False) \newline Conv2d(128, 256, kernel size=(3, 3), stride=(1, 1), padding=(1, 1)) \newline BatchNorm2d(256, eps=1e-05, momentum=0.1, affine=True, track running stats=True) \newline LeakyReLU(negative slope=0.05) \newline Conv2d(256, 256, kernel size=(3, 3), stride=(1, 1), padding=(1, 1)) \newline BatchNorm2d(256, eps=1e-05, momentum=0.1, affine=True, track running stats=True) \newline LeakyReLU(negative slope=0.05) \newline AdaptiveAvgPool2d(output size=(3, 3)) \newline Dropout(p=0.3, inplace=False) \newline Linear(in features=2304, out features=256, bias=True) \newline BatchNorm1d(256, eps=1e-05, momentum=0.1, affine=True, track running stats=True) \newline LeakyReLU(negative slope=0.05) \newline Dropout(p=0.3, inplace=False) \newline Linear(in features=256, out features=256, bias=True) \newline BatchNorm1d(256, eps=1e-05, momentum=0.1, affine=True, track running stats=True) \newline LeakyReLU(negative slope=0.05) \newline Dropout(p=0.3, inplace=False) \newline Linear(in features=256, out features=K, bias=True)  

    \item \textbf{ResNet18} \newline
    Conv2d(I, 64, kernel size=(7, 7), stride=(2, 2), padding=(3, 3), bias=False) \newline BatchNorm2d(64, eps=1e-05, momentum=0.1, affine=True, track running stats=True) \newline LeakyReLU(negative slope=0.05, inplace=True) \newline MaxPool2d(kernel size=3, stride=2, padding=1, dilation=1, ceil mode=False) \newline BasicBlock(in featurs=64, out features=64, downsampling=False) \newline BasicBlock(in featurs=64, out features=64, downsampling=False) \newline BasicBlock(in featurs=64, out features=128, downsampling=True) \newline BasicBlock(in featurs=128, out features=128, downsampling=False) \newline BasicBlock(in featurs=128, out features=256, downsampling=True) \newline BasicBlock(in featurs=256, out features=256, downsampling=False) \newline BasicBlock(in featurs=256, out features=512, downsampling=True) \newline BasicBlock(in featurs=512, out features=512, downsampling=False) \newline AdaptiveAvgPool2d(output size=(2, 2)) \newline Sequential( \newline Dropout(p=0.3, inplace=False) \newline Linear(in features=2048, out features=256, bias=True) \newline BatchNorm1d(256, eps=1e-05, momentum=0.1, affine=True, track running stats=True) \newline LeakyReLU(negative slope=0.05) \newline Dropout(p=0.3, inplace=False) \newline Linear(in features=256, out features=K, bias=True)  
    
    \item \textbf{VGG16} \newline
Conv2d(I, 64, kernel size=(3, 3), stride=(1, 1), padding=(1, 1)) \newline BatchNorm2d(64, eps=1e-05, momentum=0.1, affine=True, track running stats=True) \newline LeakyReLU(negative slope=0.05, inplace=True) \newline Conv2d(64, 64, kernel size=(3, 3), stride=(1, 1), padding=(1, 1)) \newline BatchNorm2d(64, eps=1e-05, momentum=0.1, affine=True, track running stats=True) \newline LeakyReLU(negative slope=0.05, inplace=True) \newline MaxPool2d(kernel size=2, stride=2, padding=0, dilation=1, ceil mode=False) \newline Conv2d(64, 128, kernel size=(3, 3), stride=(1, 1), padding=(1, 1)) \newline BatchNorm2d(128, eps=1e-05, momentum=0.1, affine=True, track running stats=True) \newline LeakyReLU(negative slope=0.05, inplace=True) \newline Conv2d(128, 128, kernel size=(3, 3), stride=(1, 1), padding=(1, 1)) \newline BatchNorm2d(128, eps=1e-05, momentum=0.1, affine=True, track running stats=True) \newline LeakyReLU(negative slope=0.05, inplace=True) \newline MaxPool2d(kernel size=2, stride=2, padding=0, dilation=1, ceil mode=False) \newline Conv2d(128, 256, kernel size=(3, 3), stride=(1, 1), padding=(1, 1)) \newline BatchNorm2d(256, eps=1e-05, momentum=0.1, affine=True, track running stats=True) \newline LeakyReLU(negative slope=0.05, inplace=True) \newline Conv2d(256, 256, kernel size=(3, 3), stride=(1, 1), padding=(1, 1)) \newline BatchNorm2d(256, eps=1e-05, momentum=0.1, affine=True, track running stats=True) \newline LeakyReLU(negative slope=0.05, inplace=True) \newline Conv2d(256, 256, kernel size=(3, 3), stride=(1, 1), padding=(1, 1)) \newline BatchNorm2d(256, eps=1e-05, momentum=0.1, affine=True, track running stats=True) \newline LeakyReLU(negative slope=0.05, inplace=True) \newline MaxPool2d(kernel size=2, stride=2, padding=0, dilation=1, ceil mode=False) \newline Conv2d(256, 512, kernel size=(3, 3), stride=(1, 1), padding=(1, 1)) \newline BatchNorm2d(512, eps=1e-05, momentum=0.1, affine=True, track running stats=True) \newline LeakyReLU(negative slope=0.05, inplace=True) \newline Conv2d(512, 512, kernel size=(3, 3), stride=(1, 1), padding=(1, 1)) \newline BatchNorm2d(512, eps=1e-05, momentum=0.1, affine=True, track running stats=True) \newline LeakyReLU(negative slope=0.05, inplace=True) \newline Conv2d(512, 512, kernel size=(3, 3), stride=(1, 1), padding=(1, 1)) \newline BatchNorm2d(512, eps=1e-05, momentum=0.1, affine=True, track running stats=True) \newline LeakyReLU(negative slope=0.05, inplace=True) \newline MaxPool2d(kernel size=2, stride=2, padding=0, dilation=1, ceil mode=False) \newline Conv2d(512, 512, kernel size=(3, 3), stride=(1, 1), padding=(1, 1)) \newline BatchNorm2d(512, eps=1e-05, momentum=0.1, affine=True, track running stats=True) \newline LeakyReLU(negative slope=0.05, inplace=True) \newline Conv2d(512, 512, kernel size=(3, 3), stride=(1, 1), padding=(1, 1)) \newline BatchNorm2d(512, eps=1e-05, momentum=0.1, affine=True, track running stats=True) \newline LeakyReLU(negative slope=0.05, inplace=True) \newline Conv2d(512, 1024, kernel size=(3, 3), stride=(1, 1), padding=(1, 1)) \newline BatchNorm2d(1024, eps=1e-05, momentum=0.1, affine=True, track running stats=True) \newline LeakyReLU(negative slope=0.05, inplace=True) \newline AdaptiveAvgPool2d(output size=(2, 2)) \newline Dropout(p=0.3, inplace=False) \newline Linear(in features=4096, out features=4096, bias=True) \newline BatchNorm1d(4096, eps=1e-05, momentum=0.1, affine=True, track running stats=True) \newline LeakyReLU(negative slope=0.05, inplace=True) \newline Dropout(p=0.3, inplace=False) \newline Linear(in features=4096, out features=4096, bias=True) \newline BatchNorm1d(4096, eps=1e-05, momentum=0.1, affine=True, track running stats=True) \newline LeakyReLU(negative slope=0.05, inplace=True) \newline Dropout(p=0.3, inplace=False) \newline Linear(in features=4096, out features=K, bias=True)  

    \item \textbf{AlexNet} \newline
Conv2d(I, 64, kernel size=(5, 5), stride=(1, 1), padding=(2, 2)) \newline BatchNorm2d(64, eps=1e-05, momentum=0.1, affine=True, track running stats=True) \newline LeakyReLU(negative slope=0.05) \newline MaxPool2d(kernel size=3, stride=2, padding=0, dilation=1, ceil mode=False) \newline Conv2d(64, 192, kernel size=(5, 5), stride=(1, 1), padding=(2, 2)) \newline BatchNorm2d(192, eps=1e-05, momentum=0.1, affine=True, track running stats=True) \newline LeakyReLU(negative slope=0.05) \newline MaxPool2d(kernel size=3, stride=2, padding=0, dilation=1, ceil mode=False) \newline Conv2d(192, 384, kernel size=(3, 3), stride=(1, 1), padding=(1, 1)) \newline BatchNorm2d(384, eps=1e-05, momentum=0.1, affine=True, track running stats=True) \newline LeakyReLU(negative slope=0.05) \newline Conv2d(384, 256, kernel size=(3, 3), stride=(1, 1), padding=(1, 1)) \newline BatchNorm2d(256, eps=1e-05, momentum=0.1, affine=True, track running stats=True) \newline LeakyReLU(negative slope=0.05) \newline Conv2d(256, 512, kernel size=(3, 3), stride=(1, 1), padding=(1, 1)) \newline BatchNorm2d(512, eps=1e-05, momentum=0.1, affine=True, track running stats=True) \newline LeakyReLU(negative slope=0.05) \newline AdaptiveAvgPool2d(output size=(2, 2)) \newline Dropout(p=0.3, inplace=False) \newline Linear(in features=2048, out features=4096, bias=True) \newline BatchNorm1d(4096, eps=1e-05, momentum=0.1, affine=True, track running stats=True) \newline LeakyReLU(negative slope=0.05) \newline Dropout(p=0.3, inplace=False) \newline Linear(in features=4096, out features=4096, bias=True) \newline BatchNorm1d(4096, eps=1e-05, momentum=0.1, affine=True, track running stats=True) \newline LeakyReLU(negative slope=0.05) \newline Dropout(p=0.3, inplace=False) \newline Linear(in features=4096, out features=K, bias=True) 
\end{itemize}{}

\end{document}